\documentclass[10pt,twocolumn,letterpaper]{article}

\usepackage{wacv}
\usepackage[accsupp]{axessibility}

\newcommand{\preenPSNR}{24.63}
\newcommand{\preenSSIM}{0.914}
\newcommand{\preenHCorr}{0.473}
\newcommand{\preenLPIPS}{0.094}
\newcommand{\vividPSNR}{22.27}
\newcommand{\vividSSIM}{0.899}
\newcommand{\vividHCorr}{0.401}
\newcommand{\vividLPIPS}{0.127}
\newcommand{\phaseOnePSNR}{23.60}
\newcommand{\phaseOneSSIM}{0.907}
\newcommand{\phaseOneHCorr}{0.383}
\newcommand{\phaseOneLPIPS}{0.123}
\newcommand{\dpRasterPSNR}{23.29}
\newcommand{\dpRasterSSIM}{0.893}
\newcommand{\dpRasterHCorr}{0.411}
\newcommand{\dpRasterLPIPS}{0.107}
\newcommand{\dpPresetPSNR}{22.71}
\newcommand{\dpPresetSSIM}{0.895}
\newcommand{\dpPresetHCorr}{0.385}
\newcommand{\dpPresetLPIPS}{0.117}
\newcommand{\latencyMs}{283}

\usepackage{tikz}
\usetikzlibrary{arrows.meta, positioning, calc}
\pgfdeclarelayer{fg}
\pgfsetlayers{main,fg}

\definecolor{wacvblue}{rgb}{0.21,0.49,0.74}
\usepackage[pagebackref,breaklinks,colorlinks,allcolors=wacvblue]{hyperref}

\newcommand{\PREENtitle}{PREEN: Exemplar-Based Color Style Transfer with Smartphone Editor Presets}
\newcommand{\PREENauthor}{Beomjin Ahn\\
Samsung Research\\
Seoul, Korea\\
{\tt\small beomjin.ahn@samsung.com}
}

\edef\PREENsupplayout{%
  \textfloatsep=\the\textfloatsep\relax
  \floatsep=\the\floatsep\relax
  \intextsep=\the\intextsep\relax
  \dbltextfloatsep=\the\dbltextfloatsep\relax
  \dblfloatsep=\the\dblfloatsep\relax
  \abovecaptionskip=\the\abovecaptionskip\relax}
\newcommand{\PREENmainlayout}{%
  \setlength{\textfloatsep}{9pt plus 2pt minus 3pt}%
  \setlength{\floatsep}{6pt plus 2pt minus 2pt}%
  \setlength{\intextsep}{6pt plus 2pt minus 2pt}%
  \setlength{\dbltextfloatsep}{9pt plus 2pt minus 3pt}%
  \setlength{\dblfloatsep}{6pt plus 2pt minus 2pt}%
  \setlength{\abovecaptionskip}{4pt}%
}

\newcommand{\PREENsuppnumbering}{%
  \renewcommand{\thetable}{S\arabic{table}}%
  \renewcommand{\thefigure}{S\arabic{figure}}}

\PREENmainlayout

\title{\PREENtitle}
\author{\PREENauthor}

\makeatletter
\AtBeginDocument{%
  \gdef\@extra@binfo{}%
  \immediate\write\@auxout{\string\gdef\string\@extra@binfo{}}%
  \global\let\PREEN@maketitle\maketitle
  \global\let\PREEN@@maketitle\@maketitle
  \global\let\PREEN@title\title
  \global\let\PREEN@author\author
  \global\let\PREEN@and\and
  \global\let\PREEN@thanks\thanks
}
\newcommand{\PREENbeginsupp}{%
  \clearpage
  \setcounter{page}{1}%
  \setcounter{section}{0}%
  \setcounter{subsection}{0}%
  \setcounter{subsubsection}{0}%
  \setcounter{figure}{0}%
  \setcounter{table}{0}%
  \setcounter{equation}{0}%
  \setcounter{footnote}{0}%
  \PREENsuppnumbering
  \PREENsupplayout
  \gdef\@extra@b@citeb{@supp}%
  \immediate\write\@auxout{\string\gdef\string\@extra@binfo{@supp}}%
  \def\theHsection{supp.\arabic{section}}%
  \def\theHfigure{supp.\arabic{figure}}%
  \def\theHtable{supp.\arabic{table}}%
  \def\theHequation{supp.\arabic{equation}}%
  \def\theHfootnote{supp.\arabic{footnote}}%
  \def\Hy@PageAnchorSlide{\def\thepage{supp.\arabic{page}}}%
  \def\backrefxxx##1##2##3{\hyperlink{page.supp.##1}{##1}}%
  \global\let\maketitle\PREEN@maketitle
  \global\let\@maketitle\PREEN@@maketitle
  \global\let\title\PREEN@title
  \global\let\author\PREEN@author
  \global\let\and\PREEN@and
  \global\let\thanks\PREEN@thanks
  \let\thetitle\@undefined
  \title{\PREENtitle\\
(Supplementary Material)}%
  \author{\PREENauthor}%
}
\makeatother

\begin{document}
\maketitle

\begin{abstract}
We present \textbf{PREEN} (Preset Regression via an Exemplar-based Editor Network), which casts color style transfer as regression onto the $29$ parameters of a deterministic image editor. Given a content image and a style reference, PREEN predicts a preset, a vector of editor slider values that is interpretable and editor-consumable, not a raster image. The key obstacle is supervision, since an arbitrary real reference lacks a preset ground truth. We therefore supervise with a style-transfer teacher, whose target looks we invert into the editor's parameter space as pseudo-presets, then pretrain a CLIP-conditioned regressor and refine it with LoRA under an image-level CIELAB loss. On the human-curated PST50 benchmark, PREEN applies the reference's color while preserving structural fidelity, leading the baselines on pixel- and perceptual-fidelity metrics. The preset is compact and predicted at sub-second on-device latency. The low-dimensional subspace also exposes a single inference-time strength knob that traces a continuous default-to-vivid trajectory in one forward pass, with every intermediate state a valid editor preset. A two-axis human study shows that raters prefer the vivid end for matching the reference but the default as a standalone photo, a trade-off the knob spans without retraining.
\end{abstract}

\section{Introduction}

\begin{figure}[!t]
  \centering
  \setlength{\tabcolsep}{0pt}
  \renewcommand{\arraystretch}{0.45}
  \newcommand{\cellwidth}{0.32\linewidth}
  \begin{tabular}{@{}c@{\hspace{2pt}}c@{\hspace{5pt}}c@{}}
    \small\bfseries Input & \small\bfseries Reference & \small\bfseries PREEN \\[0.5pt]
    \includegraphics[width=\cellwidth]{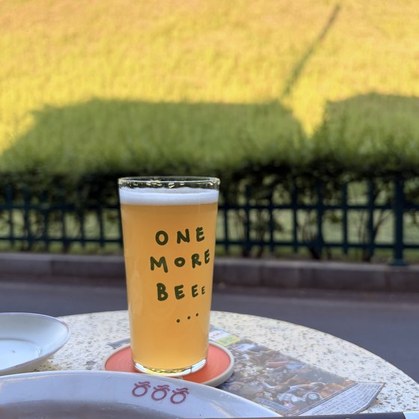} & \includegraphics[width=\cellwidth]{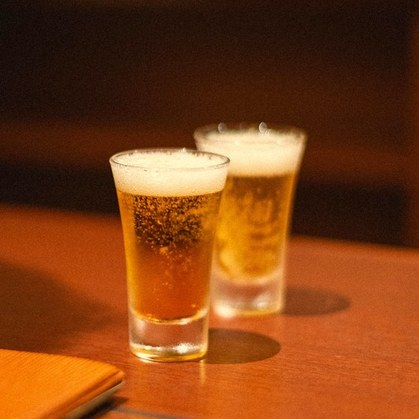} & \includegraphics[width=\cellwidth]{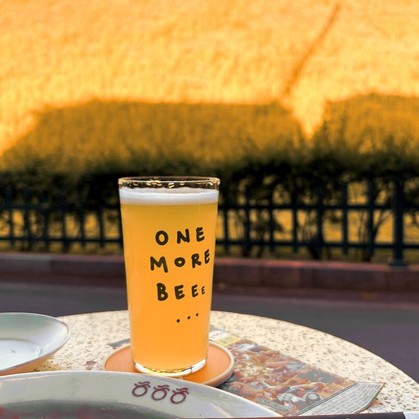} \\
    \includegraphics[width=\cellwidth]{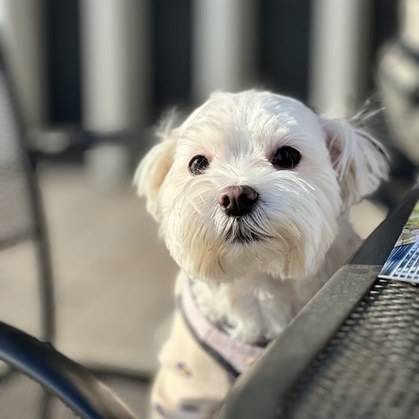} & \includegraphics[width=\cellwidth]{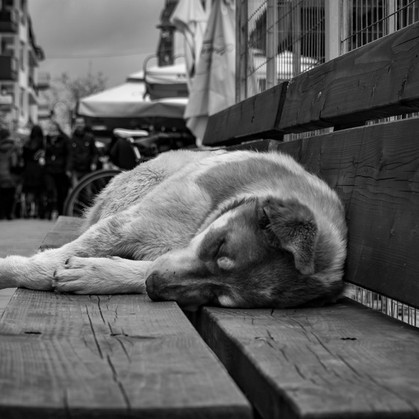} & \includegraphics[width=\cellwidth]{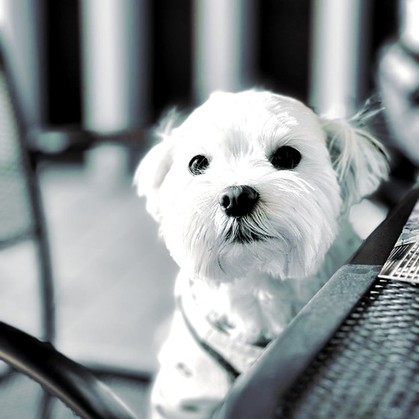} \\
    \includegraphics[width=\cellwidth]{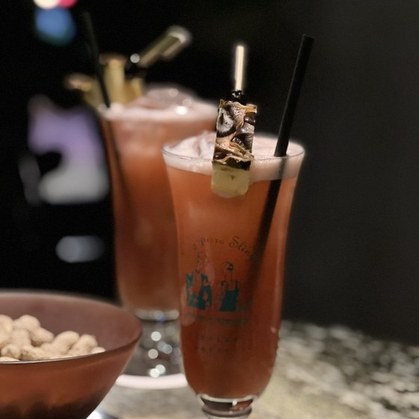} & \includegraphics[width=\cellwidth]{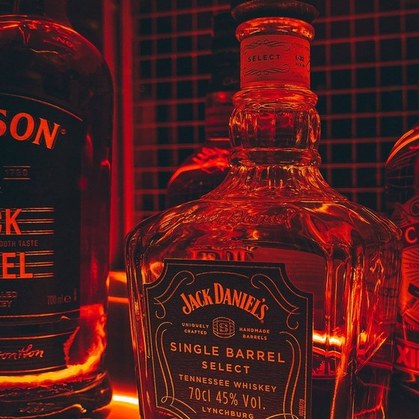} & \includegraphics[width=\cellwidth]{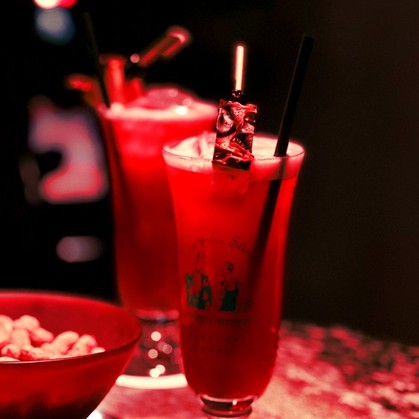} \\
    \includegraphics[width=\cellwidth]{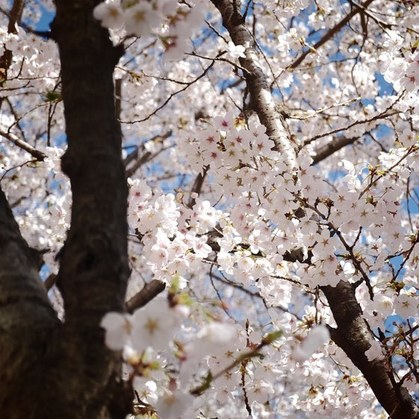} & \includegraphics[width=\cellwidth]{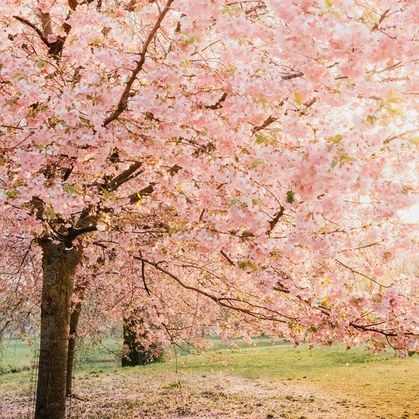} & \includegraphics[width=\cellwidth]{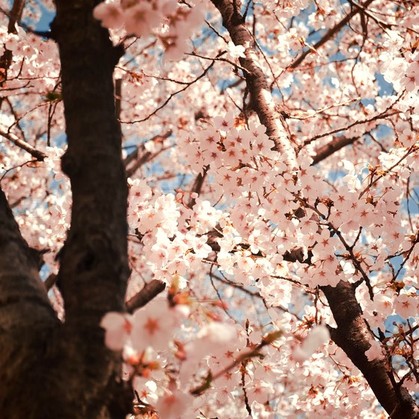} \\
  \end{tabular}
  \caption{PREEN restyles an input toward a reference via a predicted $29$-parameter editor preset (shown at the Vivid strength setting).}
  \label{fig:teaser}
\end{figure}

\begin{figure*}[!t]
\centering
\includegraphics[width=\linewidth]{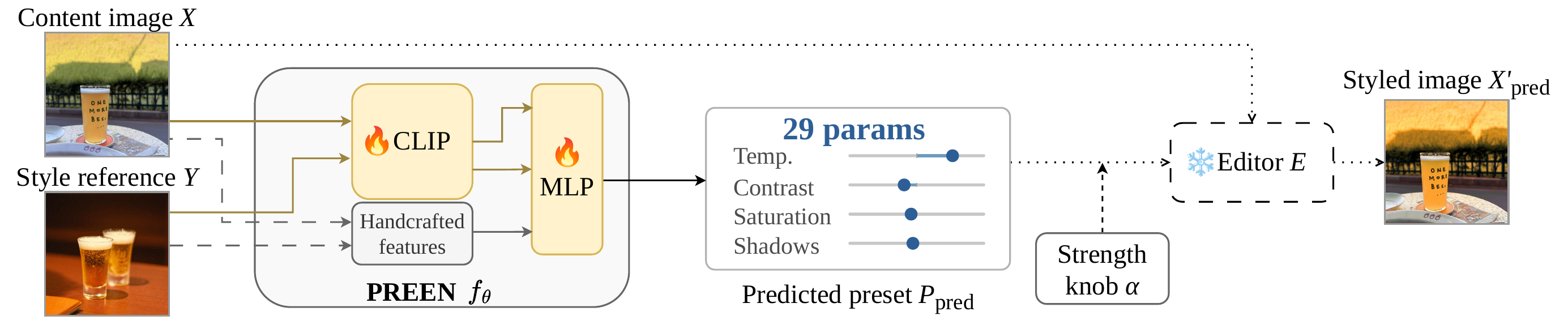}
\caption{Overview of PREEN. Given a content image $X$ and a style reference $Y$, PREEN predicts a $29$-parameter editor preset $P_{\mathrm{pred}}$ in a single forward pass. Applying $P_{\mathrm{pred}}$ with our differentiable editor $E$ yields $X'_{\mathrm{pred}}$; the strength knob $\alpha$ scales the preset's chroma parameters before editing. PREEN's output is a compact, human-editable preset rather than a rendered image. Flame icons mark the modules updated during training (CLIP only through low-rank adapters), and the snowflake marks the fixed editor.}
\label{fig:overview}
\end{figure*}

Exemplar-based color style transfer, also studied as photorealistic style transfer (PST), is a routine operation in modern photo editing. A user picks a reference image $Y$ whose look they admire, wanting their own shot $X$ recolored to match it (Figure~\ref{fig:teaser}). Mainstream academic approaches are pixel-generative, from neural style transfer~\cite{huang2017adain} to diffusion editing~\cite{meng2021sdedit}. Whatever their quality, they emit rasters with no exposed editing parameters, and the diffusion-based ones add multi-second latency and stochastic sampling that resists deterministic re-editing, an awkward fit for on-device, user-in-the-loop consumer pipelines.

An alternative is photo preset transfer~\cite{ho2020deeppreset}, predicting the parameters $P$ of a deterministic image editor $E$ from the pair $(X, Y)$ such that $E(X, P)$ approximates the look of $Y$. The output $P$, a vector of slider values, is interpretable and compact. Unlike a rendered image, it can be stored and re-edited, with no generative model at inference time.

The central challenge is supervision. Parametric retouching benchmarks such as MIT-Adobe FiveK~\cite{bychkovsky2011fivek} and PPR10K~\cite{liang2021ppr10k} provide human-authored target edits, but operate in the auto-enhancement regime without a separate style reference. No large-scale dataset pairs a content image with a reference and a matching expert-graded target. The principal prior method~\cite{ho2020deeppreset} sidesteps this by sampling presets, thereby bounding the styles it learns; the only reference-conditioned set with expert-graded targets we know of~\cite{gong2025salut} offers just 50 pairs (PST50) for evaluation.

We close this gap with teacher supervision. We propose \textbf{PREEN} (Preset Regression via an Exemplar-based Editor Network). At inference, it maps a content--reference pair $(X, Y)$ to a $29$-parameter preset that a smartphone editor applies to produce the result (Figure~\ref{fig:overview}). A two-stage pipeline trains it. Phase~1 pretrains a CLIP-conditioned regressor on a large corpus of pseudo-labeled triples recovered from a style-transfer teacher. Phase~2 freezes all base weights and trains only low-rank adapters under an image-level CIELAB loss evaluated on the differentiable editor output, addressing the perceptual non-uniformity of Phase~1's parameter-space loss.

Our contributions are: (i)~a teacher-supervised recipe that frees reference-based preset regression from sampled-preset supervision, substituting a style-transfer teacher on real references; (ii)~a strength-interpolation primitive, an inference-time scalar $\alpha$ on the chromaticity parameters that traces a continuous default-to-vivid trajectory without re-running inference; and (iii)~a two-axis human study surfacing a three-way fidelity\,/\,style-match\,/\,aesthetic divergence across the $\alpha$ trajectory.

\section{Related Work}

\subsection{Photo Preset Transfer}
Editing photos by predicting the parameters of interpretable white-box operations has a long history. Learning-based automatic retouching predicts sequences of editing operations~\cite{hu2018exposure}, and the same idea recurs in compositing harmonization~\cite{ke2022harmonizer}. Both regress parameters toward a fixed goal, enhancement or in-image harmonization, rather than a separate style reference. Deep Preset~\cite{ho2020deeppreset} established the reference-conditioned setting we adopt, regressing an editor preset (69~develop controls) from a content--reference pair, but its supervision is \emph{synthetic by construction}. Training references are images re-rendered with sampled presets, so a ground-truth preset always exists, but the styles it can learn are confined to that sampled set. The setting has remained underexplored since then; the closest later work, NCST~\cite{jiang2024ncst}, regresses a smaller global-only parameter set from a reference to perform video grading. We revisit it \emph{without} sampled-preset supervision: a style-transfer teacher on real reference pairs supplies targets we invert into the editor's parameter space as pseudo-presets (Section~\ref{sec:method}).

\begin{table*}[!t]
\centering
\caption{\label{tab:main} PST50 (paired) metrics. The Output column gives each method's edit representation, and the Dim column gives its dimensionality. Raster outputs have no fixed dimension because their size scales with the image resolution. Time is the per-image CPU latency (supplementary); for LUT and preset methods it covers generating the LUT or preset, not applying it. Bold marks the best value per metric column. Per-image Wilcoxon significance is reported in the supplementary.}
\small
\setlength{\tabcolsep}{4pt}
\begin{tabular}{@{}l@{\hspace{13pt}}l@{\hspace{13pt}}l@{\hspace{16pt}}c c c c c r@{}}
\toprule
\textbf{Method} & \textbf{Output} & \multicolumn{1}{c}{\textbf{Dim}} & \textbf{PSNR$\uparrow$} & \textbf{SSIM$\uparrow$} & \textbf{HCorr$\uparrow$} & \textbf{LPIPS$\downarrow$} & $\mathbf{\Delta E^{*}_{ab}\downarrow}$ & \textbf{Time (s)} \\
\midrule
\textit{No edit (identity)}                        & -- & -- & 22.67 & 0.906 & 0.348 & 0.127 & 12.32 & -- \\
\midrule
Reinhard~\cite{reinhard2001color}                  & Raster & -- & 20.13 & 0.835 & 0.447 & 0.122 & 13.40 & 0.004 \\
AdaIN~\cite{huang2017adain}                        & Raster & -- & 19.06 & 0.704 & 0.504 & 0.314 & 14.38 & 0.12 \\
Deep Preset ($\hat{Y}$)~\cite{ho2020deeppreset}   & Raster & -- & \dpRasterPSNR & \dpRasterSSIM & \dpRasterHCorr & \dpRasterLPIPS & 10.53 & 0.19 \\
ModFlows~\cite{larchenko2025modflows}               & Raster & -- & 19.98 & 0.853 & 0.510 & 0.118 & 12.77 & 1.57 \\
\midrule
NLUT~\cite{nlut}                                   & LUT & $\mathbb{R}^{3{\times}33^3}$ & 23.43 & 0.885 & \textbf{0.545} & 0.129 & 9.81 & N/A$^{\dagger}$ \\
SA-LUT~\cite{gong2025salut}                        & LUT & $\mathbb{R}^{3{\times}2{\times}17^3}$ & 22.54 & 0.884 & 0.516 & 0.133 & 9.63 & 0.73 \\
\midrule
Deep Preset ($\hat{P}$)~\cite{ho2020deeppreset}           & Preset & $\mathbb{R}^{69}$ & \dpPresetPSNR & \dpPresetSSIM & \dpPresetHCorr & \dpPresetLPIPS & 12.13 & 0.10 \\
\textbf{PREEN (Ours)}               & Preset & $\mathbb{R}^{29}$ & \textbf{\preenPSNR} & \textbf{\preenSSIM} & {\preenHCorr} & \textbf{\preenLPIPS} & \textbf{9.07} & {0.36} \\
\textbf{PREEN Vivid (Ours, $\alpha\!=\!5$)}                   & Preset & $\mathbb{R}^{29}$ & {\vividPSNR} & {\vividSSIM} & {\vividHCorr} & {\vividLPIPS} & 12.18 & {0.36} \\
\bottomrule
\end{tabular}

{\footnotesize\raggedright $^{\dagger}$NLUT requires a CUDA-only operator, so its metrics are measured on GPU, and CPU timing is not available.\par}
\end{table*}

\subsection{Look-Up-Table Stylization}
A separate line maps colors through a learned look-up table (LUT), a discretized grid over the RGB cube. Image-adaptive 3D LUT~\cite{zeng2020lut} and NLUT~\cite{nlut} predict one such grid applied uniformly across the image, and AceTone~\cite{ma2026acetone} predicts one from a reference with a vision-language model. Spatially aware variants blend multiple LUTs per pixel~\cite{wang2021lut}, and SA-LUT~\cite{gong2025salut} extends these to a spatially adaptive 4D LUT, which we adopt as our teacher. SA-LUT also contributes the human-curated PST50 benchmark used for our evaluation (Section~\ref{sec:results:eval}). Either way, the output is a dense grid of $10^4$--$10^5$ values (Table~\ref{tab:main}) rather than a few named sliders, leaving the edit hard for a user to inspect or refine.

\subsection{Pixel-Generative Stylization}
A large body of work renders stylization through pixel transformations, from classic global color-statistic transfer~\cite{reinhard2001color} to AdaIN~\cite{huang2017adain}, diffusion-based editing such as SDEdit~\cite{meng2021sdedit}, and recent neural color-style methods including Neural Preset~\cite{ke2023neuralpreset}, ModFlows~\cite{larchenko2025modflows}, and INRetouch~\cite{elezabi2024inretouch}. Concurrent work pushes further into generative modeling. SemiNFT~\cite{yang2026seminft} casts reference preset transfer as a Diffusion-Transformer task refined with reinforcement learning for aesthetic perception. These methods emit a raster image directly, without exposing user-editable parameters.

Neither the LUT nor the pixel-generative family yields an editable preset; we compare experimentally against both families and against Deep Preset, which targets the same task.

\section{Method}
\label{sec:method}

\subsection{Problem Formulation}
\label{sec:method:problem}

Let $E:\mathcal{X}\times\mathcal{P}\to\mathcal{X}$ denote a deterministic image editor mapping a content image $X$ and a parameter vector $P\in[-1,1]^d$ to an edited image $E(X,P)$. We target a $d=29$ editor representative of consumer mobile photo applications: $8$ global controls (temperature, tint, brightness, exposure, contrast, highlights, shadows, saturation) and $21$ per-hue HSL channels (hue/saturation/luminance for each of $7$ color zones: red, orange, yellow, green, cyan, blue, magenta). We split the $29$ parameters into a tone set $\mathcal{I}_L$ ($12$: contrast, brightness, exposure, highlights, shadows, per-hue luminance) and a chromaticity set $\mathcal{I}_{\mathrm{ab}}$ ($17$: temperature, tint, saturation, per-hue hue/saturation). Equivalently, the $8$ global controls split $5$ tone\,/\,$3$ chroma and the $21$ per-hue channels split $7$\,/\,$14$, a split reused by the regressor's two output trunks (Section~\ref{sec:method:phase1}) and the $\alpha$ control (Section~\ref{sec:method:alpha}). We implement these controls, inspired by consumer smartphone photo editors, as a differentiable module (stage-by-stage pipeline in the supplementary), so both the Phase~1 inversion (Eq.~\ref{eq:inversion}) and the Phase~2 image-level loss can backpropagate through $E$, the same module that produces every PREEN result we report. Given $(X, Y)$, the goal is a regressor $f_\theta(X,Y)=P_{\mathrm{pred}}$ such that $X'_{\mathrm{pred}}=E(X,P_{\mathrm{pred}})$ approximates the look of $Y$.

\subsection{Teacher Supervision}

We instantiate the teacher with SA-LUT~\cite{gong2025salut}, a spatially adaptive 4D LUT producing a pseudo-target $\tilde{X}'=T(X,Y)$ for any pair. A LUT teacher is structure-preserving and dominated by color/tone transforms that lie close to our editor's subspace, unlike a pixel-generative teacher's edits. Both phases use the published teacher unmodified; a spatially global variant appears only in the ablation (Table~\ref{tab:teacher}).
\begin{figure*}[!t]
    \centering
    \setlength{\tabcolsep}{0pt}
    \renewcommand{\arraystretch}{0.45}
    \newcommand{\cellwidth}{0.1615\linewidth}
    \begin{tabular}{@{}c@{\hspace{1.5pt}}c@{\hspace{5pt}}c@{\hspace{1.5pt}}c@{\hspace{1.5pt}}c@{\hspace{5pt}}c@{}}
      \small\bfseries $X$ & \small\bfseries $Y$ & \small\bfseries SA-LUT & \small\bfseries Deep Preset & \small\bfseries PREEN & \small\bfseries Target \\[0.5pt]
      \includegraphics[width=\cellwidth]{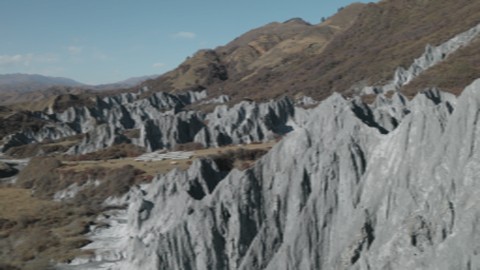} &
      \includegraphics[width=\cellwidth]{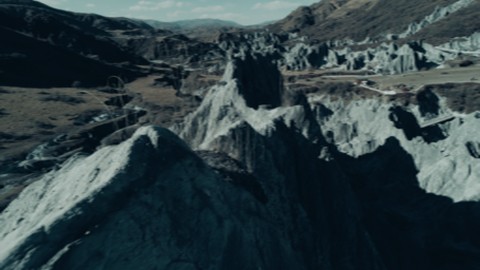} &
      \includegraphics[width=\cellwidth]{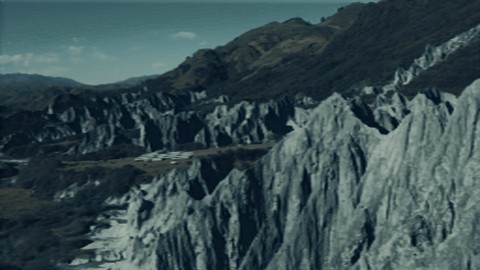} &
      \includegraphics[width=\cellwidth]{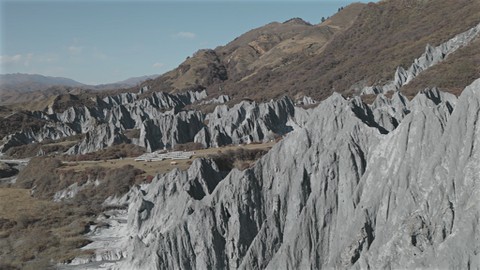} &
      \includegraphics[width=\cellwidth]{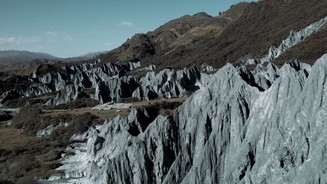} &
      \includegraphics[width=\cellwidth]{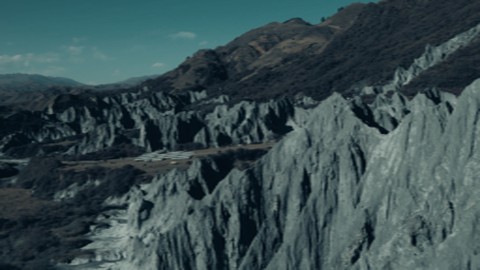} \\
      \includegraphics[width=\cellwidth]{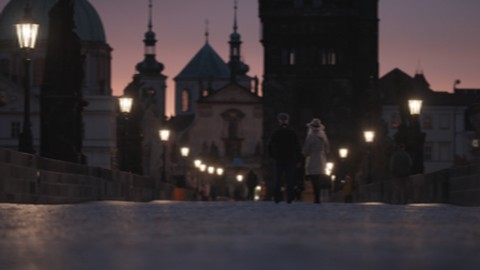} &
      \includegraphics[width=\cellwidth]{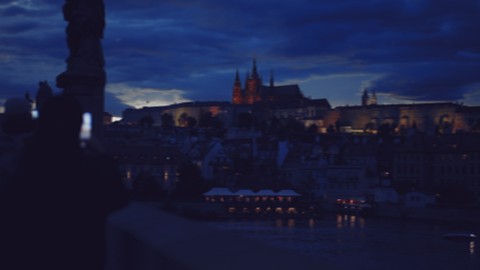} &
      \includegraphics[width=\cellwidth]{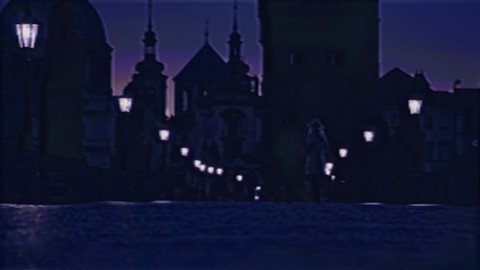} &
      \includegraphics[width=\cellwidth]{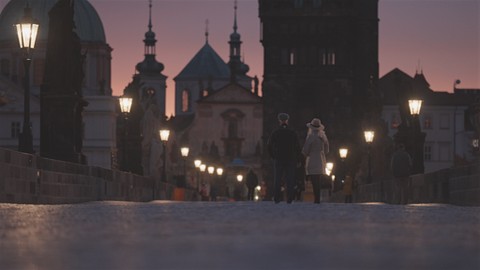} &
      \includegraphics[width=\cellwidth]{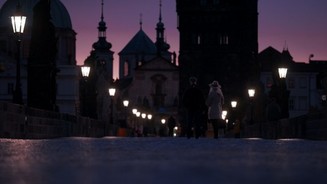} &
      \includegraphics[width=\cellwidth]{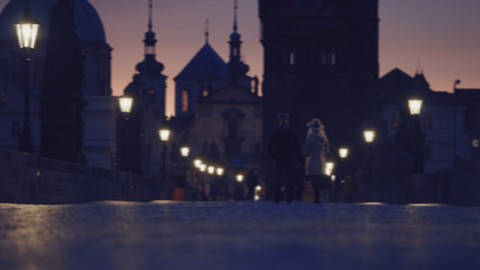} \\
      \includegraphics[width=\cellwidth]{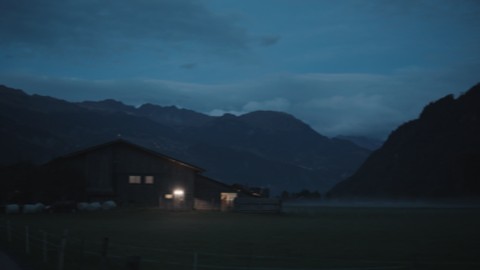} &
      \includegraphics[width=\cellwidth]{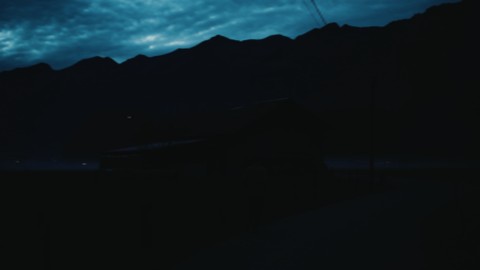} &
      \includegraphics[width=\cellwidth]{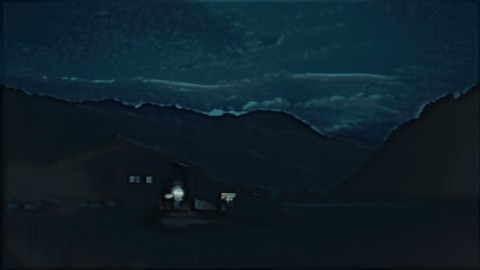} &
      \includegraphics[width=\cellwidth]{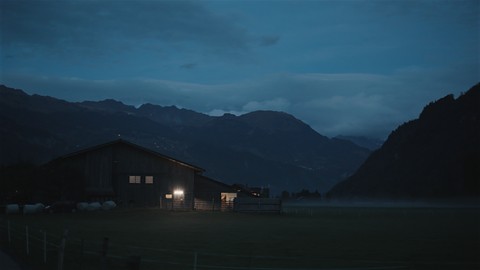} &
      \includegraphics[width=\cellwidth]{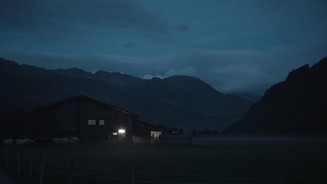} &
      \includegraphics[width=\cellwidth]{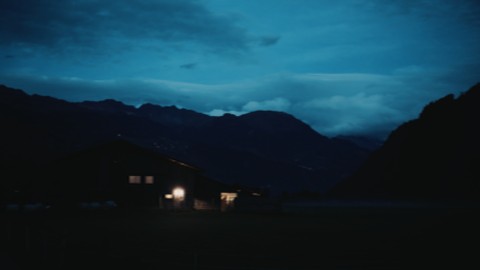} \\
      \includegraphics[width=\cellwidth]{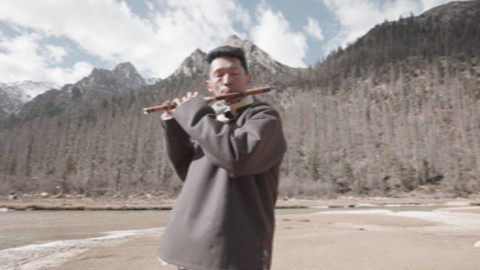} &
      \includegraphics[width=\cellwidth]{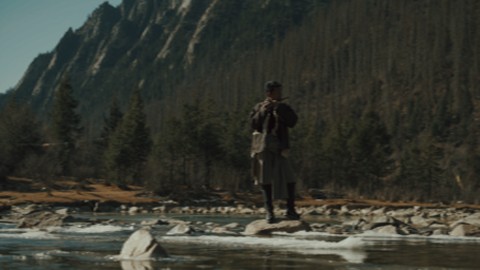} &
      \includegraphics[width=\cellwidth]{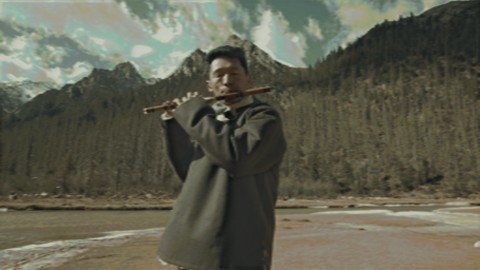} &
      \includegraphics[width=\cellwidth]{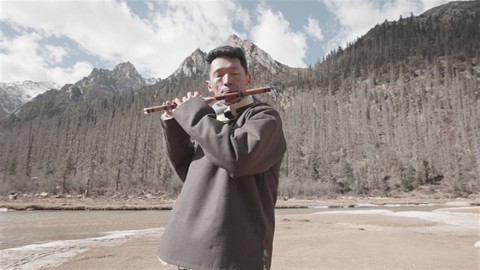} &
      \includegraphics[width=\cellwidth]{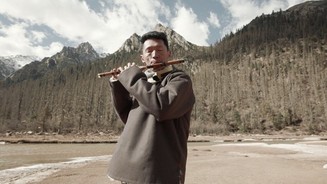} &
      \includegraphics[width=\cellwidth]{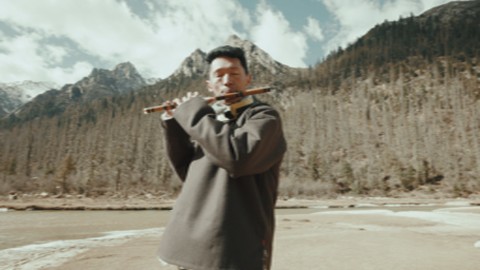} \\
      \includegraphics[width=\cellwidth]{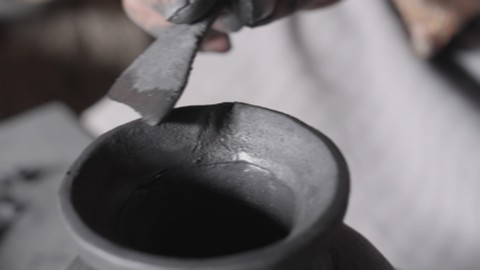} &
      \includegraphics[width=\cellwidth]{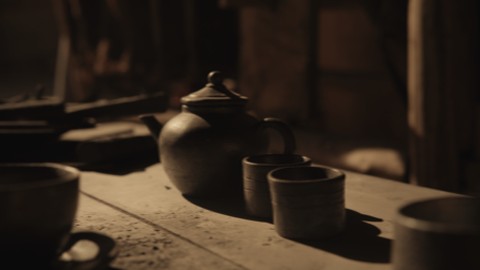} &
      \includegraphics[width=\cellwidth]{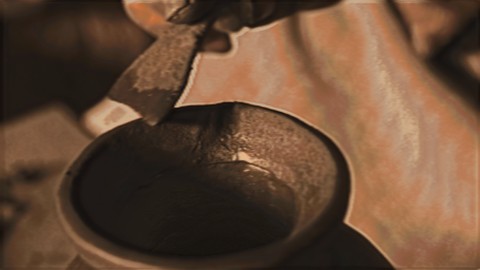} &
      \includegraphics[width=\cellwidth]{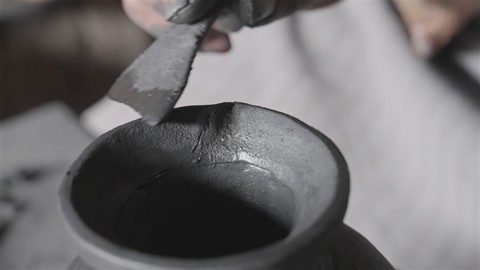} &
      \includegraphics[width=\cellwidth]{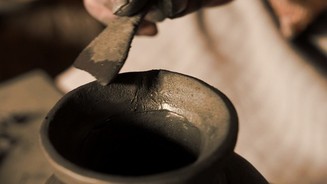} &
      \includegraphics[width=\cellwidth]{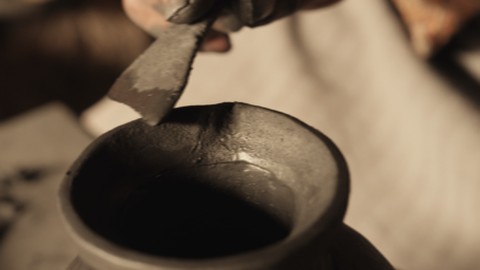} \\
    \end{tabular}
    \caption{Head-to-head on PST50 (first five samples). Columns: content $X$, reference $Y$, SA-LUT~\cite{gong2025salut}, Deep Preset ($\hat{P}$)~\cite{ho2020deeppreset}, PREEN ($\alpha\!=\!1$), and the expert target.}
    \label{fig:qual}
\end{figure*}

\subsection{Phase 1: Pseudo-Label Recovery and Regression}
\label{sec:method:phase1}

For each $(X,\tilde{X}')$, we project the teacher's richer edit into the editor's constrained $29$-parameter subspace, recovering a pseudo-preset $\tilde{P}$ by gradient-based optimization through the differentiable editor~\cite{tseng2019hyperparameter}:
\begin{equation}
    \tilde{P} = \arg\min_{P\in[-1,1]^{29}}\, \mathcal{L}_{\mathrm{rec}}\!\bigl(E(X,P),\,\tilde{X}'\bigr),
    \label{eq:inversion}
\end{equation}
where $\mathcal{L}_{\mathrm{rec}} = \mathrm{MSE}_{\mathrm{RGB}} + \lambda_{\mathrm{lab}}\bigl(\mathrm{MSE}_{L^{*}} + \mathrm{MSE}_{a^{*}b^{*}}\bigr)$, each term the mean-squared error between $E(X,P)$ and $\tilde{X}'$ over the indicated channels, an RGB anchor plus CIELAB luminance and chromaticity, and the weight $\lambda_{\mathrm{lab}}$ is set on a held-out subset. We run AdamW for a fixed step budget per pair, yielding the $(X, Y, \tilde{P})$ triples on which the regressor below is trained.

The regressor $f_\theta$ is a frozen CLIP visual encoder~\cite{radford2021clip} feeding a simple residual MLP. Both $X$ and $Y$ pass through the shared encoder, and we concatenate the per-image $768$-d CLIP image embeddings (the projected encoder output) with five handcrafted color-statistic features of $X$ and $Y$: (i)~the RGB log-ratio $\log(\mu_Y/\mu_X)$, (ii)~the HSL channel-mean difference, (iii)~low-level RGB mean/std, (iv)~a soft RGB color histogram, and (v)~per-channel RGB percentile differences. These low-frequency color statistics supplement CLIP's semantic features. The MLP runs two parallel residual trunks over this shared vector, each $[1024,1024,1024,256]$: a tone trunk ending in a linear head for the $\mathcal{I}_L$ parameters and a chroma trunk for the $\mathcal{I}_{\mathrm{ab}}$ parameters, each head applying Softsign so $P_{\mathrm{pred}}\in[-1,1]^{29}$. Phase~1 minimizes $\mathcal{L}_{\mathrm{ph1}}=H_{0.1}\!\bigl(P_{\mathrm{pred}}-\tilde{P}\bigr)+\lambda\|P_{\mathrm{pred}}\|_{2}^{2}$ with the CLIP backbone frozen, where $H_{0.1}$ is the Huber loss ($\beta\!=\!0.1$) and the $\ell_2$ term lightly regularizes the predicted preset.

\subsection{Phase 2: Adaptation under Image-Level Loss}

Phase~1 already yields a working regressor, but its parameter-space loss is not perceptually uniform. Because a $\pm0.1$ shift in brightness is far more visible than the same shift in a per-hue blue-hue parameter, Phase~2 refines the model with an image-level loss, matching the edited image so that capacity concentrates on the perceptually salient changes that the parameter-space loss underweights.

Phase~2 freezes all base weights of $f_\theta$ and trains only rank-16 LoRA adapters~\cite{hu2022lora} on every CLIP attention and FFN projection. Adapters of the same rank are also placed on every linear layer of the MLP. After training, the adapters are merged, leaving inference cost identical to that of Phase~1. The Phase~2 objective passes predicted presets through the differentiable editor and applies a CIELAB reconstruction loss between $X'_{\mathrm{pred}}=E(X,P_{\mathrm{pred}})$ and $\tilde{X}'$, plus an $\ell_2$ regularizer on $P_{\mathrm{pred}}$:
\begin{equation}
    \mathcal{L}_{\mathrm{ph2}} = w_{\mathrm{pix}}\mathcal{L}_{\mathrm{pix}} + w_{\mathrm{ab}}\mathcal{L}_{\mathrm{ab}} + w_{L}\mathcal{L}_{L} + \lambda\|P_{\mathrm{pred}}\|_{2}^{2},
    \label{eq:phase2_loss}
\end{equation}
where $\mathcal{L}_{\mathrm{ab}}$, $\mathcal{L}_{L}$, and the auxiliary anchor $\mathcal{L}_{\mathrm{pix}}$ are mean-squared errors in CIELAB $a^{*}b^{*}$, $L^{*}$, and RGB, with $(w_{\mathrm{pix}},w_{\mathrm{ab}},w_{L})=(0.2,5.0,1.0)$ set on a held-out subset. The term $\mathcal{L}_{L}$ backpropagates only into the tone trunk and $\mathcal{L}_{\mathrm{ab}}$ only into the chroma trunk (the complementary slice is detached). The RGB anchor reaches both. The $\ell_2$ regularizer reuses the Phase~1 weight $\lambda$ and keeps parameters at $0$ when they have little effect (e.g., HSL channels with no corresponding color present).

The Phase~2 target is the full teacher applied to an augmented reference $Y_{\mathrm{aug}}$ (a random crop/flip of $Y$; Section~\ref{sec:results:setup}), $\tilde{X}'=T(X,Y_{\mathrm{aug}})$. Identity samples ($Y\!=\!X$, at a $0.1$ ratio) use $X$ itself as the target.

\subsection{Strength Interpolation via \texorpdfstring{$\mathcal{I}_{\mathrm{ab}}$ $\alpha$-Scaling}{Iab Alpha-Scaling}}
\label{sec:method:alpha}

Operating in an explicit low-dimensional parameter subspace admits a parameter-space controllability primitive. At editing time, a scalar $\alpha$ rescales the chromaticity slice $\mathcal{I}_{\mathrm{ab}}$ of $P_{\mathrm{pred}}$ (Section~\ref{sec:method:problem}), leaving the tone set $\mathcal{I}_L$ unchanged, to dial edit intensity. We define
\begin{equation}
    P^{\alpha}_i = \begin{cases}
        \dfrac{\alpha\, P_i}{1+(\alpha-1)\,|P_i|}, & i \in \mathcal{I}_{\mathrm{ab}}, \\[4pt]
        P_i, & i \in \mathcal{I}_L.
    \end{cases}
\end{equation}
This form is simply the head's Softsign ($P\!=\!z/(1{+}|z|)$) re-applied after rescaling its pre-activation $z\!\mapsto\!\alpha z$, so $\alpha\!=\!1$ recovers $P_{\mathrm{pred}}$ and it inherits Softsign's sign- and $[-1,1]$-preservation with smooth saturation as $\alpha\!\to\!\infty$. Setting $\alpha\!>\!1$ thus intensifies the entire predicted color edit (saturation and hue shifts alike), producing a vivid, stronger variant that emphasizes color shift over fidelity to the teacher distribution.

Restricting the rescaling to chroma is deliberate, since rescaling luminance parameters as well co-saturates global tone, clips highlights, and crushes shadows (supplementary). Chroma-only scaling preserves the tonal anchor while the color identity intensifies, and one $f_\theta$ serves both modes.

\section{Experiments}

\subsection{Experimental Setup}
\label{sec:results:setup}

Content images $X$ are drawn from OpenImages~\cite{kuznetsova2020openimages}, and style references $Y$ from the top $20\%$ of the AVA aesthetics database~\cite{murray2012ava} by mean aesthetic score, so references are well-composed, attractively graded photos. We embed every image with CLIP ViT-L/14~\cite{radford2021clip} and pair each $X$ with one of its top-$10$ nearest references by cosine similarity, so the reference is semantically compatible, not arbitrary. This yields a few hundred thousand $(X, Y)$ training pairs, supervised by the SA-LUT~\cite{gong2025salut} teacher. The Phase~1 pseudo-presets $\tilde{P}$ are recovered via Eq.~\ref{eq:inversion}.

All optimizers are AdamW~\cite{loshchilov2019adamw}. Phase~1 inversion (Eq.~\ref{eq:inversion}) runs $1{,}000$ steps at learning rate $0.2$ per pair with the CIELAB weight $\lambda_{\mathrm{lab}}\!=\!20$, chosen on a held-out subset, and a small pre-activation magnitude decay ($10^{-3}$). Phase~1 regression uses learning rate $3{\times}10^{-5}$. Phase~2 trains rank-16 LoRA at learning rate $1{\times}10^{-4}$, batch size $64$, \texttt{bf16}, $\lambda = 10^{-2}$, on an NVIDIA L4 GPU. The augmentation $Y_{\mathrm{aug}}$ is a random square crop ($80$--$100\%$ of the short side) plus a horizontal flip.

\subsection{Evaluation}
\label{sec:results:eval}

We evaluate on PST50~\cite{gong2025salut}, a human-curated benchmark of (content, reference, expert-edited target) triples. We adopt the four metrics reported with it: PSNR, SSIM, HCorr (per-channel HSV histogram similarity to the target), and LPIPS~\cite{zhang2018lpips}, all computed on outputs and ground truth resized to $256{\times}256$, SA-LUT's training resolution. We additionally report $\Delta E^{*}_{ab}$, the mean CIELAB color difference to the expert target, used as an evaluation metric in the paired-target retouching literature~\cite{liang2021ppr10k,zeng2020lut}, where CIELAB-space training losses are also common~\cite{moran2020deeplpf,moran2021curl}.

\begin{table}[tb]
\centering
\caption{\label{tab:ablation} Pipeline ablation of PREEN on PST50. The top block trains a single phase; each remaining row changes one factor of the full pipeline (bottom row).}
\small
\setlength{\tabcolsep}{3pt}
\resizebox{\linewidth}{!}{%
\begin{tabular}{@{}l c c c c c@{}}
\toprule
\textbf{Setting} & \textbf{PSNR} & \textbf{SSIM} & \textbf{HCorr} & \textbf{LPIPS} & $\mathbf{\Delta E^{*}_{ab}}$ \\
\midrule
Phase~1 only    & \phaseOnePSNR & \phaseOneSSIM & \phaseOneHCorr & \phaseOneLPIPS & 10.60 \\
Phase~2 only & 24.43 & 0.918 & 0.453 & 0.105 & 9.12 \\
\midrule
MLP-only LoRA                 & 24.39 & 0.915 & 0.423 & 0.115 & 9.89 \\
w/o gradient routing          & 24.25 & 0.898 & 0.473 & 0.097 & 9.44 \\
\midrule
w/o identity samples          & 22.36 & 0.833 & 0.391 & 0.118 & 10.87 \\
RGB loss only    & 24.89 & 0.922 & 0.462 & 0.103 & 9.59 \\
\midrule
\textbf{PREEN (Ours)}         & \preenPSNR & \preenSSIM & \preenHCorr & \preenLPIPS & 9.07 \\
\bottomrule
\end{tabular}}
\end{table}

\begin{table}[tb]
\centering
\caption{\label{tab:teacher} Teacher ablation: PREEN trained under the published SA-LUT teacher vs.\ a \emph{global} variant (per-pixel context map collapsed to a per-image mean). Metrics are PREEN's.}
\small
\setlength{\tabcolsep}{3pt}
\resizebox{\linewidth}{!}{%
\begin{tabular}{@{}l c c c c c@{}}
\toprule
\textbf{Teacher} & \textbf{PSNR} & \textbf{SSIM} & \textbf{HCorr} & \textbf{LPIPS} & $\mathbf{\Delta E^{*}_{ab}}$ \\
\midrule
SA-LUT Global          & 24.74 & 0.917 & 0.465 & 0.094 & 9.23 \\
\midrule
\textbf{SA-LUT (Our setting)}  & \preenPSNR & \preenSSIM & \preenHCorr & \preenLPIPS & 9.07 \\
\bottomrule
\end{tabular}}
\end{table}

\begin{table}[tb]
\centering
\setlength{\tabcolsep}{4pt}
\caption{\label{tab:backbone} Phase~1 backbone ablation. Params = backbone + MLP head. Each backbone uses its native input resolution.}
\small
\resizebox{\linewidth}{!}{%
\begin{tabular}{@{}l r c c c c c@{}}
\toprule
\textbf{Backbone} & \textbf{Params} & \textbf{PSNR} & \textbf{SSIM} & \textbf{HCorr} & \textbf{LPIPS} & $\mathbf{\Delta E^{*}_{ab}}$ \\
\midrule
\textit{No backbone}                        &   6M & 23.03 & 0.883 & 0.359 & 0.141 & 11.36 \\
DINOv3~\cite{simeoni2025dinov3}             & 317M & 23.04 & 0.884 & 0.370 & 0.141 & 11.39 \\
SigLIP\,2~\cite{tschannen2025siglip2}       & 330M & 23.25 & 0.889 & 0.388 & 0.128 & 11.58 \\
InternViT~\cite{chen2024internvl}           & 318M & 23.40 & 0.906 & 0.371 & 0.127 & 11.22 \\
\midrule
\textbf{CLIP}~\cite{radford2021clip} & 315M & \phaseOnePSNR & \phaseOneSSIM & \phaseOneHCorr & \phaseOneLPIPS & 10.60 \\
\bottomrule
\end{tabular}}
\end{table}

\subsection{Main Results on PST50}

Table~\ref{tab:main} reports the five PST50 metrics (Section~\ref{sec:results:eval}) averaged over the paired triples. NCST~\cite{jiang2024ncst}, another parametric prior, is not benchmarked, as it releases no public code or weights and was not evaluated on PST50.

PREEN improves on Deep Preset in both its forms, the generator's rendered output $\hat{Y}$ (a raster) and the predicted preset $\hat{P}$ applied in Lightroom. In Table~\ref{tab:main}, it leads every baseline on PSNR, SSIM, and LPIPS, by $+1.34$\,dB PSNR over Deep Preset ($\hat{Y}$), $+1.20$\,dB over the next-best NLUT, and $+1.96$\,dB over the no-edit (identity) row, while lifting HCorr from $0.348$ (no-edit) to $\preenHCorr$, though it trails most free-form Raster/LUT baselines there. On $\Delta E^{*}_{ab}$ ($9.07$), it ties the LUT methods within per-image noise (supplementary). Section~\ref{sec:discussion} discusses how the restrained PST50 targets shape these scores.

\begin{figure*}[!t]
    \centering
    \setlength{\tabcolsep}{0pt}
    \renewcommand{\arraystretch}{0.45}
    \newcommand{\cellwidth}{0.163\linewidth}
    \begin{tabular}{@{}c@{\hspace{1.5pt}}c@{\hspace{5pt}}c@{\hspace{1.5pt}}c@{\hspace{1.5pt}}c@{\hspace{1.5pt}}c@{}}
      \small\bfseries $X$ & \small\bfseries $Y$ & \small\bfseries SA-LUT & \small\bfseries Deep Preset & \small\bfseries PREEN & \small\bfseries PREEN Vivid \\[0.5pt]
      \includegraphics[width=\cellwidth]{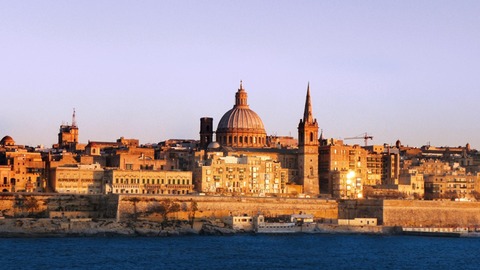} &
      \includegraphics[width=\cellwidth]{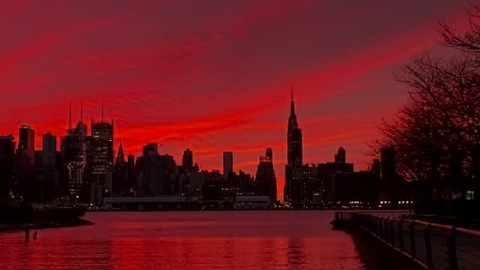} &
      \includegraphics[width=\cellwidth]{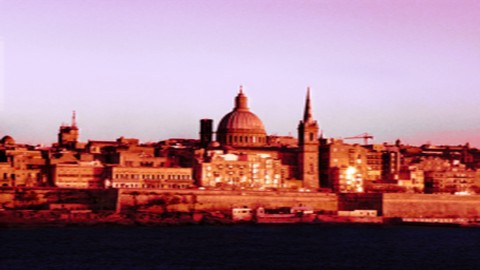} &
      \includegraphics[width=\cellwidth]{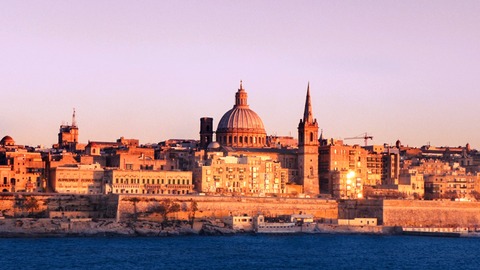} &
      \includegraphics[width=\cellwidth]{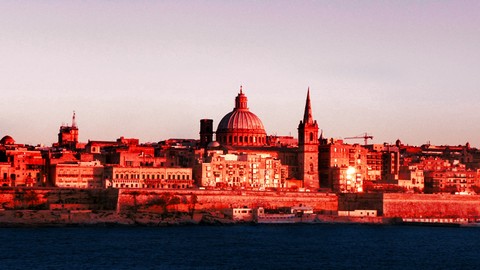} &
      \includegraphics[width=\cellwidth]{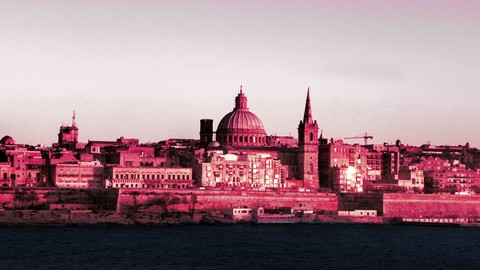} \\
      \includegraphics[width=\cellwidth]{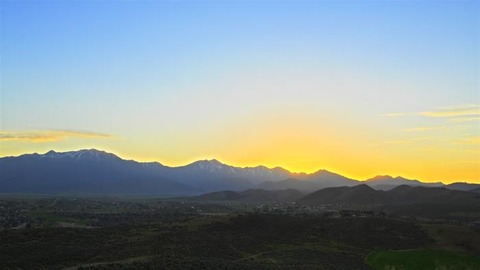} &
      \includegraphics[width=\cellwidth]{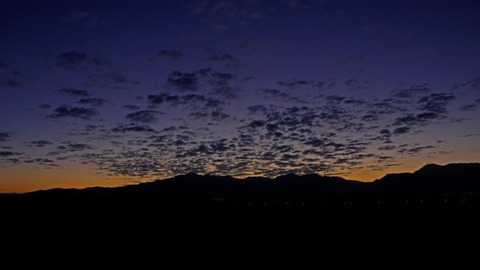} &
      \includegraphics[width=\cellwidth]{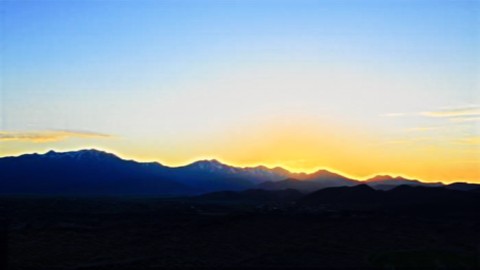} &
      \includegraphics[width=\cellwidth]{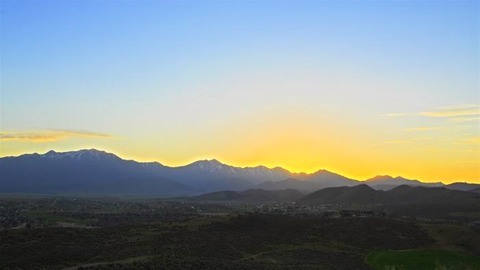} &
      \includegraphics[width=\cellwidth]{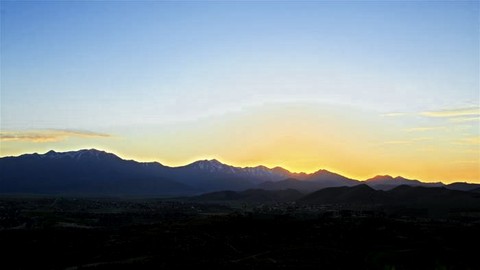} &
      \includegraphics[width=\cellwidth]{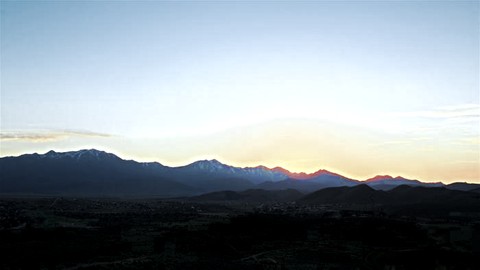} \\
      \includegraphics[width=\cellwidth]{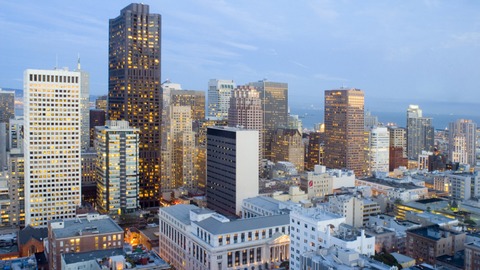} &
      \includegraphics[width=\cellwidth]{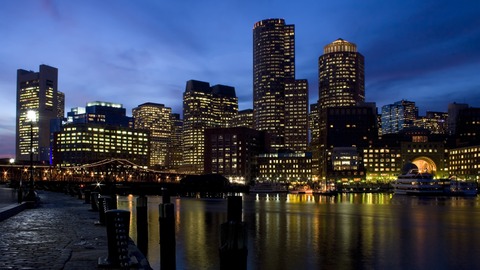} &
      \includegraphics[width=\cellwidth]{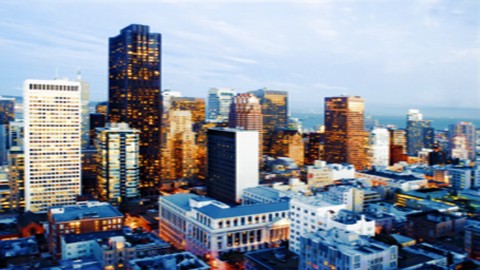} &
      \includegraphics[width=\cellwidth]{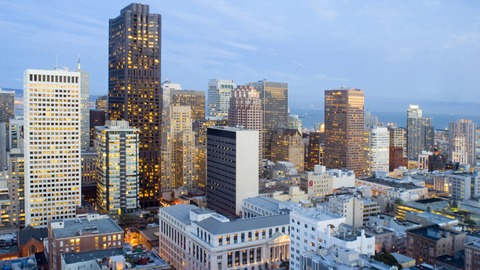} &
      \includegraphics[width=\cellwidth]{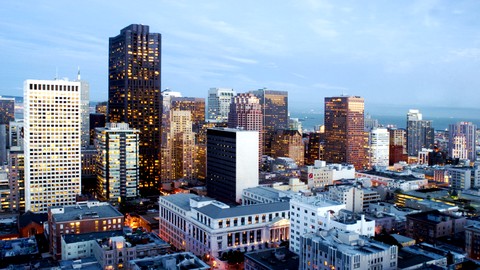} &
      \includegraphics[width=\cellwidth]{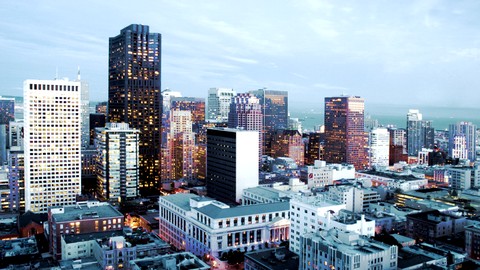} \\
    \end{tabular}
    \caption{Qualitative comparison on three DPST~\cite{luan2017deepphoto} pairs with distinct content (no paired ground truth). Columns: $X$, $Y$, SA-LUT~\cite{gong2025salut}, Deep Preset ($\hat{P}$)~\cite{ho2020deeppreset}, PREEN ($\alpha\!=\!1$), and PREEN Vivid ($\alpha\!=\!5$).}
    \label{fig:dpst}
\end{figure*}

A global preset lacks the local adaptivity of spatially varying baselines such as SA-LUT. What Table~\ref{tab:main} shows is that, on this benchmark, the $29$-parameter representation is not itself the bottleneck. It stays within range of LUT grids over $1000\times$ larger in dimension ($\mathbb{R}^{3\times2\times17^3}$ vs.\ $\mathbb{R}^{29}$; Table~\ref{tab:main}). The HCorr gap to the LUTs is prediction headroom, not a subspace limit (oracle HCorr~$0.672$, supplementary). Figure~\ref{fig:qual} shows a head-to-head on the first five PST50 samples, where PREEN's edit lands close to the expert target.

PREEN also generalizes beyond PST50. Figure~\ref{fig:dpst} gives qualitative results on DPST~\cite{luan2017deepphoto} pairs (no paired ground truth, so no metrics). It restyles content on par with Deep Preset while retaining $\alpha$ control. The full PST50 grid, two more DPST pairs, the predicted presets for example pairs, and an in-the-wild gallery are in the supplementary.

\subsection{Ablation Studies}

\subsubsection{Component Ablation}
Table~\ref{tab:ablation} isolates the contribution of each training phase. Phase~1 alone, a frozen-CLIP MLP regressor trained on pseudo-labels, already reaches \phaseOnePSNR\,dB PSNR, above both Deep Preset forms. Phase~2 LoRA adaptation adds the largest gains, most prominently in chromaticity ($\phaseOneHCorr\!\to\!\preenHCorr$ HCorr). By contrast, \emph{Phase~2 only}, a single stage from random initialization with no Phase~1 that fully fine-tunes the MLP under the image-level loss and adapts CLIP via LoRA as in PREEN, does not surpass the full pipeline on the task's target metrics. Although it edges SSIM, it trails on HCorr ($0.453$ vs.\ $\preenHCorr$) and LPIPS and reaches only a comparable $\Delta E^{*}_{ab}$ ($9.12$ vs.\ $9.07$) before peaking early and degrading, despite its larger fully fine-tuned capacity. The benefit of Phase~1's initialization is thus not recovered by single-stage optimization; replacing that stage's fine-tuning with LoRA fares worse still ($\Delta E^{*}_{ab}\,9.57$; supplementary), consistent with a low-rank update on an untrained base being underexpressive and with Phase~1's pretrained base enabling the efficient LoRA adaptation.

Single-factor variants locate the gains: MLP-only LoRA (frozen CLIP) keeps fidelity but recovers under half the chromaticity gain (HCorr $0.423$), so adaptation must reach the representation, not just the head; disabling the dual-trunk gradient routing (joint $L$/$ab$ gradients) costs $0.38$\,dB PSNR and $0.016$ SSIM at unchanged HCorr, confirming the routing aids tonal fidelity; and dropping identity samples collapses fidelity (SSIM $0.833$, PSNR below the no-edit floor). Removing the CIELAB terms (plain RGB MSE) instead gains pixel fidelity but cedes every color-facing metric (Table~\ref{tab:ablation}). A single RGB term underweights chroma, settling into damped color edits, whereas CIELAB lets us weight chroma above luminance ($w_{\mathrm{ab}}\!=\!5$ vs.\ $w_L\!=\!1$, Eq.~\ref{eq:phase2_loss}), committing to the color transfer the task demands.

\subsubsection{Teacher Choice}
SA-LUT varies its edit across the image, whereas our editor applies one global preset, so Phase~1 must project spatially varying edits onto global parameters. To test whether a teacher matched to the editor yields cleaner pseudo-labels, we also tried a \emph{global} variant that collapses SA-LUT's per-pixel context map to a per-image mean. It brings no consistent improvement (Table~\ref{tab:teacher}). The variant edges PSNR by ${\sim}0.1$\,dB and SSIM by $0.003$ but lowers HCorr ($\preenHCorr\!\to\!0.465$) and worsens $\Delta E^{*}_{ab}$ ($9.07\!\to\!9.23$), the two metrics that score color most directly. The two teachers differ significantly on no metric (supplementary), so the published teacher can be used as is, without tailoring it to the editor.

\subsubsection{Backbone Choice}
We probe the backbone by swapping the frozen CLIP ViT-L/14 encoder for three recent vision encoders at matched capacity and training budget (Table~\ref{tab:backbone}): the self-supervised DINOv3~\cite{simeoni2025dinov3}, the contrastive SigLIP\,2~\cite{tschannen2025siglip2}, and the VLM-oriented InternViT~\cite{chen2024internvl,gao2024miniinternvl} (checkpoints in the supplementary). Removing the backbone entirely, leaving only the handcrafted color statistics, degrades every metric. The converse, CLIP without the handcrafted features, scores $23.15$\,dB (not shown), between no-backbone ($23.03$) and full ($\phaseOnePSNR$; Table~\ref{tab:backbone}), so the backbone and the color features contribute complementarily, neither subsuming the other. On PSNR, the backbones are close; on $\Delta E^{*}_{ab}$, only CLIP ($10.60$) and InternViT beat the no-backbone floor ($11.36$), with CLIP best. We adopt CLIP, the standard, widely deployed choice with mature on-device support.

\begin{table}[tb]
\centering
\caption{\label{tab:groups} Incremental preset-group contribution on PST50. The predicted preset is masked at inference to retain only the listed parameter groups, with the rest set to their neutral value.}
\small
\setlength{\tabcolsep}{3pt}
\resizebox{\linewidth}{!}{%
\begin{tabular}{@{}l c c c c c c@{}}
\toprule
\textbf{Active groups} & \textbf{\#} & \textbf{PSNR} & \textbf{SSIM} & \textbf{HCorr} & \textbf{LPIPS} & $\mathbf{\Delta E^{*}_{ab}}$ \\
\midrule
\textit{No edit}      & 0  & 22.67 & 0.906 & 0.348 & 0.127 & 12.32 \\
\,+\,G1 (WB)          & 2  & 21.99 & 0.898 & 0.403 & 0.102 & 11.71 \\
\,+\,G2 (global tone) & 7  & 24.21 & 0.913 & 0.451 & 0.099 & 10.08 \\
\,+\,G3 (saturation)  & 8  & 24.40 & 0.919 & 0.443 & 0.096 & 9.60 \\
\midrule
\textbf{\,+\,G4 (per-hue HSL)} & 29 & \preenPSNR & \preenSSIM & \preenHCorr & \preenLPIPS & 9.07 \\
\bottomrule
\end{tabular}}
\end{table}

\begin{figure*}[t]
  \centering
  \setlength{\tabcolsep}{0pt}
  \renewcommand{\arraystretch}{0.45}
  \newcommand{\cellwidth}{0.163\linewidth}
  \begin{tabular}{@{}c@{\hspace{1.2pt}}c@{\hspace{5pt}}c@{\hspace{1.2pt}}c@{\hspace{1.2pt}}c@{\hspace{1.2pt}}c@{}}
    \small\bfseries $X$ & \small\bfseries $Y$ & \small\bfseries $\alpha\!=\!1$ (Default) & \small\bfseries $\alpha\!=\!3$ & \small\bfseries $\alpha\!=\!5$ (Vivid) & \small\bfseries $\alpha\!=\!10$ \\[0.5pt]
      \includegraphics[width=\cellwidth]{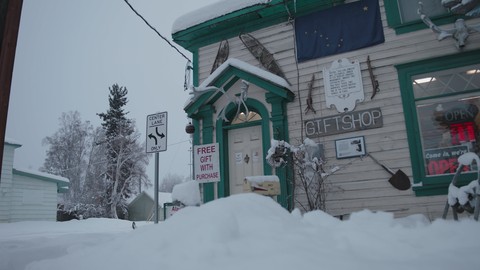} &
      \includegraphics[width=\cellwidth]{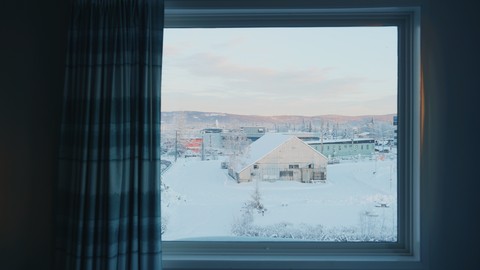} &
      \includegraphics[width=\cellwidth]{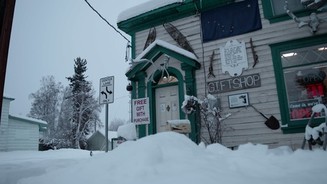} &
      \includegraphics[width=\cellwidth]{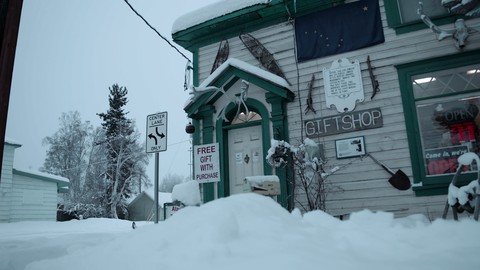} &
      \includegraphics[width=\cellwidth]{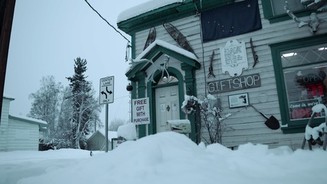} &
      \includegraphics[width=\cellwidth]{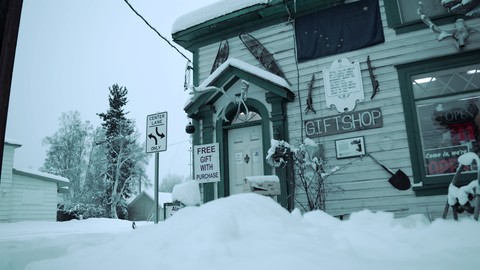} \\
      \includegraphics[width=\cellwidth]{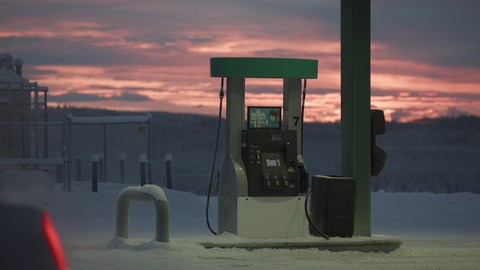} &
      \includegraphics[width=\cellwidth]{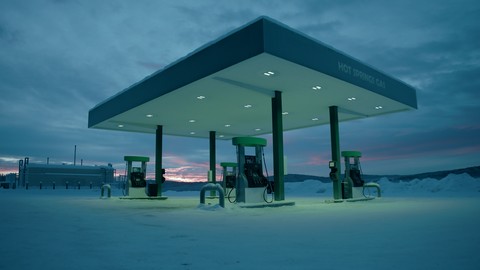} &
      \includegraphics[width=\cellwidth]{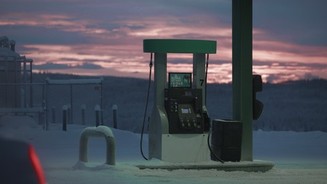} &
      \includegraphics[width=\cellwidth]{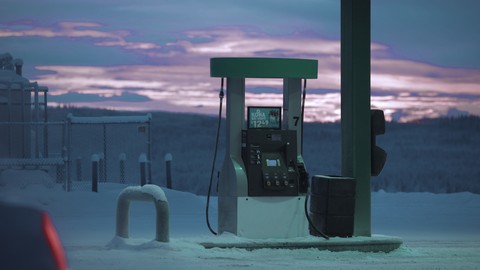} &
      \includegraphics[width=\cellwidth]{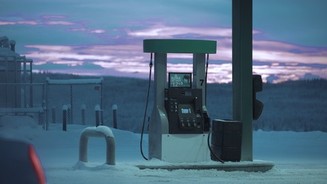} &
      \includegraphics[width=\cellwidth]{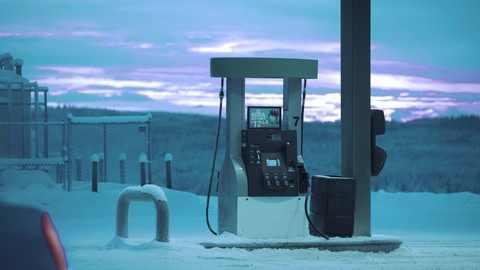} \\
      \includegraphics[width=\cellwidth]{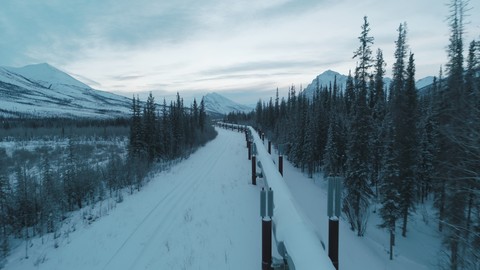} &
      \includegraphics[width=\cellwidth]{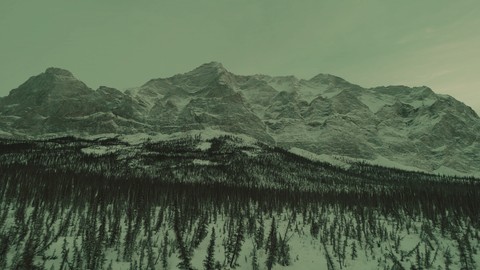} &
      \includegraphics[width=\cellwidth]{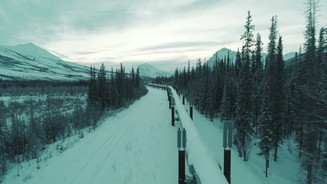} &
      \includegraphics[width=\cellwidth]{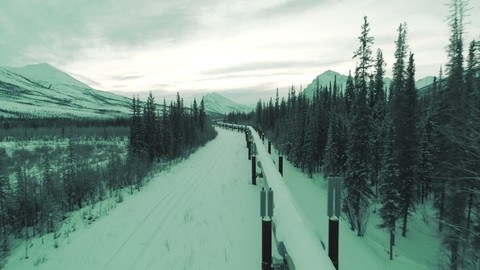} &
      \includegraphics[width=\cellwidth]{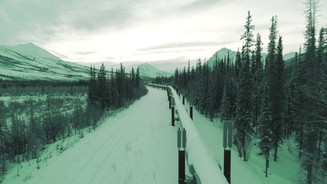} &
      \includegraphics[width=\cellwidth]{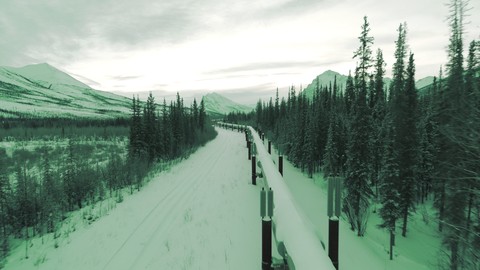} \\
  \end{tabular}
  \caption{Visual $\alpha$ sweep on the last three PST50 samples. One forward pass, with only the inference-time chroma scalar changing between operating points. Columns: $X$, $Y$, $\alpha\!=\!1$, $\alpha\!=\!3$, $\alpha\!=\!5$, $\alpha\!=\!10$.}
  \label{fig:alpha_sweep}
\end{figure*}

\subsubsection{Preset-Group Contribution}
To see which of the $29$ parameters carry the edit, we mask the predicted preset during inference to retain only the specified groups, setting the rest to their neutral values (Table~\ref{tab:groups}). The contribution is interpretable and largely monotonic. White balance alone (G1) sets the gross color cast (its PSNR sits below no-edit until tone joins), the global tone parameters (G2) add the largest single jump ($+2.2$\,dB over G1), global saturation (G3) helps fidelity while dipping HCorr ($0.451\!\to\!0.443$), and the $21$ per-hue HSL channels (G4) close the remaining HCorr gap ($0.443\!\to\!\preenHCorr$), the color-fidelity component the global parameters cannot reach.

\subsection{Strength Interpolation (\texorpdfstring{$\alpha$}{alpha} Sweep)}

Figure~\ref{fig:alpha_sweep} shows color saturation intensifying smoothly from $\alpha\!=\!1$ through $\alpha\!=\!5$ (Vivid) to $\alpha\!=\!10$ with no re-inference. PSNR peaks at the default and degrades monotonically as $\alpha$ grows while perceptual saturation keeps increasing, a trade-off that motivates our two operating points. We fix Vivid at $\alpha\!=\!5$ by visual inspection, not by tuning on the benchmark. The look is clearly intensified but not yet degenerate. The full per-$\alpha$ table and a luminance-rescale failure case are in the supplementary.

\subsection{Off-the-Shelf Foundation Models}
\label{sec:results:foundation}

\subsubsection{Direct Parameter Prediction}
We prompt three frontier vision-language models (Gemini~3.1~Pro~\cite{gemini31pro}, Claude~Opus~4.8~\cite{anthropic2025claude4}, and the open-weight Qwen3-VL~235B~\cite{qwenteam2025qwen3vl}) to predict the $29$ parameters from the content--reference pair, providing the full channel specification, valid ranges, and a forced structured-output schema, with one query per image (Table~\ref{tab:foundation}; verbatim prompt in the supplementary). Under a neutral prompt (ranges and sign conventions only), all three produce much larger edits and score worse on every metric of the supplementary comparison (PSNR $-2.9$ to $-6.0$\,dB). The VLM rows of Table~\ref{tab:foundation} therefore use the conservative guided prompt. They cluster at $21.4$--$22.5$\,dB PSNR, \mbox{$2.1$--$3.2$\,dB} below PREEN. Calibrating sliders from text alone, without the editor's feedback, is hard.

\begin{table}[tb]
\centering
\setlength{\tabcolsep}{4pt}
\caption{\label{tab:foundation} Off-the-shelf foundation models on PST50: frontier VLMs predicting the preset and a generative image model, with PREEN for reference.}
\small
\begin{tabular}{l c c c c c}
\toprule
\textbf{Method} & \textbf{PSNR} & \textbf{SSIM} & \textbf{HCorr} & \textbf{LPIPS} & $\mathbf{\Delta E^{*}_{ab}}$ \\
\midrule
\multicolumn{6}{@{}l}{\emph{Parameter prediction (Output: Preset)}}\\
Gemini 3.1 Pro        & 21.40 & 0.788 & 0.343 & 0.157 & 12.57 \\
Claude Opus 4.8       & 22.49 & 0.857 & 0.364 & 0.124 & 11.41 \\
Qwen3-VL 235B         & 21.87 & 0.886 & 0.352 & 0.124 & 12.02 \\
\midrule
\multicolumn{6}{@{}l}{\emph{Image-to-image (Output: Raster)}}\\
Nano Banana Pro       & 19.66 & 0.745 & 0.461 & 0.198 & 15.24 \\
\midrule
\textbf{PREEN (Ours)}        & \preenPSNR & \preenSSIM & \preenHCorr & \preenLPIPS & 9.07 \\
\bottomrule
\end{tabular}
\end{table}

\subsubsection{Direct Pixel Synthesis}
The complementary route skips parameters entirely. Nano Banana Pro (Gemini~3~Pro~Image)~\cite{nanobananapro}, run once per pair on PST50 and prompted to transfer the reference's color and mood while preserving structure, matches PREEN on HCorr but trails it in fidelity by ${\sim}5$\,dB PSNR, $0.17$ SSIM, and over $2\times$ LPIPS (Table~\ref{tab:foundation}). Synthesized pixels drift from the content's fine structure, and the output is a non-deterministic raster with no editable parameters.

\subsection{User Study}

Automatic metrics capture fidelity, not perceived quality. We run pairwise preference studies: PREEN Vivid ($\alpha\!=\!5$) and the default each against (i)~Deep Preset ($\hat{P}$) and (ii)~Claude~Opus~4.8 predicting the same preset, plus Vivid vs.\ default (isolating $\alpha$) and both vs.\ the SA-LUT teacher. Raters judge style match (closer to the reference) and aesthetic appeal (more pleasing as a standalone photo). The VLM opponent uses its guided prompt (Section~\ref{sec:results:foundation}). Figure~\ref{fig:user} reports PREEN's win rates, and the supplementary lists each with its $95\%$ confidence interval.

On style match, Vivid wins against every compared preset opponent (Deep Preset $61.6\%$, Claude $65.7\%$, PREEN's own default $73.4\%$; pooled $66.9\%$, $[62.2,71.6]$) and roughly ties the teacher itself ($44.6\%$, $[36.2,53.3]$), to which the default loses ($28.3\%$). On aesthetic appeal, the ordering inverts. Vivid drops to $43.3\%$ pooled, $[38.9,47.9]$, with raters favoring the default and Claude as standalone images, yet both PREEN modes are preferred to the teacher's full-strength edit ($68.8\%$/$73.8\%$).

This three-way split is the study's central finding. The metrics rank the \emph{default} highest, human style-match ranks \emph{Vivid} highest, and standalone aesthetic preference returns to the \emph{default}. ``Matches the reference'' and ``looks best on its own'' are different objectives. The $\alpha$ knob spans them without retraining.

The default also holds its own. It beats Deep Preset on both axes ($61.8\%$ style, $[53.0,70.6]$; $67.5\%$ aesthetic, $[60.6,74.0]$) and roughly matches Claude~Opus~4.8 ($52.0\%$/$52.5\%$) while remaining deterministic and far cheaper.

\subsection{On-Device Deployment}
\label{sec:results:ondevice}

We export $f_\theta$ (LoRA merged) to TFLite with 8-bit dynamic-range quantization (INT8 weights, FP32 activations) and benchmark on a Snapdragon~8~Elite~Gen~5 smartphone (8 CPU threads, median of 20 warm runs). The preset is predicted end-to-end in $\latencyMs$\,ms per pair, CPU-only. Latency is dominated by the CLIP backbone ($304$M of $315$M parameters), yet within a sub-second UX budget. Quantization preserves accuracy. On PST50, the INT8 artifact scores $24.34$/$0.915$/$0.479$/$0.093$/$9.02$, within $0.3$\,dB PSNR of Table~\ref{tab:main} and matching it on the other four metrics.

\section{Discussion and Limitations}
\label{sec:discussion}

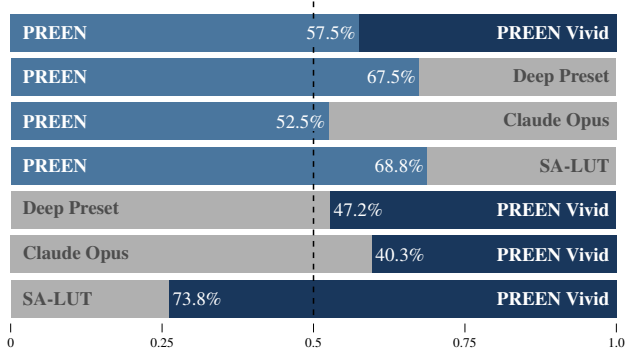
\begin{figure}[!tb]
\centering
\newcommand{\barlen}{11.2cm}            %
\newcommand{\barheight}{0.59}
\newcommand{\rowstep}{0.70}             %
\newcommand{\blockgap}{1.42}            %
\newcommand{\yscale}{1.17}
\newcommand{\labelfont}{\normalsize\bfseries}
\newcommand{\pctfont}{\normalsize}
\newcommand{\titlefont}{\Large\bfseries}
\newcommand{\titlegap}{0.5}
\edef\secondblock{\fpeval{-(6*\rowstep+\barheight+\blockgap)}}   %
\definecolor{vividblue}{RGB}{25,55,92}
\definecolor{defaultblue}{RGB}{74,118,158}
\definecolor{opponentgray}{RGB}{176,176,176}
\newcommand{\barDefaultVsOpponent}[6]{%
  \fill[#3] (0,#1) rectangle ({#2*\barlen},{#1-\barheight});
  \fill[opponentgray] ({#2*\barlen},#1) rectangle (\barlen,{#1-\barheight});
  \begin{pgfonlayer}{fg}\node[font=\labelfont,white,anchor=west] at (0.10,{#1-0.5*\barheight}) {#4};\end{pgfonlayer}
  \begin{pgfonlayer}{fg}\node[font=\labelfont,black!70,anchor=east] at ({\barlen-0.10},{#1-0.5*\barheight}) {#5};\end{pgfonlayer}
  \begin{pgfonlayer}{fg}\node[font=\pctfont,white,fill=#3,inner sep=1.5pt,anchor=east] at ({#2*\barlen-0.07},{#1-0.5*\barheight}) {#6};\end{pgfonlayer}}
\newcommand{\barDefaultVsVivid}[6]{%
  \fill[#3] (0,#1) rectangle ({#2*\barlen},{#1-\barheight});
  \fill[vividblue] ({#2*\barlen},#1) rectangle (\barlen,{#1-\barheight});
  \begin{pgfonlayer}{fg}\node[font=\labelfont,white,anchor=west] at (0.10,{#1-0.5*\barheight}) {#4};\end{pgfonlayer}
  \begin{pgfonlayer}{fg}\node[font=\labelfont,white,anchor=east] at ({\barlen-0.10},{#1-0.5*\barheight}) {#5};\end{pgfonlayer}
  \begin{pgfonlayer}{fg}\node[font=\pctfont,white,fill=#3,inner sep=1.5pt,anchor=east] at ({#2*\barlen-0.07},{#1-0.5*\barheight}) {#6};\end{pgfonlayer}}
\newcommand{\barOpponentVsVivid}[6]{%
  \fill[opponentgray] (0,#1) rectangle ({(1-#2)*\barlen},{#1-\barheight});
  \fill[#3] ({(1-#2)*\barlen},#1) rectangle (\barlen,{#1-\barheight});
  \begin{pgfonlayer}{fg}\node[font=\labelfont,black!70,anchor=west] at (0.10,{#1-0.5*\barheight}) {#4};\end{pgfonlayer}
  \begin{pgfonlayer}{fg}\node[font=\labelfont,white,anchor=east] at ({\barlen-0.10},{#1-0.5*\barheight}) {#5};\end{pgfonlayer}
  \begin{pgfonlayer}{fg}\node[font=\pctfont,white,fill=#3,inner sep=1.5pt,anchor=west] at ({(1-#2)*\barlen+0.07},{#1-0.5*\barheight}) {#6};\end{pgfonlayer}}
\resizebox{\columnwidth}{!}{%
\begin{tikzpicture}[y=\yscale cm]
\node[anchor=west,font=\titlefont] at (0,\titlegap) {Style match};
\barDefaultVsVivid{0}{0.266}{defaultblue}{PREEN}{PREEN Vivid}{26.6\%}
\barDefaultVsOpponent{-\rowstep}{0.618}{defaultblue}{PREEN}{Deep Preset}{61.8\%}
\barDefaultVsOpponent{-2*\rowstep}{0.520}{defaultblue}{PREEN}{Claude Opus}{52.0\%}
\barDefaultVsOpponent{-3*\rowstep}{0.283}{defaultblue}{PREEN}{SA-LUT}{28.3\%}
\barOpponentVsVivid{-4*\rowstep}{0.616}{vividblue}{Deep Preset}{PREEN Vivid}{61.6\%}
\barOpponentVsVivid{-5*\rowstep}{0.657}{vividblue}{Claude Opus}{PREEN Vivid}{65.7\%}
\barOpponentVsVivid{-6*\rowstep}{0.446}{vividblue}{SA-LUT}{PREEN Vivid}{44.6\%}
\draw[black,thick,dashed] ({0.5*\barlen},0.20) -- ({0.5*\barlen},{-6*\rowstep-\barheight-0.05});
\node[anchor=west,font=\titlefont] at (0,{\secondblock+\titlegap}) {Aesthetic appeal};
\barDefaultVsVivid{\secondblock}{0.575}{defaultblue}{PREEN}{PREEN Vivid}{57.5\%}
\barDefaultVsOpponent{\secondblock-\rowstep}{0.675}{defaultblue}{PREEN}{Deep Preset}{67.5\%}
\barDefaultVsOpponent{\secondblock-2*\rowstep}{0.525}{defaultblue}{PREEN}{Claude Opus}{52.5\%}
\barDefaultVsOpponent{\secondblock-3*\rowstep}{0.688}{defaultblue}{PREEN}{SA-LUT}{68.8\%}
\barOpponentVsVivid{\secondblock-4*\rowstep}{0.472}{vividblue}{Deep Preset}{PREEN Vivid}{47.2\%}
\barOpponentVsVivid{\secondblock-5*\rowstep}{0.403}{vividblue}{Claude Opus}{PREEN Vivid}{40.3\%}
\barOpponentVsVivid{\secondblock-6*\rowstep}{0.738}{vividblue}{SA-LUT}{PREEN Vivid}{73.8\%}
\draw[black,thick,dashed] ({0.5*\barlen},{\secondblock+0.20}) -- ({0.5*\barlen},{\secondblock-6*\rowstep-\barheight-0.05});
\foreach \t in {0,0.25,0.5,0.75,1.0}{
  \draw ({\t*\barlen},{\secondblock-6*\rowstep-\barheight-0.10}) -- ({\t*\barlen},{\secondblock-6*\rowstep-\barheight-0.24});
  \node[font=\scriptsize,below] at ({\t*\barlen},{\secondblock-6*\rowstep-\barheight-0.22}) {\t};}
\end{tikzpicture}}
\setlength{\abovecaptionskip}{0.3pt}
\caption{\label{fig:user} Human pairwise preference on PST50. Each bar splits the judgments between the two labeled methods. The printed percentage is the win rate of the segment it sits in, and the dashed line marks $50\%$.}
\end{figure}

More than the numbers, PREEN's payoff is the form of its output, an interpretable, deterministic preset that an editor consumes directly. The explicit $29$-dimensional subspace also makes controllability free: a single inference-time scalar $\alpha$ rescales the preset's chroma along a default-to-vivid trajectory of valid, re-editable presets. Across that trajectory, the three-way divergence that our user study surfaces (Figure~\ref{fig:user}) makes the best operating point a user preference, not a fixed target.

Beyond controllability, the fidelity numbers invite a second reading. A global preset is structurally more constrained than the spatial LUT teacher, so PREEN's higher PST50 fidelity scores (Table~\ref{tab:main}) do not indicate greater expressivity; they reflect PST50's restrained expert targets. The teacher overshoots these conservative edits, while PREEN's identity sampling (Table~\ref{tab:ablation}) and $\ell_2$ preset regularizer keep it calibrated. Damping the teacher toward the input confirms this, improving it on four of the five PST50 metrics (full diagnostic in the supplementary).

PREEN also inherits the editor's failure modes. Looks that \emph{remap} color instead of adjusting it (split-toned monochromes, wholesale palette shifts) lie outside $E$'s space, as do local effects such as grain, vignettes, and local burns; for the former, even the per-image oracle fails (supplementary). It also assumes a semantically compatible reference: our pairs are matched by content, and we do not test transfer to disjoint references. Shrinking the CLIP-dominated footprint (Section~\ref{sec:results:ondevice}) with a lighter encoder (e.g., MobileCLIP or FastViT) is left to future work.

\section{Conclusion}

We presented PREEN, which recasts exemplar-based color style transfer as regression onto a deterministic editor's $29$ parameters. Lacking ground-truth presets, we invert a style-transfer teacher's output into the editor's parameter space, train a CLIP-conditioned regressor, and adapt it with LoRA under an image-level CIELAB loss. On PST50, the preset recovers the reference's color shift while preserving structure; it is interpretable, ${\sim}1000\times$ more compact than the teacher's LUT, and runs on-device in sub-second time. A single chromaticity scalar spans default and vivid looks, exposing strength as a free, re-editable control. A human study shows that the two ends of that range serve different goals, matching the reference and standing alone as a photo.

\PREENbeginsupp
\maketitle
\thispagestyle{empty}

\section{Editor Pipeline Details}
The editor applies the $29$ parameters in a fixed order typical of consumer smartphone photo editors. Each parameter maps to a specific operator and stage (Table~\ref{supp:tab:editor}). Parameters act in $[-1,1]$ and are internally rescaled to the device ranges. Stages~1--4 operate in RGB, stages~5--8 on the BT.601 (Rec.\,601) luma $Y$, and stage~9 in HSL with per-hue Gaussian masks. Every operation is differentiable almost everywhere, so gradients from the Phase~2 image-level loss flow back to $P_{\mathrm{pred}}$.

Table~\ref{supp:tab:editor} gives the update rule of every stage. The input pixel is $X=(X_r,X_g,X_b)\in[0,1]^3$, and we write $s\equiv100\,p$ for the (rescaled) parameter of each stage.

\begin{table*}[t]
\centering
\footnotesize
\renewcommand{\arraystretch}{1.55}
\setlength{\tabcolsep}{6pt}
\caption{The $29$-parameter editor in fixed application order, with the per-stage update rule ($s\equiv100\,p$ for the stage's parameter; stage-1 multipliers $m_c$ in Table~\ref{supp:tab:wb}; $h{=}128/255$; constants rounded and hard branches shown for readability). Stages 1--4 act in RGB, 5--8 in YCbCr luma $Y$, 9 in HSL.}
\label{supp:tab:editor}
\begin{tabular}{@{}c l c l@{}}
\toprule
\textbf{Stage} & \textbf{Parameter} & \textbf{Idx} & \textbf{Update rule} \\
\midrule
1 & Temperature & 0 & $X_c \leftarrow m_c\,X_c$\quad ($m_c$: white-balance LUT, Table~\ref{supp:tab:wb}) \\
2 & Contrast    & 1 & $X \leftarrow \tfrac12+\gamma\,(X-\tfrac12),\quad \gamma=\big((s+400)/400\big)^2$ \\
\addlinespace[5pt]
3 & Tint        & 2 & $X_r\mathrel{+}=\rho\,\phi\cdot(0.0032\,s)\,X_r,\ \ X_g\mathrel{+}=(1{-}\rho)\,\phi\cdot(0.0034\,(-s))\,X_g,\ \ X_b\mathrel{+}=\rho\,\phi\cdot(0.0034\,s)\,X_b$ \\
  & & & \quad $\ell=0.2126X_r{+}0.7152X_g{+}0.0722X_b,\ \ \phi=(\ell/\sigma)^2$ if $\ell{<}\tau$ else $((1{-}\ell)/\sigma)^2$ \\
  & & & \quad $\rho=\mathrm{sig}(0.5s),\ \tau{=}\tfrac{128}{255},\ \sigma{=}\tfrac{170}{255}$ \\
\addlinespace[5pt]
4 & Saturation  & 3 & $X_c \leftarrow \big(W X_c + \mathrm{Gain}\,(g_rX_r{+}g_gX_g{+}g_bX_b)\big)/1024$ \\
  & & & \quad $W=\mathrm{clip}(0.1(103(s{+}100){-}10300){+}1024,\,0,\,2048),\ \ \mathrm{Gain}{=}1024{-}W,$ \\
  & & & \quad $(g_r,g_g,g_b){=}(316,624,84)/1024$ \\
\addlinespace[5pt]
5 & Brightness  & 4 & $Y \leftarrow Y + 0.8\,(s/127)\cdot\big(Y\ \text{if}\ Y{<}h\ \text{else}\ 1{-}Y\big)$ \\
6 & Exposure    & 5 & $Y \leftarrow Y + s\,|s|/25500$ \\
7 & Highlights  & 6 & $Y \leftarrow Y + 0.0028\,s\,Y\,(255Y/200-\tfrac12)$ \\
\addlinespace[5pt]
8 & Shadows     & 7 & $s{<}0$: $Y\,{+}\,0.003(\tau_m{-}Y)s$ (gated to $Y{<}\tau_m$) \\
  & & & \quad $s{>}0$: blend $Y\,{+}\,0.28\,s/255$ ($Y{\le}h$) and $Y\,{+}\,0.0022(1{-}Y)s$;\ \ $\tau_m{=}\tfrac{159}{255}$ \\
\addlinespace[5pt]
9 & Per-hue HSL & 8--28 & $\{H,S,L\}\mathrel{+}=\textstyle\sum_{j=1}^{7} w^{(\cdot)}_j\,s^{(\cdot)}_j\,g_{(\cdot)},\ \ w^{(\cdot)}_j=e^{-d_j^2/2\sigma_{(\cdot)}^2},\ \ d_j=\mathrm{wrap}(H{-}H_j)\in[-\tfrac12,\tfrac12]$ \\
  & & & \quad $H_j{=}\{0,40,60,120,180,240,300\}/360$ \\
  & & & \quad $\sigma_{\rm h,s,l}{=}\{22.2,14.3,13.6\}/360,\ \ g_{\rm h,s,l}{=}\{0.45/360,\,0.58/100,\,0.21/100\}$ \\
  & & & \quad $s^{(\cdot)}_j{=}100\,p_{8+3(j-1)+\{0,1,2\}}$: each hue center $j$ carries its own $H/S/L$ parameter \\
\bottomrule
\end{tabular}
\end{table*}

\begin{table}[ht]
\centering
\caption{Stage-1 white-balance calibration LUT: per-channel multipliers $M_{c,j}$ at the 9 temperature knots (UI units, here mapped to $p_0\in[-1,1]$).}
\small
\setlength{\tabcolsep}{4.5pt}
\label{supp:tab:wb}
\resizebox{\columnwidth}{!}{%
\begin{tabular}{l ccccccccc}
\toprule
knot & $-100$ & $-75$ & $-50$ & $-25$ & $0$ & $25$ & $50$ & $75$ & $100$ \\
\midrule
$M_R$ & 1.00 & 1.00 & 1.00 & 1.00 & 1.00 & 1.06 & 1.13 & 1.20 & 1.25 \\
$M_G$ & 1.67 & 1.34 & 1.17 & 1.08 & 1.00 & 1.06 & 1.10 & 1.13 & 1.16 \\
$M_B$ & 4.71 & 2.02 & 1.40 & 1.16 & 1.00 & 1.00 & 1.00 & 1.00 & 1.00 \\
\bottomrule
\end{tabular}}
\end{table}

\section{Full \texorpdfstring{$\alpha$}{alpha} Sweep Analysis}

Table~\ref{supp:tab:alpha} reports the full chroma-only $\alpha$ sweep on PST50, extending the two operating points used in the main paper. Toward $\alpha\!\to\!0$ the chroma edit is under-applied and the color-facing metrics (HCorr, $\Delta E^{*}_{ab}$) weaken, and for large $\alpha$ ($5$, $10$) over-saturation degrades every metric. Performance stays close to the $\alpha\!=\!1$ values for small departures ($\alpha\!\in\![0.5,2]$) and falls off more clearly by $\alpha\!=\!3$ (PSNR $24.63\!\to\!23.53$, $\Delta E^{*}_{ab}$ $9.07\!\to\!10.29$). We adopt $\alpha\!=\!1$ as the default operating point.

\begin{table}[ht]
\centering
\caption{Full $\alpha$ sweep on PST50. The first row is the identity (all $29$ parameters neutral, the ``No edit (identity)'' row of Table~1 of the main paper), shown for reference; it is \emph{not} the $\alpha{=}0$ point of the main paper's Eq.~(3), which would zero only the chroma slice while retaining the predicted tone parameters.}
\small
\setlength{\tabcolsep}{3pt}
\label{supp:tab:alpha}
\begin{tabular}{l c c c c c}
\toprule
\textbf{$\alpha$} & \textbf{PSNR$\uparrow$} & \textbf{SSIM$\uparrow$} & \textbf{HCorr$\uparrow$} & \textbf{LPIPS$\downarrow$} & $\mathbf{\Delta E^{*}_{ab}}\downarrow$ \\
\midrule
No edit          & 22.67 & 0.906 & 0.348 & 0.127 & 12.32 \\
$0.25$           & 24.11 & 0.909 & 0.412 & 0.112 & 10.45 \\
$0.5$            & 24.40 & 0.912 & 0.437 & 0.102 & 9.71 \\
$1.0$ (default)  & \preenPSNR & \preenSSIM & \preenHCorr & \preenLPIPS & 9.07 \\
$2.0$            & 24.26 & 0.913 & 0.481 & 0.095 & 9.36 \\
$3.0$            & 23.53 & 0.908 & 0.445 & 0.105 & 10.29 \\
$5.0$ (Vivid)    & \vividPSNR & \vividSSIM & \vividHCorr & \vividLPIPS & 12.18 \\
$7.0$            & 21.41 & 0.889 & 0.372 & 0.147 & 13.72 \\
$10.0$           & 20.49 & 0.876 & 0.332 & 0.169 & 15.51 \\
\bottomrule
\end{tabular}
\end{table}

\begin{figure*}[t]
  \centering
  \setlength{\tabcolsep}{0pt}
  \newcommand{\cellwidth}{0.163\linewidth}
  {\small\bfseries Chroma-only $\alpha$-scaling (Ours)}\\[1pt]
  \begin{tabular}{@{}c@{\hspace{1.2pt}}c@{\hspace{5pt}}c@{\hspace{1.2pt}}c@{\hspace{1.2pt}}c@{\hspace{1.2pt}}c@{}}
    \small\bfseries $X$ & \small\bfseries $Y$ & \small\bfseries $\alpha\!=\!1$ & \small\bfseries $\alpha\!=\!3$ & \small\bfseries $\alpha\!=\!5$ & \small\bfseries $\alpha\!=\!10$ \\[1.5pt]
    \includegraphics[width=\cellwidth]{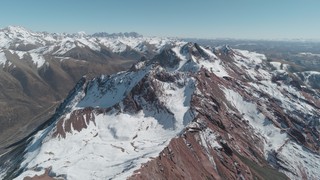} &
    \includegraphics[width=\cellwidth]{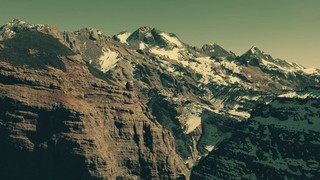} &
    \includegraphics[width=\cellwidth]{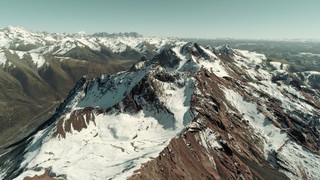} &
    \includegraphics[width=\cellwidth]{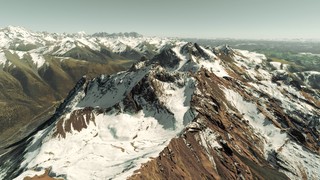} &
    \includegraphics[width=\cellwidth]{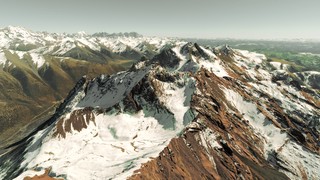} &
    \includegraphics[width=\cellwidth]{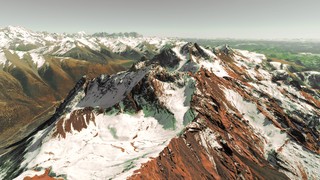} \\
  \end{tabular}\\[3pt]
  {\small\bfseries Full-parameter rescale (luminance + chroma)}\\[1pt]
  \begin{tabular}{@{}c@{\hspace{1.2pt}}c@{\hspace{5pt}}c@{\hspace{1.2pt}}c@{\hspace{1.2pt}}c@{\hspace{1.2pt}}c@{}}
    \includegraphics[width=\cellwidth]{figures/luma/luma_x} &
    \includegraphics[width=\cellwidth]{figures/luma/luma_y} &
    \includegraphics[width=\cellwidth]{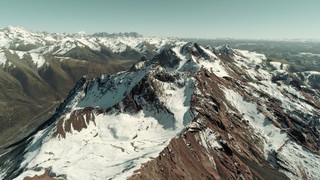} &
    \includegraphics[width=\cellwidth]{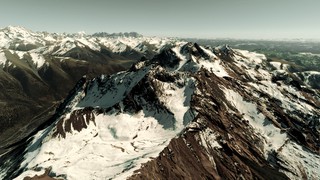} &
    \includegraphics[width=\cellwidth]{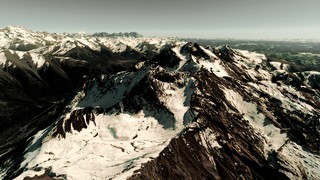} &
    \includegraphics[width=\cellwidth]{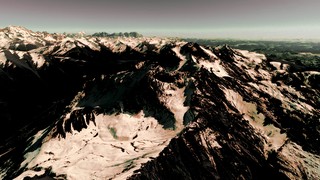} \\
  \end{tabular}
  \caption{Chroma-only versus full-parameter $\alpha$-scaling on one PST50 sample with a pronounced warm edit. Top: scaling only the chromaticity slice (ours) intensifies color while preserving tone. Bottom: scaling all parameters co-saturates the tonal channels, degrading the photograph as $\alpha$ grows. Columns: content $X$, reference $Y$, and $\alpha\!=\!1$, $3$, $5$, $10$.}
  \label{supp:fig:luma}
\end{figure*}

\noindent\textbf{Why chroma-only? Luminance rescale co-saturates tone.} Figure~\ref{supp:fig:luma} contrasts our chroma-only $\alpha$-scaling with rescaling all $29$ parameters (luminance included) on a PST50 sample with a pronounced warm edit. Chroma-only scaling intensifies the color identity while the tonal anchor stays fixed, so the image keeps a similar luminance scale even at $\alpha\!=\!10$. Rescaling the luminance parameters also co-saturates exposure, brightness, and contrast, driving the tones toward crushed shadows or blown highlights and flattening the photograph as $\alpha$ grows. This motivates restricting the $\alpha$ knob to the chromaticity slice $\mathcal{I}_{\mathrm{ab}}$.

\noindent\textbf{Does $\alpha$ rotate hue?} A natural worry is that scaling the chromaticity parameters reorients hue rather than merely intensifying it. We bound the per-channel hue rotation the $\alpha$ knob actually applies (an upper bound, as the per-hue Gaussian weight $w_h\!\le\!1$), measured over the $7$ per-hue channels of all $50$ PST50 images. At the default $\alpha\!=\!1$ no per-hue channel rotates by more than $30^\circ$ (median $2.1^\circ$), and even at the Vivid setting ($\alpha\!=\!5$) only $0.3\%$ do (median $8.9^\circ$). Pushing to $\alpha\!=\!10$, twice Vivid, still leaves the median at $14.9^\circ$ with no channel beyond $60^\circ$ and only $3.7\%$ past $30^\circ$. The editor's bounded $[-1,1]$ per-hue range thus keeps hue identity largely intact as saturation grows.

\section{Metric Protocol}

Every method receives PST50's Rec.\,709 content, and outputs and the expert target are bilinearly resized to $256{\times}256$ (the SA-LUT~\cite{gong2025salut} training resolution and kernel). PSNR is computed in sRGB. SSIM~\cite{wang2004ssim} uses an $11{\times}11$ Gaussian window ($\sigma{=}1.5$). HCorr is the mean per-channel correlation of $64$-bin HSV histograms. LPIPS~\cite{zhang2018lpips} uses pyiqa's \texttt{lpips-vgg+}, the VGG variant whose spatial average is weighted by semantic features. $\Delta E^{*}_{ab}$ is the mean per-pixel CIE76 difference $\lVert \mathrm{LAB}(X')-\mathrm{LAB}(\mathrm{target})\rVert_2$ in CIELAB. Times in Table~1 of the main paper are per-image latencies on an Intel Xeon 6975P-C CPU, with every method producing the same $256{\times}256$ output. The final checkpoint is chosen on a held-out validation split. PST50 is used only for the reported evaluation, and none of its images appear in our training data. We compute a perceptual hash (DCT pHash) for every training image and every PST50 image, and, for each PST50 image, rank every training image by pHash Hamming distance and inspect its closest matches. All are distinct photographs sharing at most coarse composition or tone, with no duplicates.

\noindent\textbf{Baseline rendering.} The Deep Preset ($\hat{P}$)~\cite{ho2020deeppreset} baseline predicts a Lightroom develop preset, which we apply to each content image and export in Adobe Lightroom Classic 15.2.1, one fixed build used for both the metrics in Table~1 of the main paper and the qualitative figures. Reinhard~\cite{reinhard2001color} is implemented as per-channel mean and standard-deviation transfer in CIELAB, and all other methods are rendered with their own released code. Neural Preset~\cite{ke2023neuralpreset} is omitted because it releases neither model weights nor training code. It is also a neural color-mapping method rather than an editor-preset predictor.

\section{Per-Image Statistical Tests}

All methods are evaluated on the same 50 PST50 triples, so paired per-image tests apply. Table~\ref{supp:tab:stats} reports PREEN's mean per-image advantage on each metric (sign-preserving difference, PREEN minus baseline) with two-sided Wilcoxon signed-rank $p$-values. At the uncorrected $0.05$ level, PREEN's PSNR and SSIM advantages are significant against every baseline, and LPIPS against all but Deep Preset ($\hat{Y}$) ($p{=}.077$). On $\Delta E^{*}_{ab}$, PREEN's point-estimate lead over the two LUT methods ($9.07$ vs.\ $9.63$/$9.81$) is within per-image noise ($p{\ge}.42$), while all other $\Delta E^{*}_{ab}$ advantages are significant. The HCorr deficits to the free-form LUT/raster methods (negative entries) are prediction headroom rather than a subspace limit, since the oracle reaches HCorr $0.672$ (Table~\ref{supp:tab:tto}). In contrast, the HCorr gain over the no-edit baseline is significant ($p{<}.001$). Under per-metric Holm correction within the family of between-method comparisons (the eight upper-block rows), every PSNR advantage over a published method remains significant (adjusted $p{\le}.031$), while the marginal effects do not (SSIM vs.\ no edit/Deep Preset ($\hat{P}$); LPIPS vs.\ Deep Preset ($\hat{Y}$)/ModFlows; $\Delta E^{*}_{ab}$ vs.\ Deep Preset ($\hat{Y}$)/ModFlows). Finally, the last row is a separate within-method test. We detect no significant per-image difference between PREEN trained under the SA-LUT-global teacher variant and under the published teacher (all $p{\ge}.56$), and the small color-metric gaps that do appear lean slightly toward the published teacher (the direction shown in Table~3 of the main paper). The global variant thus brings no consistent gain.

For absolute scale, PREEN's own per-metric means carry $95\%$ bootstrap CIs ($10$k resamples, $n{=}50$): PSNR $24.63\,[23.28,25.92]$, SSIM $0.914\,[0.894,0.932]$, HCorr $0.473\,[0.413,0.533]$, LPIPS $0.094\,[0.081,0.107]$, and $\Delta E^{*}_{ab}\,9.07\,[7.83,10.41]$.

\begin{table*}[t]
\centering
\caption{Per-image paired comparison, PREEN ($\alpha\!=\!1$) minus baseline: mean difference with two-sided Wilcoxon $p$ in parentheses. Positive $\Delta$PSNR/$\Delta$SSIM/$\Delta$HCorr and negative $\Delta$LPIPS/$\Delta\Delta E^{*}_{ab}$ favor PREEN. The last row instead compares the two teacher variants (PREEN trained under each); it is global minus published, so a positive $\Delta E^{*}_{ab}$ favors the published teacher.}
\small
\setlength{\tabcolsep}{4pt}
\label{supp:tab:stats}
\begin{tabular}{l r r r r r}
\toprule
\textbf{vs.} & $\Delta$\textbf{PSNR} & $\Delta$\textbf{SSIM} & $\Delta$\textbf{HCorr} & $\Delta$\textbf{LPIPS} & $\Delta\Delta E^{*}_{ab}$ \\
\midrule
No edit            & $+1.96$ {\tiny($<$.001)} & $+.009$ {\tiny(.041)} & $+.125$ {\tiny($<$.001)} & $-.034$ {\tiny($<$.001)} & $-3.25$ {\tiny($<$.001)} \\
Reinhard           & $+4.50$ {\tiny($<$.001)} & $+.079$ {\tiny($<$.001)} & $+.026$ {\tiny(.58)} & $-.028$ {\tiny(.002)} & $-4.34$ {\tiny(.002)} \\
AdaIN              & $+5.57$ {\tiny($<$.001)} & $+.210$ {\tiny($<$.001)} & $-.031$ {\tiny(.19)} & $-.221$ {\tiny($<$.001)} & $-5.32$ {\tiny($<$.001)} \\
Deep Preset ($\hat{Y}$) & $+1.34$ {\tiny(.011)} & $+.022$ {\tiny(.007)} & $+.062$ {\tiny(.12)} & $-.014$ {\tiny(.077)} & $-1.46$ {\tiny(.014)} \\
ModFlows            & $+4.64$ {\tiny($<$.001)} & $+.062$ {\tiny($<$.001)} & $-.037$ {\tiny(.22)} & $-.025$ {\tiny(.041)} & $-3.70$ {\tiny(.021)} \\
NLUT               & $+1.20$ {\tiny(.015)} & $+.030$ {\tiny(.002)} & $-.072$ {\tiny(.028)} & $-.035$ {\tiny($<$.001)} & $-0.74$ {\tiny(.42)} \\
SA-LUT             & $+2.09$ {\tiny(.010)} & $+.030$ {\tiny(.015)} & $-.044$ {\tiny(.17)} & $-.040$ {\tiny(.002)} & $-0.56$ {\tiny(.45)} \\
Deep Preset ($\hat{P}$)   & $+1.92$ {\tiny($<$.001)} & $+.020$ {\tiny(.026)} & $+.088$ {\tiny(.001)} & $-.023$ {\tiny(.003)} & $-3.06$ {\tiny($<$.001)} \\
\midrule
SA-LUT-global vs.\ published teacher & $+0.11$ {\tiny(.61)} & $+.003$ {\tiny(.95)} & $-.008$ {\tiny(.81)} & $.000$ {\tiny(.87)} & $+0.17$ {\tiny(.56)} \\
\bottomrule
\end{tabular}
\end{table*}

\section{Training-Free Probes and Reference Points}

We chart what training-free use of $(X, Y)$ and the teacher affords, under the same protocol as Table~1 of the main paper. (i)~\emph{Fit-to-reference} optimizes, per pair, the $29$ editor parameters (AdamW, $1{,}200$ steps) so the edited content matches the reference's color statistics, using either per-channel CIELAB mean and standard deviation, in the spirit of classic color transfer~\cite{reinhard2001color}, or the full sorted CIELAB marginals, then applies the result at full resolution. (ii)~\emph{Damped teacher output} is the raster blend $X'_{\gamma}=(1{-}\gamma)X+\gamma\,T(X,Y)$ with the SA-LUT teacher, sweeping $\gamma\in[0.1,1.0]$ and reporting the best ($\gamma{=}0.5$). (iii)~\emph{Test-time-optimization (TTO) inversion} applies Eq.~(1) of the main paper at test time to the teacher output (or to the damped teacher output), i.e., a preset-output method that needs the teacher plus a per-image optimization at inference (about one minute per image on the same CPU at $256{\times}256$, $1{,}000$ AdamW steps, vs.\ PREEN's $0.36$\,s without the teacher). (iv)~\emph{Oracle} is the same inversion against the expert target, upper-bounding the editor's subspace.

Table~\ref{supp:tab:tto} reports the results, which carry four lessons.

\emph{First}, reference-fitting reaches HCorr $0.455$/$0.471$ under the two statistic-matching objectives (on par with PREEN's $0.473$) yet collapses every fidelity metric below the no-edit baseline. The reference is a different scene, so its raw statistics are not the editing target, and over-applying them distorts the photograph (Figure~\ref{supp:fig:tto}, where PREEN's restrained edits on the same samples stay close to the target). This is the characteristic failure of a distribution-only proxy and shows why HCorr must be read alongside fidelity.

\emph{Second}, the damped teacher output shows how PST50's restrained targets reward restraint. At $\gamma{=}0.5$, the blend surpasses most methods in Table~1 of the main paper. Calibrating other baselines' edit strength does not close the gap: the same $\gamma$ sweep lifts none of the other four learned raster/LUT baselines past PREEN's $24.63$\,dB (best blends: NLUT $24.05$, ModFlows $24.16$, Deep Preset ($\hat{Y}$) $23.52$, AdaIN $23.51$\,dB).

\emph{Third}, projecting teachers into the editor changes their PSNR only slightly (SA-LUT $22.54{\to}22.67$\,dB, damped $25.09{\to}25.05$\,dB), consistent again with the subspace not being the bottleneck, though each projection costs a per-image optimization on top of the teacher. Notably, PREEN lands closer to the expert targets than the projection of its own undamped supervision ($22.67{\to}24.63$\,dB, $\Delta E^{*}_{ab}$ $9.70{\to}9.07$). The direction of this change (fidelity up, HCorr $0.505{\to}0.473$) matches the training objective's built-in restraint, since the $\ell_2$ preset regularizer and identity sampling damp predicted edits. PREEN's presets are also roughly $4\times$ smaller in norm than the teacher's projection (median $\lVert P\rVert$ ratio $0.24$) and only weakly aligned with it in direction (median cosine $0.28$).

\emph{Fourth}, the oracle reaches $32.56$\,dB and HCorr $0.672$, far above every method on PSNR, HCorr, and $\Delta E^{*}_{ab}$ (and best on SSIM, if narrowly), though its LPIPS ($0.137$) is worse than even the no-edit baseline ($0.127$). The gap to PREEN on those four metrics is prediction error from observing only $(X,Y)$, which we leave to future work. (The oracle minimizes a CIELAB-weighted reconstruction objective, trading perceptual texture for color/tone fidelity, so it upper-bounds the editor's color/tone subspace, not its perceptual quality.)

\begin{table}[ht]
\centering
\caption{Training-free probes and reference points on PST50, evaluated against the expert target. $\dagger$\,=\,requires the SA-LUT teacher at inference; $\ddagger$\,=\,per-image test-time optimization. The bottom row is the oracle fit to the target (upper bound on the editor's subspace).}
\small
\setlength{\tabcolsep}{4pt}
\label{supp:tab:tto}
\resizebox{\columnwidth}{!}{%
\begin{tabular}{l c c c c c}
\toprule
\textbf{Method} & \textbf{PSNR} & \textbf{SSIM} & \textbf{HCorr} & \textbf{LPIPS} & $\mathbf{\Delta E^{*}_{ab}}$ \\
\midrule
No edit                 & 22.67 & 0.906 & 0.348 & 0.127 & 12.32 \\
Fit-to-ref (mean+std)$^{\ddagger}$   & 19.84 & 0.812 & 0.455 & 0.191 & 13.86 \\
Fit-to-ref (marginals)$^{\ddagger}$  & 19.59 & 0.793 & 0.471 & 0.214 & 14.13 \\
Damped teacher output ($\gamma{=}0.5$, raster)$^{\dagger}$ & 25.09 & 0.922 & 0.501 & 0.088 & 8.71 \\
TTO inversion of teacher$^{\dagger\ddagger}$        & 22.67 & 0.892 & 0.505 & 0.120 & 9.70 \\
TTO inversion of damped teacher output$^{\dagger\ddagger}$ & 25.05 & 0.922 & 0.471 & 0.091 & 8.93 \\
PREEN ($\alpha\!=\!1$)  & \preenPSNR & \preenSSIM & \preenHCorr & \preenLPIPS & 9.07 \\
\midrule
\textit{Oracle (29-param fit to GT)} & \textit{32.56} & \textit{0.937} & \textit{0.672} & \textit{0.137} & \textit{4.90} \\
\bottomrule
\end{tabular}}
\end{table}

\begin{figure*}[t]
  \centering
  \setlength{\tabcolsep}{0pt}
  \newcommand{\cellwidth}{0.192\linewidth}
  \begin{tabular}{@{}c@{\hspace{1.5pt}}c@{\hspace{5pt}}c@{\hspace{1.5pt}}c@{\hspace{5pt}}c@{}}
    \small\bfseries Input & \small\bfseries Reference & \small\bfseries Fit-to-ref & \small\bfseries PREEN & \small\bfseries Target \\[1.5pt]
    \includegraphics[width=\cellwidth]{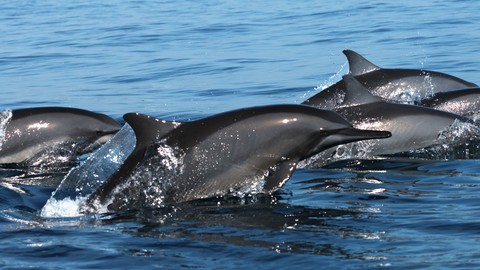} &
    \includegraphics[width=\cellwidth]{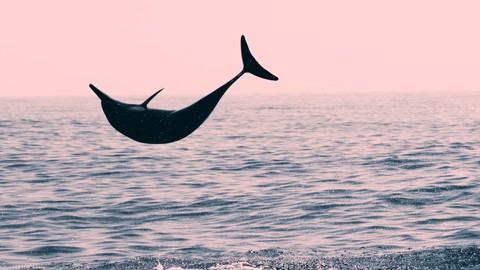} &
    \includegraphics[width=\cellwidth]{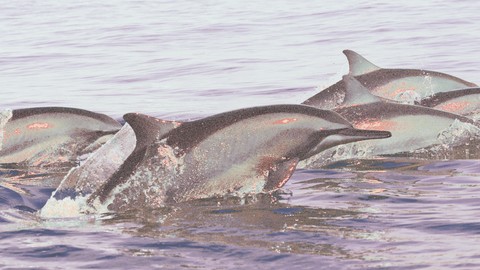} &
    \includegraphics[width=\cellwidth]{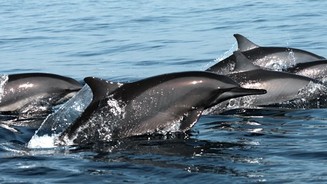} &
    \includegraphics[width=\cellwidth]{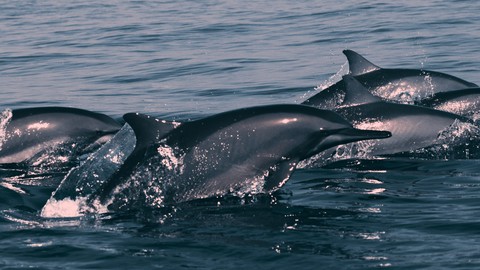} \\
    \includegraphics[width=\cellwidth]{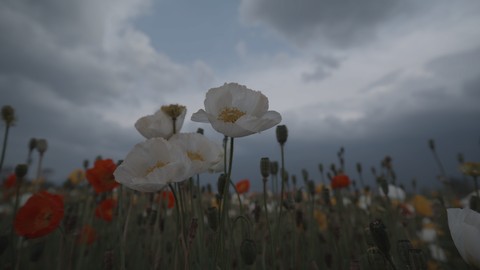} &
    \includegraphics[width=\cellwidth]{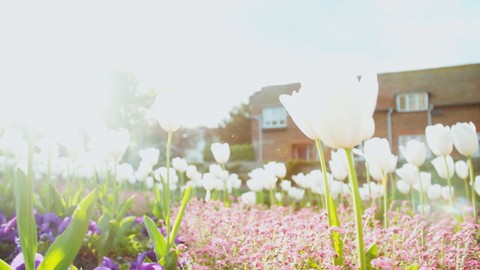} &
    \includegraphics[width=\cellwidth]{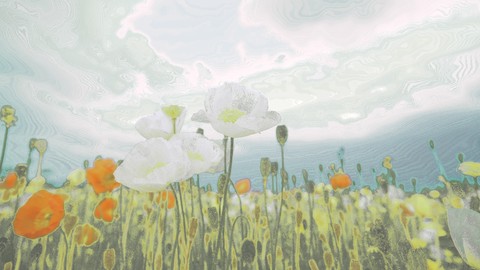} &
    \includegraphics[width=\cellwidth]{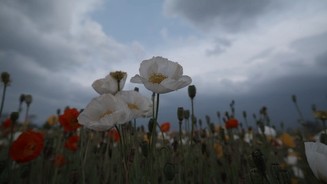} &
    \includegraphics[width=\cellwidth]{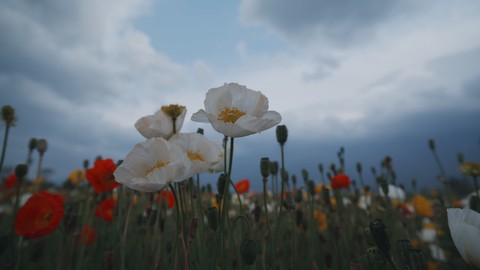} \\
    \includegraphics[width=\cellwidth]{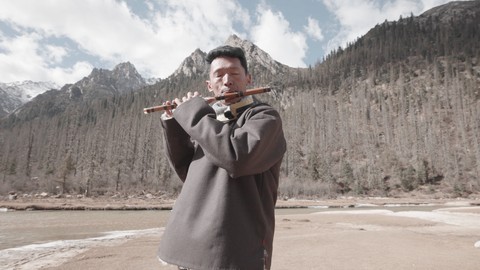} &
    \includegraphics[width=\cellwidth]{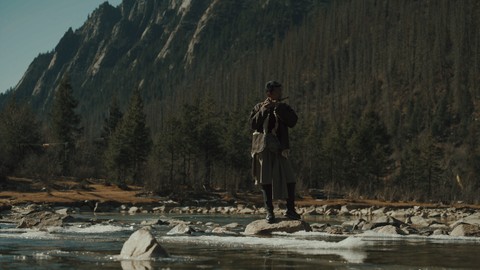} &
    \includegraphics[width=\cellwidth]{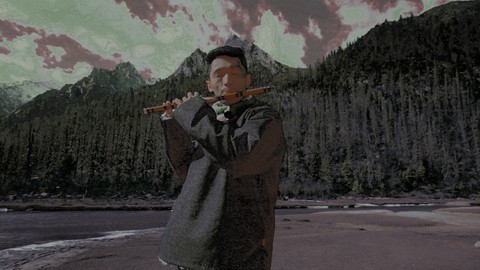} &
    \includegraphics[width=\cellwidth]{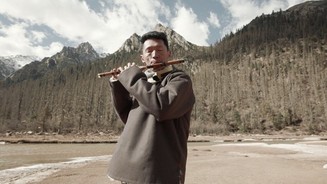} &
    \includegraphics[width=\cellwidth]{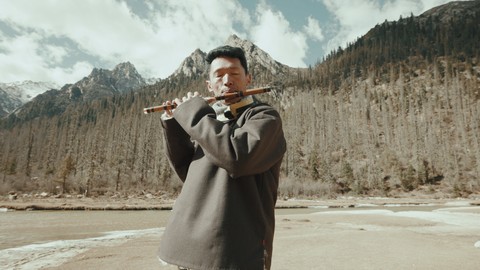} \\
  \end{tabular}
  \caption{Training-free fit-to-reference (marginals) over-applies the reference look and distorts the photograph. The three lowest-PSNR fit-to-reference results on PST50. Columns: input $X$, reference $Y$, fit-to-reference, PREEN ($\alpha\!=\!1$), and the expert target.}
  \label{supp:fig:tto}
\end{figure*}

\section{Predicted Preset Breakdown}
\begin{figure*}[p]
\centering
\includegraphics[width=0.94\textwidth]{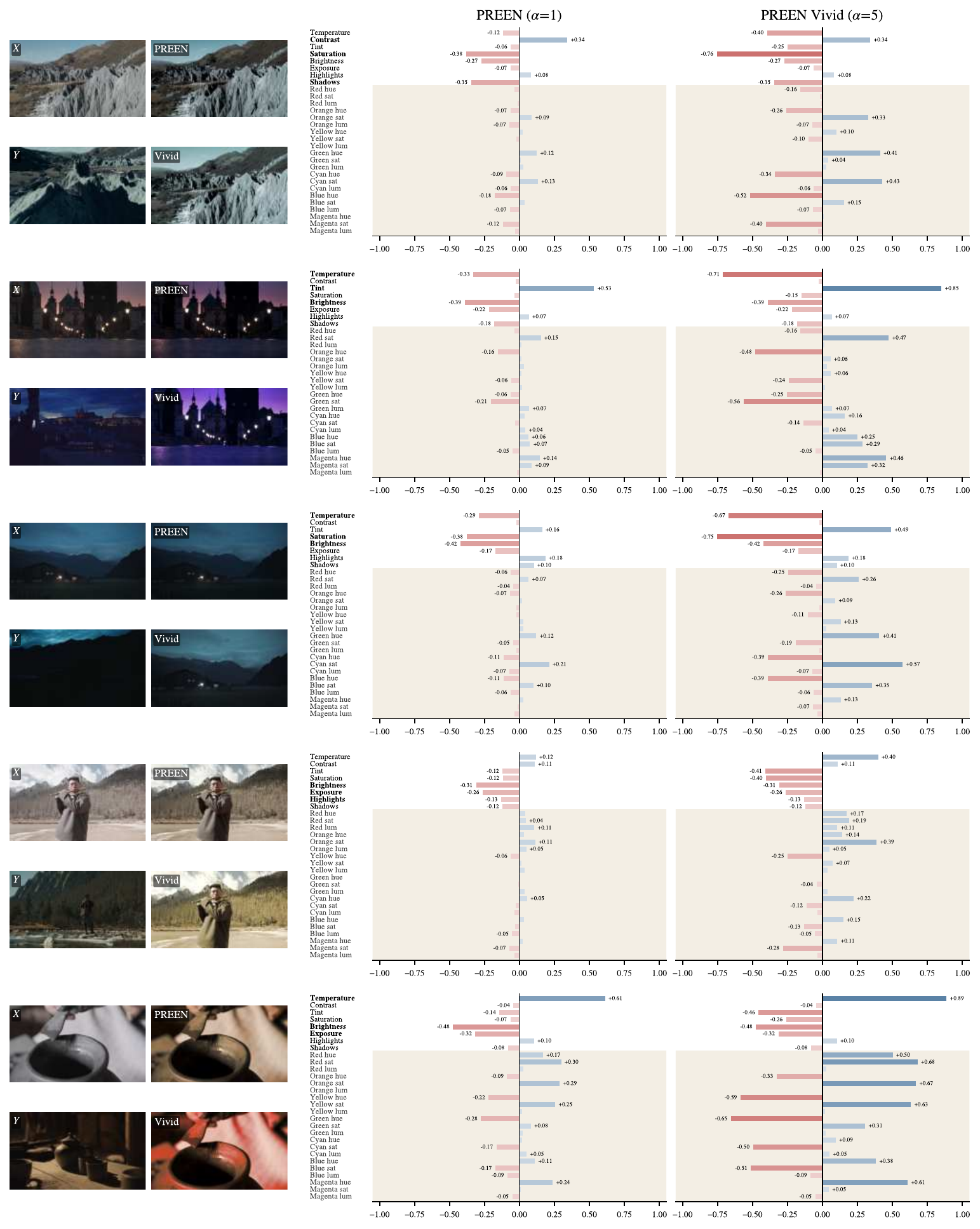}
\caption{\label{supp:fig:presets} Predicted presets on PST50 (first five pairs). Per row: inputs ($X$, $Y$), rendered outputs, and the predicted $29$-parameter preset for PREEN ($\alpha\!=\!1$) and PREEN Vivid ($\alpha\!=\!5$) as real slider values (top eight: global controls; shaded block: the $21$ per-hue HSL channels). Bar shade tracks magnitude, and the three largest-magnitude parameters (by PREEN $\alpha\!=\!1$ magnitude) are bold; Vivid scales chromaticity only, leaving tone bars unchanged.}
\end{figure*}

To make the output's interpretability concrete, Figure~\ref{supp:fig:presets} shows the actual $29$-parameter presets PREEN predicts for the first five PST50 pairs, namely the $8$ global controls and the $21$ per-hue HSL channels, as real editor-slider values in $[-1,1]$. The Vivid column ($\alpha\!=\!5$) rescales only the $17$ chromaticity parameters, leaving the $12$ tone parameters (the five global tone controls, i.e., contrast, brightness, exposure, highlights, and shadows, plus the seven per-hue luminance channels) unchanged. This is visible as identical tone bars, with the chroma bars (where nonzero) amplified between the two columns.

\section{Qualitative Gallery}

\begin{figure*}[t]
  \centering
  \setlength{\tabcolsep}{0pt}
  \newcommand{\cellwidth}{0.192\linewidth}
  \begin{tabular}{@{}c@{\hspace{1.5pt}}c@{\hspace{5pt}}c@{\hspace{1.5pt}}c@{\hspace{5pt}}c@{}}
    \small\bfseries Input & \small\bfseries Reference & \small\bfseries Oracle fit & \small\bfseries PREEN & \small\bfseries Target \\[1.5pt]
    \includegraphics[width=\cellwidth]{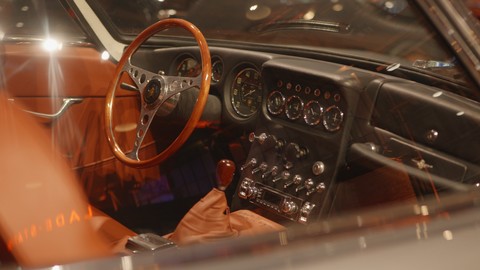} &
    \includegraphics[width=\cellwidth]{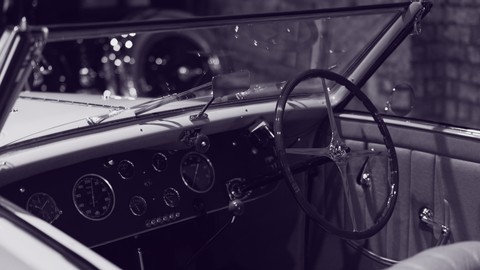} &
    \includegraphics[width=\cellwidth]{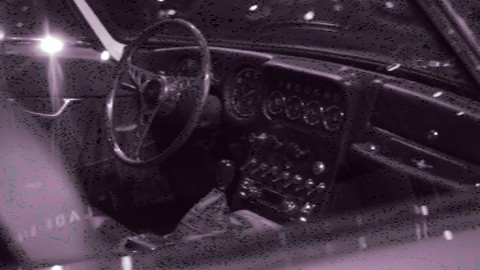} &
    \includegraphics[width=\cellwidth]{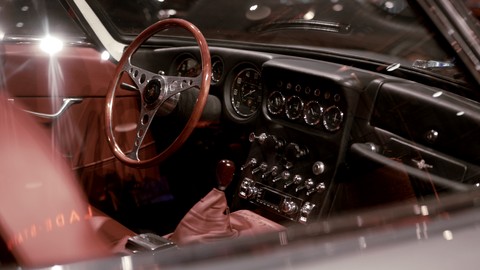} &
    \includegraphics[width=\cellwidth]{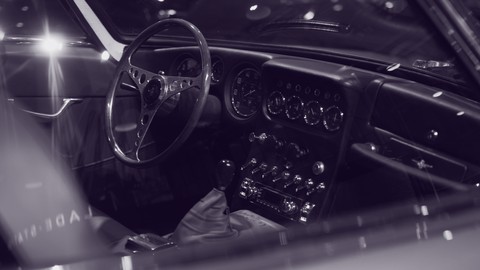} \\
    \includegraphics[width=\cellwidth]{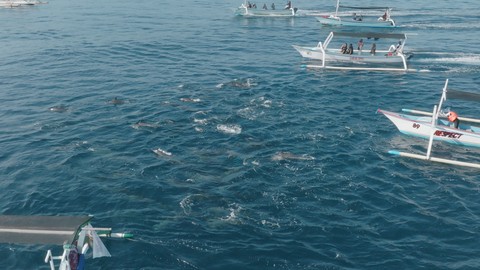} &
    \includegraphics[width=\cellwidth]{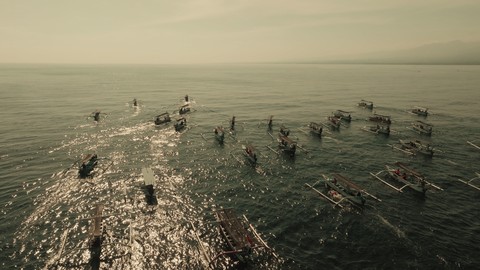} &
    \includegraphics[width=\cellwidth]{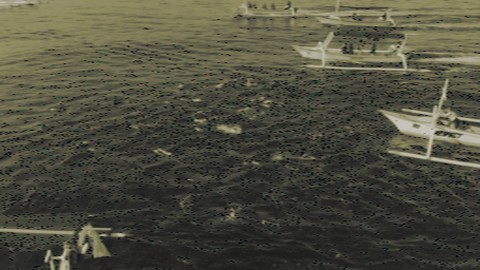} &
    \includegraphics[width=\cellwidth]{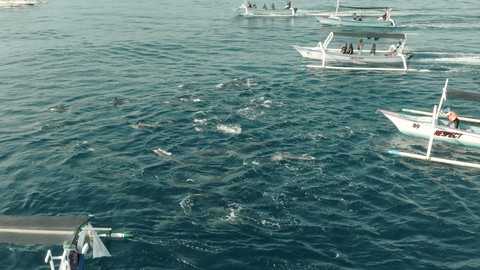} &
    \includegraphics[width=\cellwidth]{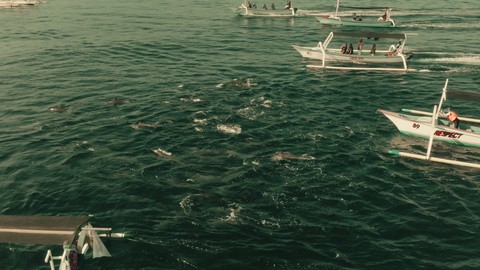} \\
    \includegraphics[width=\cellwidth]{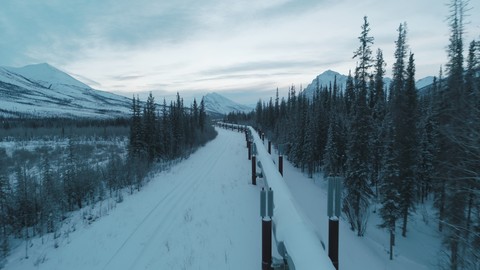} &
    \includegraphics[width=\cellwidth]{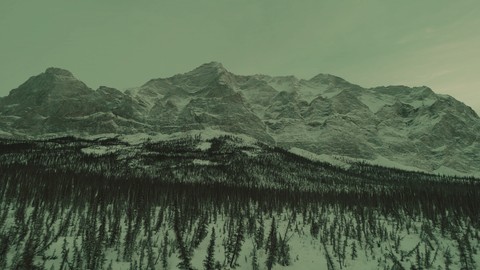} &
    \includegraphics[width=\cellwidth]{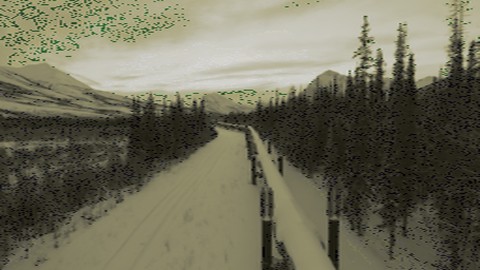} &
    \includegraphics[width=\cellwidth]{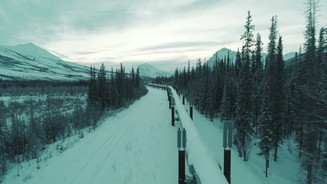} &
    \includegraphics[width=\cellwidth]{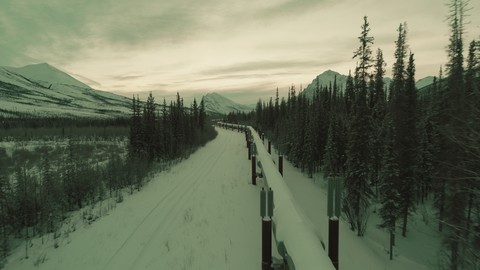} \\
    \includegraphics[width=\cellwidth]{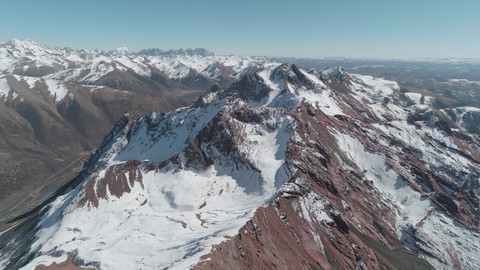} &
    \includegraphics[width=\cellwidth]{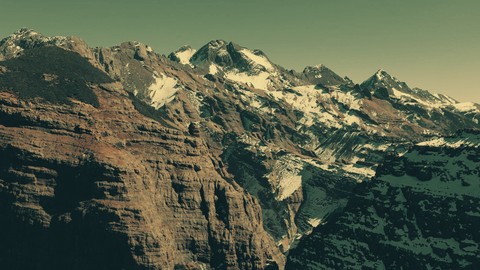} &
    \includegraphics[width=\cellwidth]{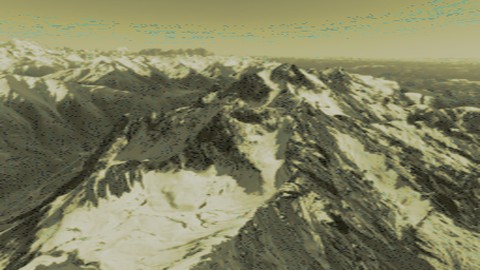} &
    \includegraphics[width=\cellwidth]{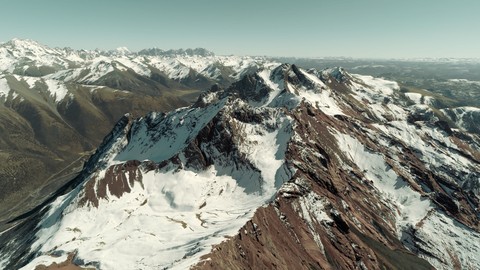} &
    \includegraphics[width=\cellwidth]{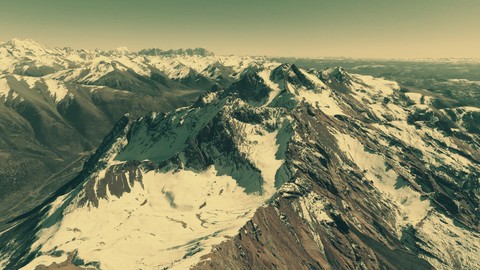} \\
    \includegraphics[width=\cellwidth]{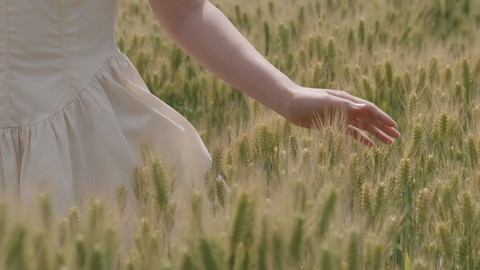} &
    \includegraphics[width=\cellwidth]{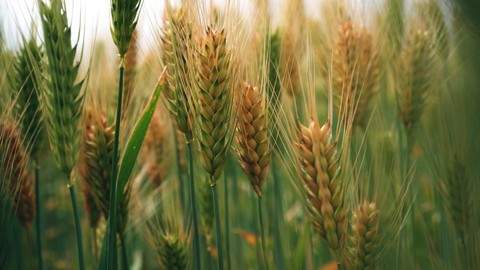} &
    \includegraphics[width=\cellwidth]{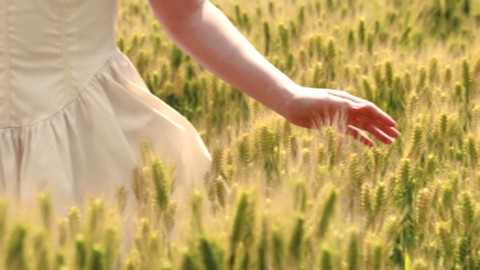} &
    \includegraphics[width=\cellwidth]{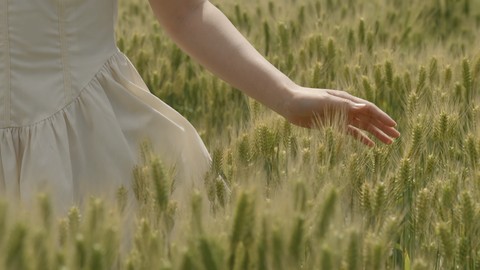} &
    \includegraphics[width=\cellwidth]{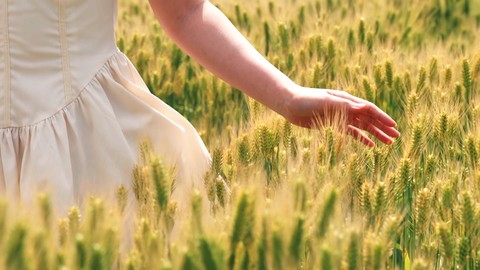} \\
  \end{tabular}
  \caption{PREEN's five worst per-image $\Delta E^{*}_{ab}$ cases on PST50: target looks outside the editor's render space (rows 1--4) and a reachable target the regressor misses (last row). Columns: input $X$, reference $Y$, oracle fit to the target, PREEN, and the expert target.}
  \label{supp:fig:failure}
\end{figure*}

\noindent\textbf{Failure cases.} Figure~\ref{supp:fig:failure} shows the failure family behind most of PREEN's worst PST50 cases, namely, target looks that the editor cannot render. We verify this with the per-image oracle fit, the Eq.~(1) inversion applied to the ground truth, shown in the third column of the figure. On these samples even the oracle misses the look (oracle $\Delta E^{*}_{ab}$ $8.7$--$11.0$ vs.\ an across-benchmark oracle median of $4.1$, mean $4.90$), because the target \emph{remaps} color (for example, a split-toned near-monochrome or a wholesale palette shift) rather than adjusting existing color. No global preset can express such remapping, regardless of regressor quality. Four of PREEN's five largest per-image $\Delta E^{*}_{ab}$ errors fall in this editor-limited family. The last row shows the complementary case, where the oracle matches the target closely (oracle $\Delta E^{*}_{ab}$ $4.6$), whereas PREEN misses it ($22.2$): plain prediction error, i.e., the headroom to the oracle that Section~4.3 of the main paper identifies.

\begin{figure*}[t]
    \centering
    \setlength{\tabcolsep}{0pt}
    \newcommand{\cellwidth}{0.196\linewidth}
    \begin{tabular}{@{}c@{\hspace{1.5pt}}c@{\hspace{5pt}}c@{\hspace{1.5pt}}c@{\hspace{1.5pt}}c@{}}
      \small\bfseries $X$ & \small\bfseries $Y$ & \small\bfseries Deep Preset ($\hat{P}$) & \small\bfseries PREEN & \small\bfseries PREEN Vivid \\[1.5pt]
      \includegraphics[width=\cellwidth]{figures/dpst/dpst_r0_x} & \includegraphics[width=\cellwidth]{figures/dpst/dpst_r0_y} & \includegraphics[width=\cellwidth]{figures/dpst/dpst_r0_dp} & \includegraphics[width=\cellwidth]{figures/dpst/dpst_r0_a1} & \includegraphics[width=\cellwidth]{figures/dpst/dpst_r0_vivid} \\
      \includegraphics[width=\cellwidth]{figures/dpst/dpst_r1_x} & \includegraphics[width=\cellwidth]{figures/dpst/dpst_r1_y} & \includegraphics[width=\cellwidth]{figures/dpst/dpst_r1_dp} & \includegraphics[width=\cellwidth]{figures/dpst/dpst_r1_a1} & \includegraphics[width=\cellwidth]{figures/dpst/dpst_r1_vivid} \\
      \includegraphics[width=\cellwidth]{figures/dpst/dpst_r2_x} & \includegraphics[width=\cellwidth]{figures/dpst/dpst_r2_y} & \includegraphics[width=\cellwidth]{figures/dpst/dpst_r2_dp} & \includegraphics[width=\cellwidth]{figures/dpst/dpst_r2_a1} & \includegraphics[width=\cellwidth]{figures/dpst/dpst_r2_vivid} \\
      \includegraphics[width=\cellwidth]{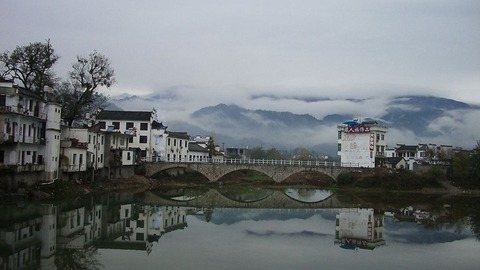} & \includegraphics[width=\cellwidth]{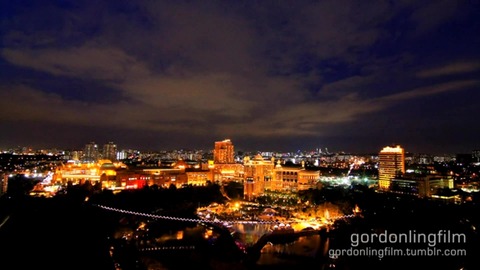} & \includegraphics[width=\cellwidth]{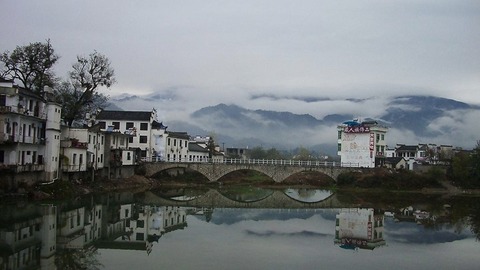} & \includegraphics[width=\cellwidth]{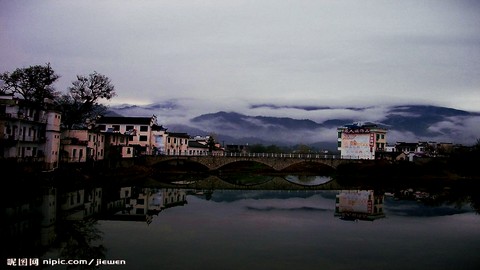} & \includegraphics[width=\cellwidth]{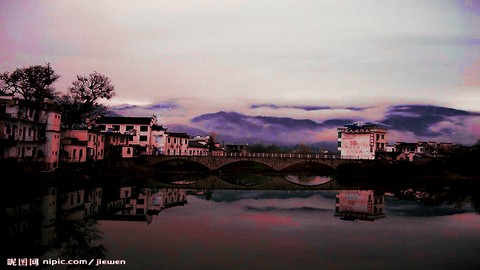} \\
      \includegraphics[width=\cellwidth]{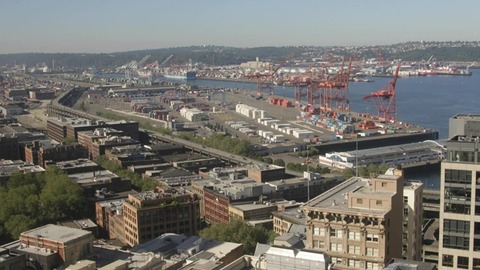} & \includegraphics[width=\cellwidth]{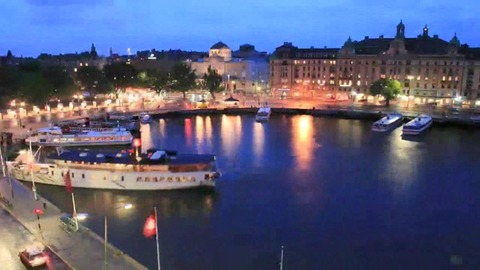} & \includegraphics[width=\cellwidth]{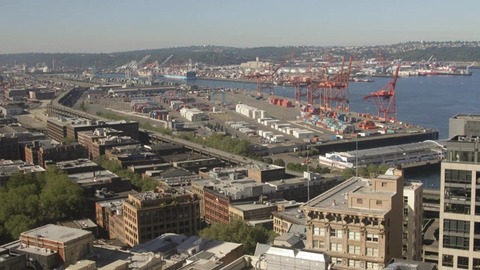} & \includegraphics[width=\cellwidth]{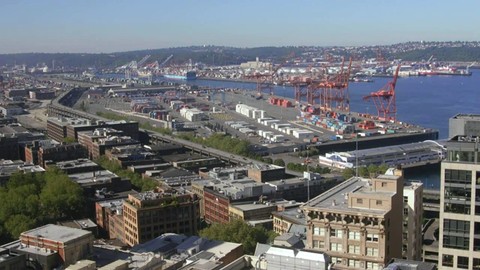} & \includegraphics[width=\cellwidth]{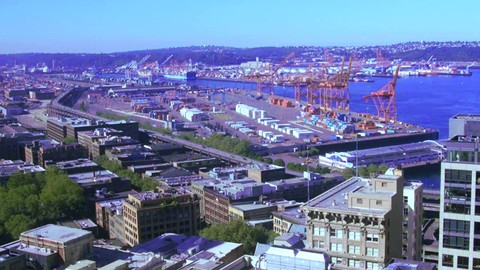} \\
    \end{tabular}
    \caption{Qualitative results on five DPST~\cite{luan2017deepphoto} pairs (distinct content; no paired ground truth). Columns: $X$, $Y$, Deep Preset ($\hat{P}$)~\cite{ho2020deeppreset}, PREEN ($\alpha\!=\!1$), and PREEN Vivid ($\alpha\!=\!5$).}
    \label{supp:fig:dpst}
\end{figure*}

\noindent\textbf{Generalization beyond PST50 (DPST).} Figure~\ref{supp:fig:dpst} shows five DPST~\cite{luan2017deepphoto} pairs with distinct content (the main paper showed the first three); there is no paired ground truth, so the comparison is qualitative. PREEN transfers the reference's color character to unseen content and retains the same $\alpha$ control.

\begin{figure*}[p]
  \centering
  \setlength{\tabcolsep}{0pt}
  \renewcommand{\arraystretch}{0}
  \newcommand{\cellwidth}{0.122\linewidth}
  \begin{tabular}{@{}c@{}c@{}c@{}c@{\hspace{4pt}}c@{}c@{}c@{}c@{}}
    \scriptsize\bfseries $X$ & \scriptsize\bfseries $Y$ & \scriptsize\bfseries PREEN & \scriptsize\bfseries PREEN Vivid & \scriptsize\bfseries $X$ & \scriptsize\bfseries $Y$ & \scriptsize\bfseries PREEN & \scriptsize\bfseries PREEN Vivid \\
    \includegraphics[width=\cellwidth]{figures/wild/wild00_x}&\includegraphics[width=\cellwidth]{figures/wild/wild00_y}&\includegraphics[width=\cellwidth]{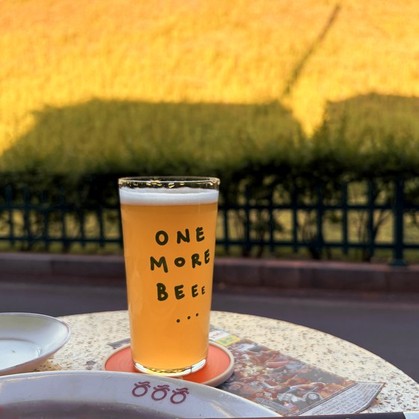}&\includegraphics[width=\cellwidth]{figures/wild/wild00_v}&\includegraphics[width=\cellwidth]{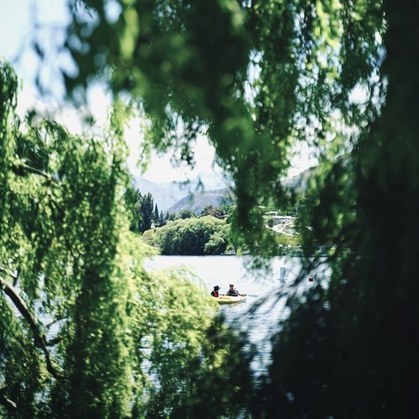}&\includegraphics[width=\cellwidth]{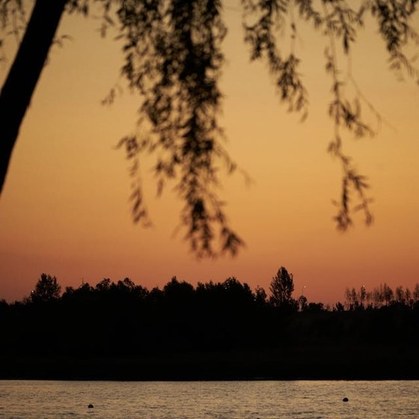}&\includegraphics[width=\cellwidth]{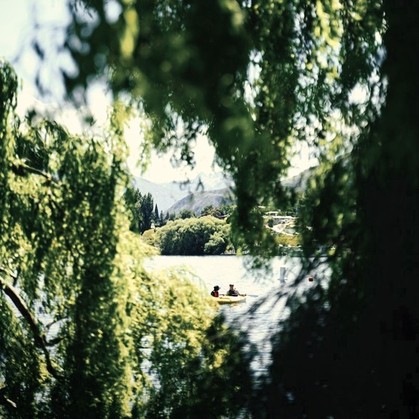}&\includegraphics[width=\cellwidth]{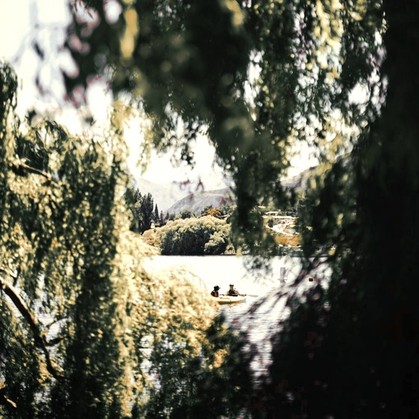} \\
    \includegraphics[width=\cellwidth]{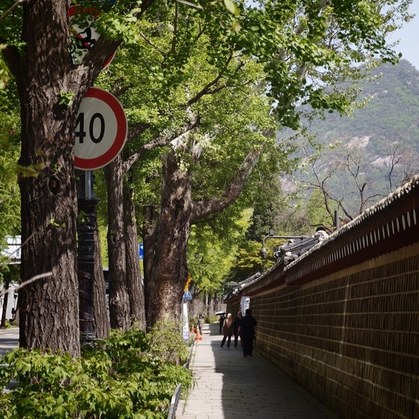}&\includegraphics[width=\cellwidth]{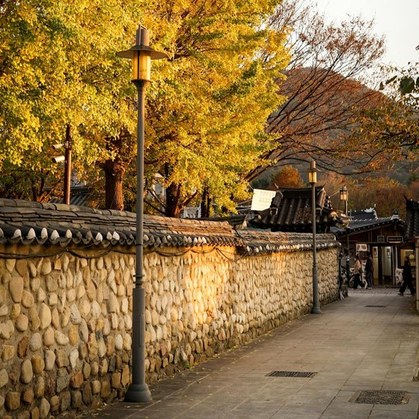}&\includegraphics[width=\cellwidth]{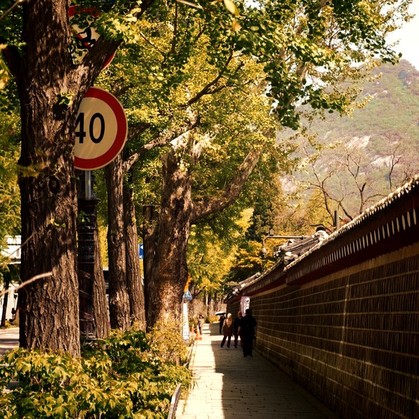}&\includegraphics[width=\cellwidth]{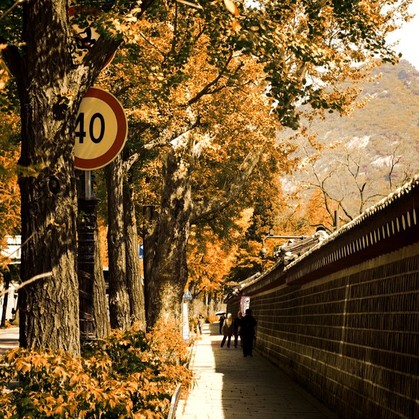}&\includegraphics[width=\cellwidth]{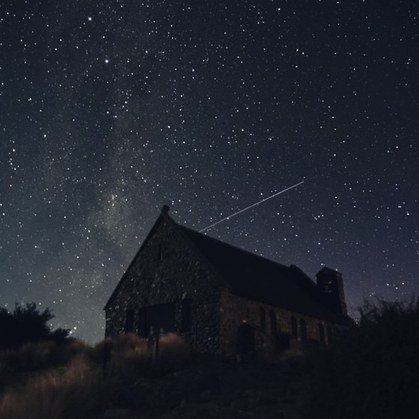}&\includegraphics[width=\cellwidth]{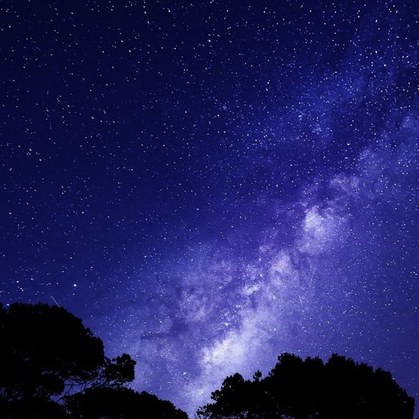}&\includegraphics[width=\cellwidth]{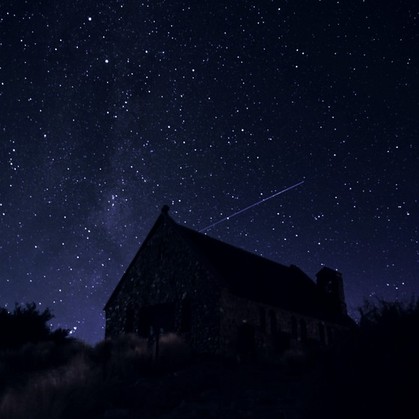}&\includegraphics[width=\cellwidth]{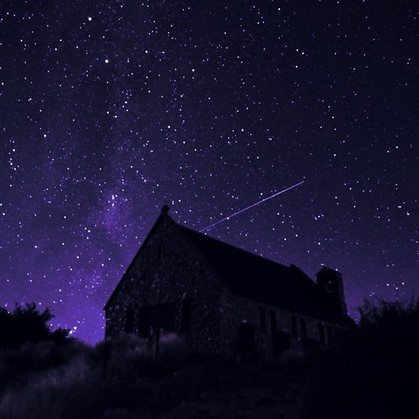} \\
    \includegraphics[width=\cellwidth]{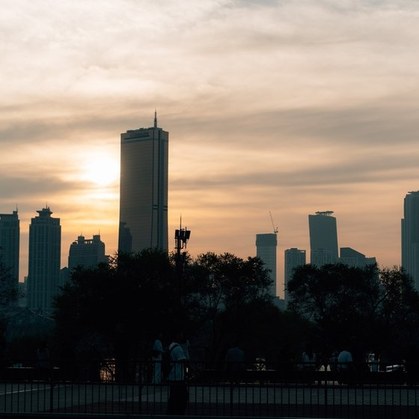}&\includegraphics[width=\cellwidth]{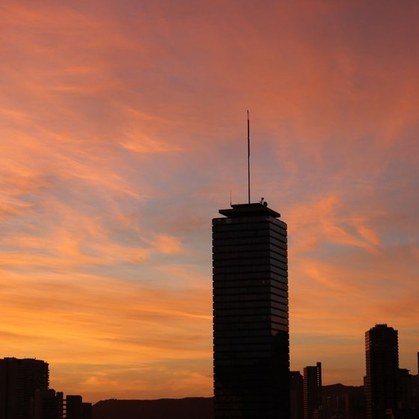}&\includegraphics[width=\cellwidth]{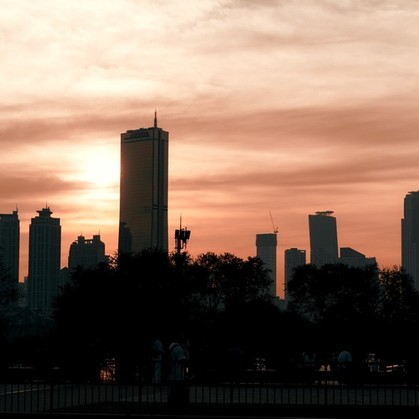}&\includegraphics[width=\cellwidth]{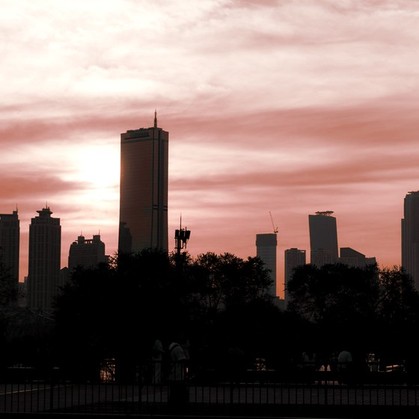}&\includegraphics[width=\cellwidth]{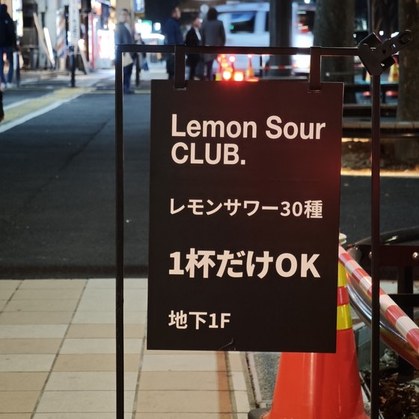}&\includegraphics[width=\cellwidth]{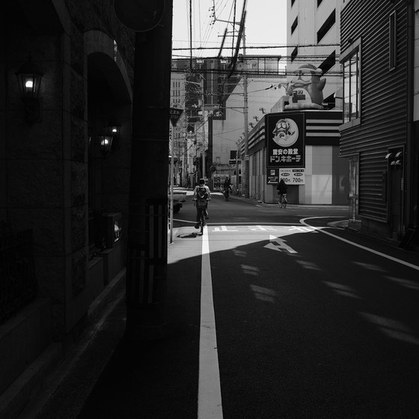}&\includegraphics[width=\cellwidth]{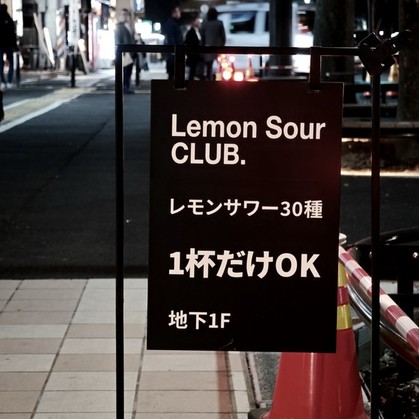}&\includegraphics[width=\cellwidth]{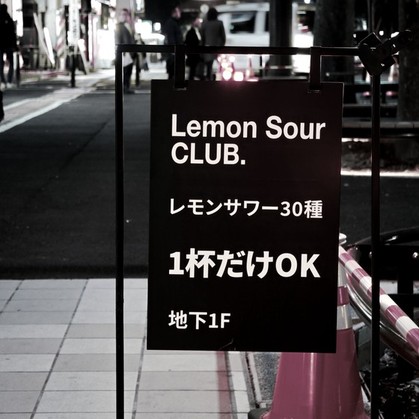} \\
    \includegraphics[width=\cellwidth]{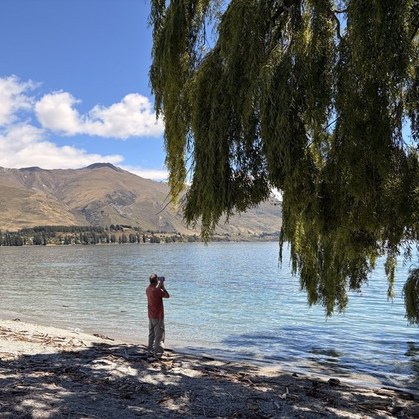}&\includegraphics[width=\cellwidth]{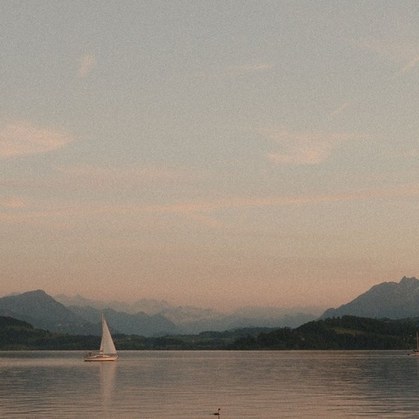}&\includegraphics[width=\cellwidth]{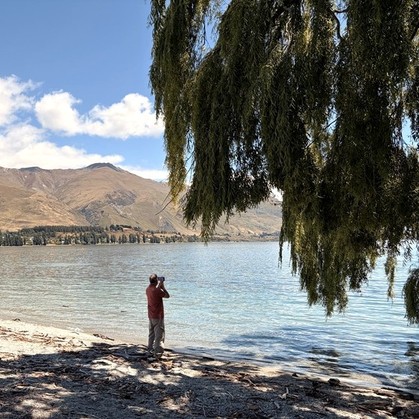}&\includegraphics[width=\cellwidth]{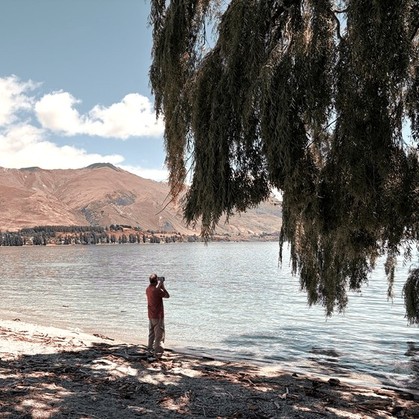}&\includegraphics[width=\cellwidth]{figures/wild/wild07_x}&\includegraphics[width=\cellwidth]{figures/wild/wild07_y}&\includegraphics[width=\cellwidth]{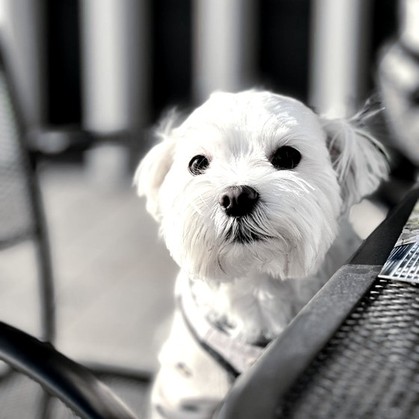}&\includegraphics[width=\cellwidth]{figures/wild/wild07_v} \\
    \includegraphics[width=\cellwidth]{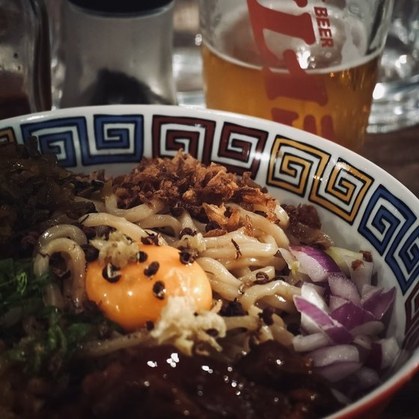}&\includegraphics[width=\cellwidth]{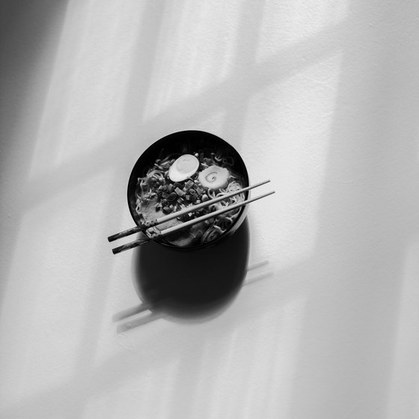}&\includegraphics[width=\cellwidth]{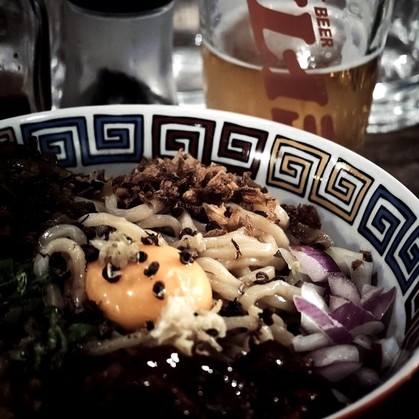}&\includegraphics[width=\cellwidth]{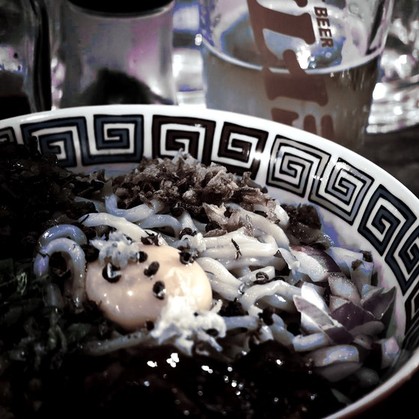}&\includegraphics[width=\cellwidth]{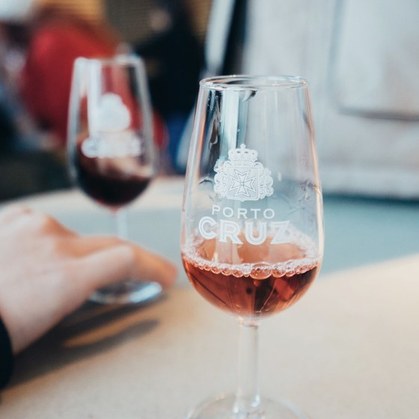}&\includegraphics[width=\cellwidth]{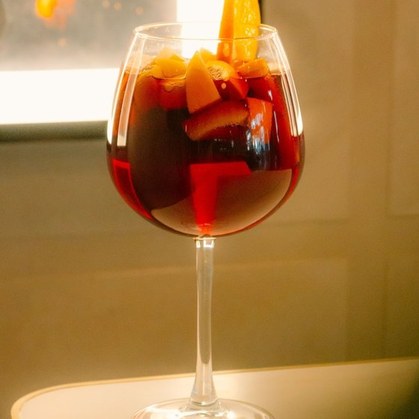}&\includegraphics[width=\cellwidth]{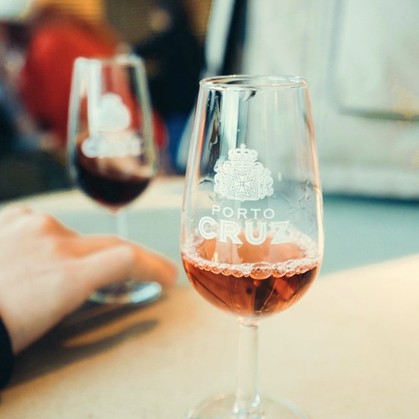}&\includegraphics[width=\cellwidth]{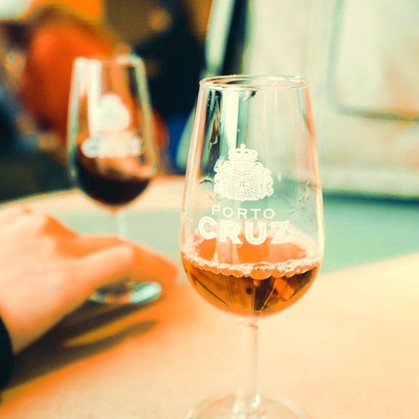} \\
    \includegraphics[width=\cellwidth]{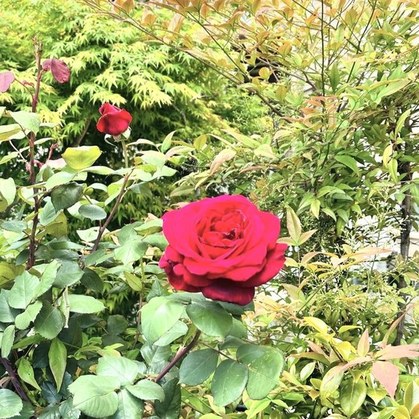}&\includegraphics[width=\cellwidth]{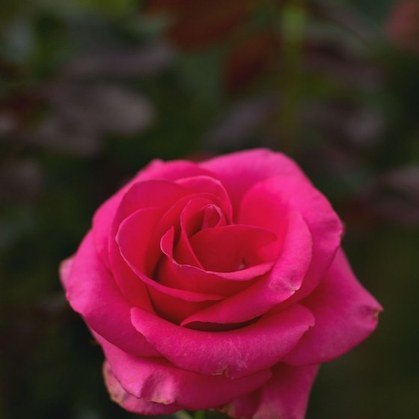}&\includegraphics[width=\cellwidth]{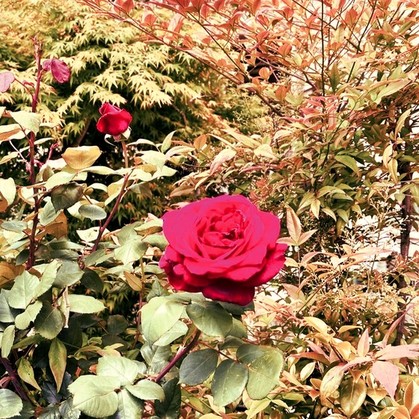}&\includegraphics[width=\cellwidth]{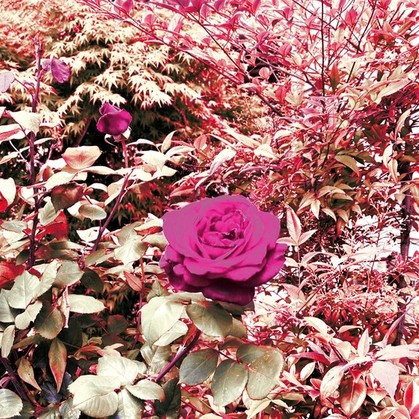}&\includegraphics[width=\cellwidth]{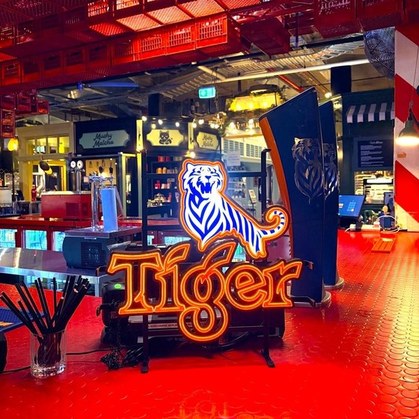}&\includegraphics[width=\cellwidth]{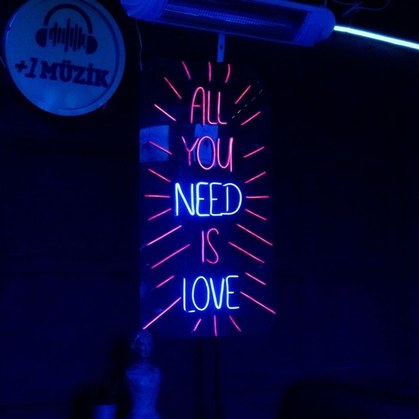}&\includegraphics[width=\cellwidth]{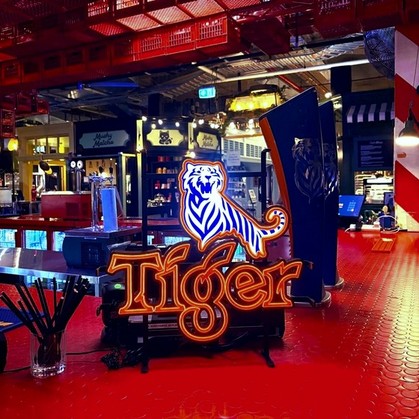}&\includegraphics[width=\cellwidth]{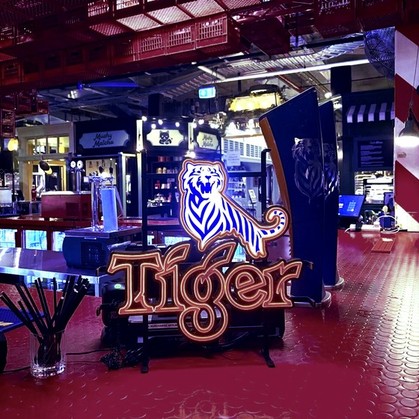} \\
  \end{tabular}
  \caption{Extended in-the-wild gallery (1/2): PREEN applied to photographs with selected references. Each group: content $X$, reference $Y$, PREEN ($\alpha\!=\!1$), and PREEN Vivid ($\alpha\!=\!5$).}
  \label{supp:fig:wild}
\end{figure*}

\begin{figure*}[p]
  \centering
  \setlength{\tabcolsep}{0pt}
  \renewcommand{\arraystretch}{0}
  \newcommand{\cellwidth}{0.122\linewidth}
  \begin{tabular}{@{}c@{}c@{}c@{}c@{\hspace{4pt}}c@{}c@{}c@{}c@{}}
    \scriptsize\bfseries $X$ & \scriptsize\bfseries $Y$ & \scriptsize\bfseries PREEN & \scriptsize\bfseries PREEN Vivid & \scriptsize\bfseries $X$ & \scriptsize\bfseries $Y$ & \scriptsize\bfseries PREEN & \scriptsize\bfseries PREEN Vivid \\
    \includegraphics[width=\cellwidth]{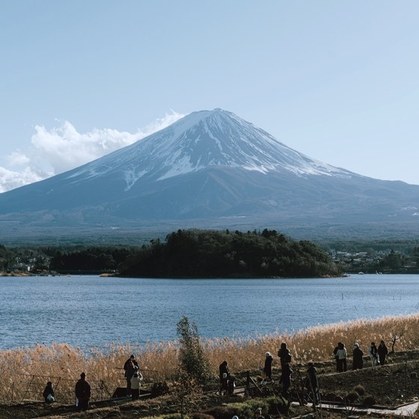}&\includegraphics[width=\cellwidth]{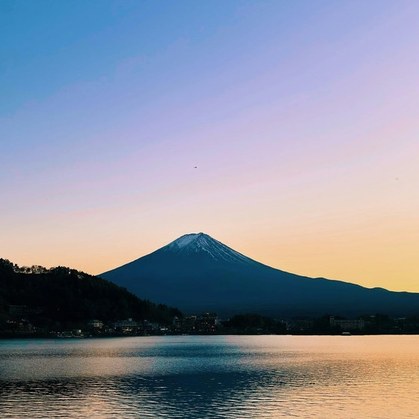}&\includegraphics[width=\cellwidth]{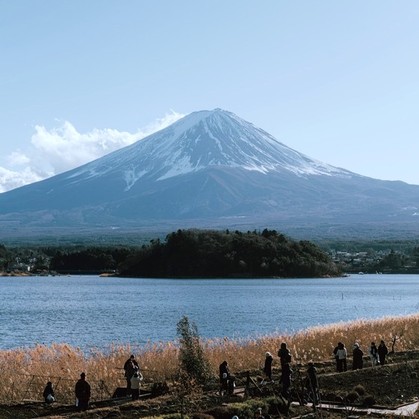}&\includegraphics[width=\cellwidth]{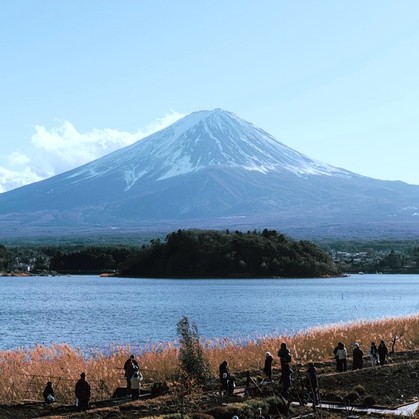}&\includegraphics[width=\cellwidth]{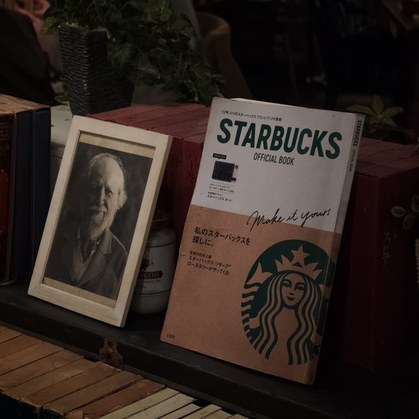}&\includegraphics[width=\cellwidth]{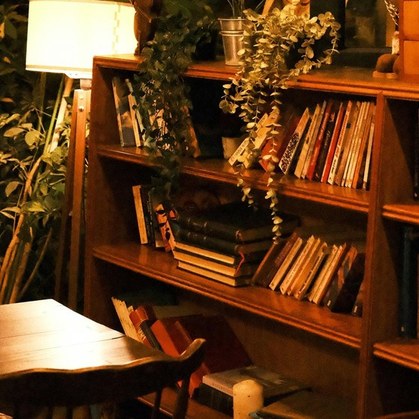}&\includegraphics[width=\cellwidth]{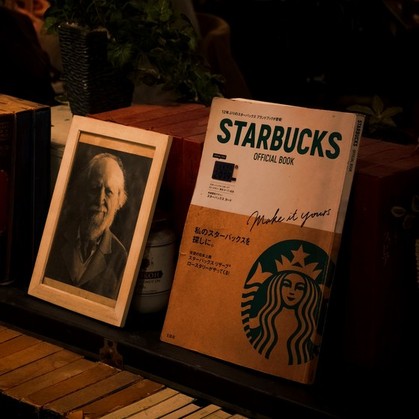}&\includegraphics[width=\cellwidth]{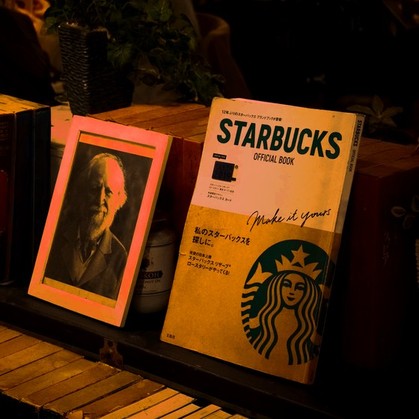} \\
    \includegraphics[width=\cellwidth]{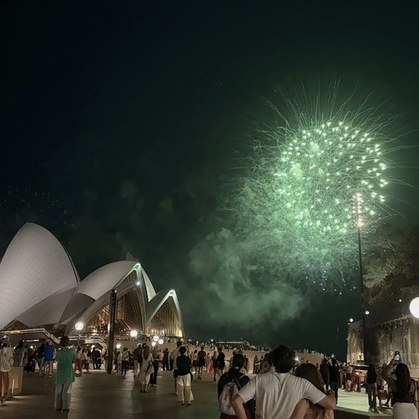}&\includegraphics[width=\cellwidth]{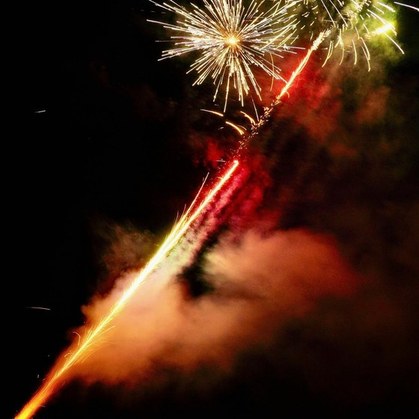}&\includegraphics[width=\cellwidth]{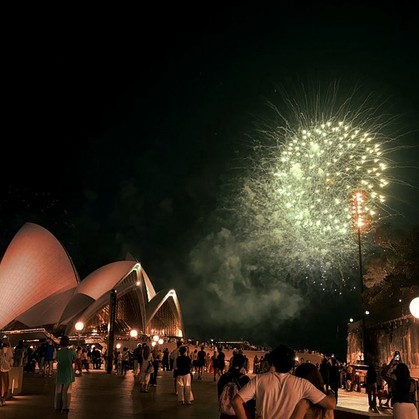}&\includegraphics[width=\cellwidth]{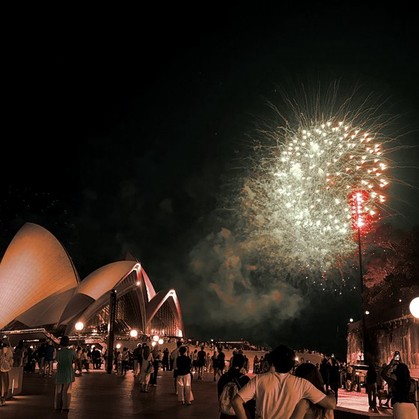}&\includegraphics[width=\cellwidth]{figures/wild/wild15_x}&\includegraphics[width=\cellwidth]{figures/wild/wild15_y}&\includegraphics[width=\cellwidth]{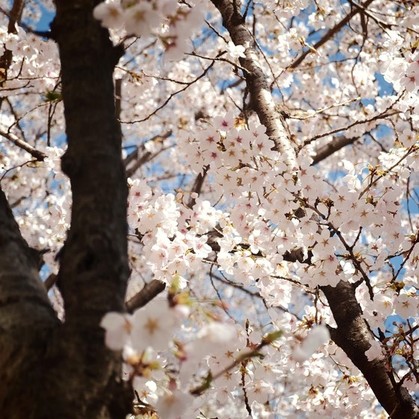}&\includegraphics[width=\cellwidth]{figures/wild/wild15_v} \\
    \includegraphics[width=\cellwidth]{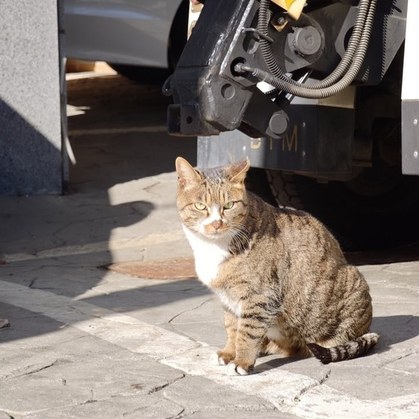}&\includegraphics[width=\cellwidth]{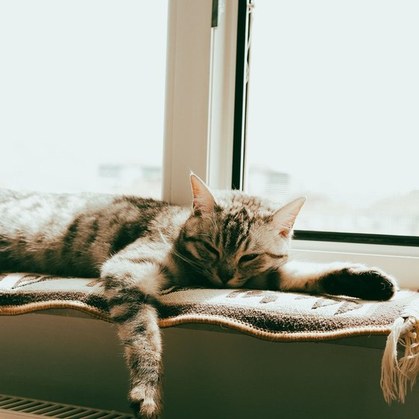}&\includegraphics[width=\cellwidth]{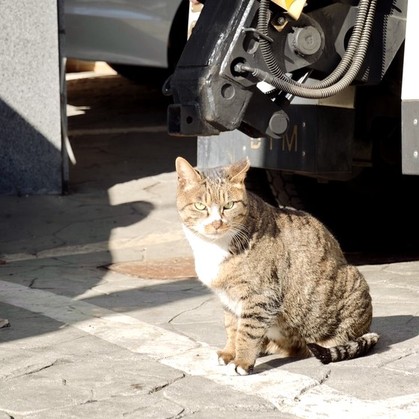}&\includegraphics[width=\cellwidth]{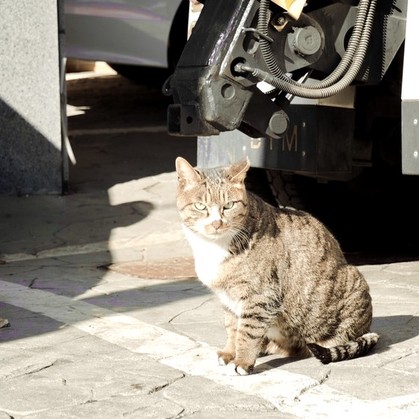}&\includegraphics[width=\cellwidth]{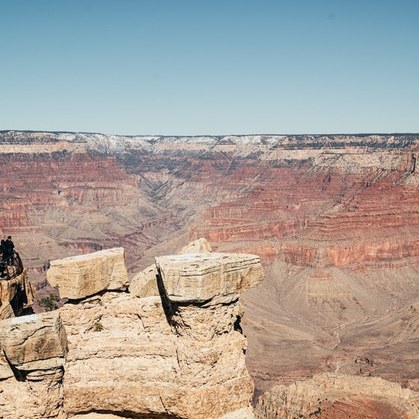}&\includegraphics[width=\cellwidth]{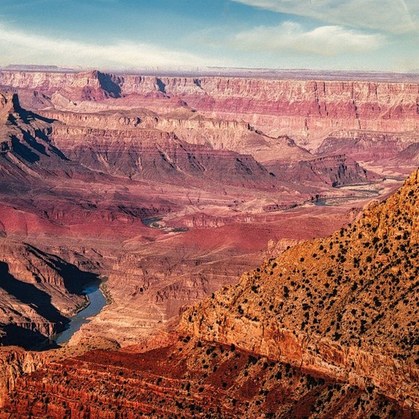}&\includegraphics[width=\cellwidth]{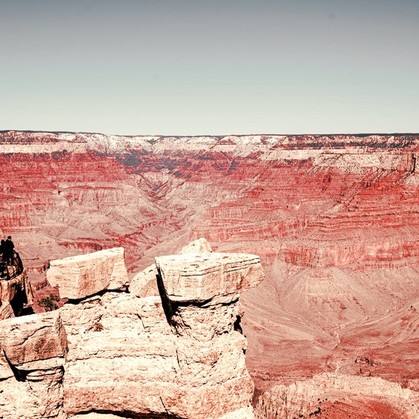}&\includegraphics[width=\cellwidth]{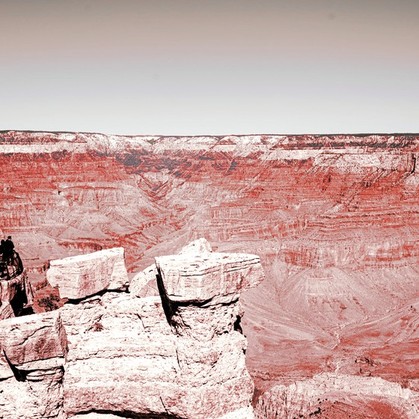} \\
    \includegraphics[width=\cellwidth]{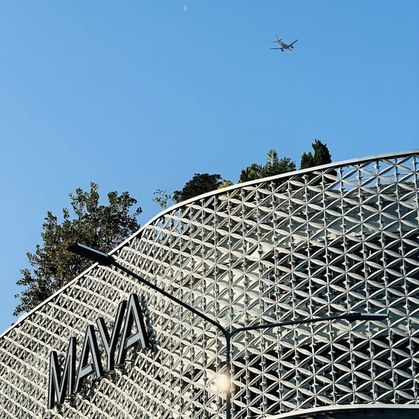}&\includegraphics[width=\cellwidth]{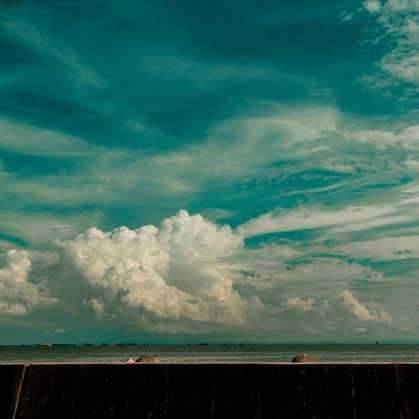}&\includegraphics[width=\cellwidth]{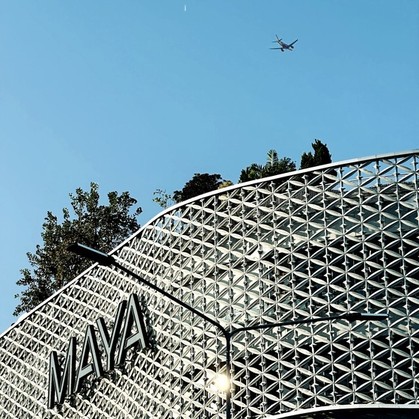}&\includegraphics[width=\cellwidth]{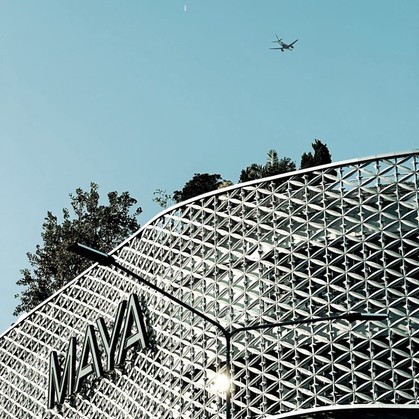}&\includegraphics[width=\cellwidth]{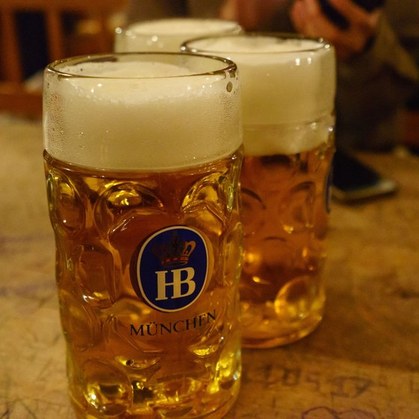}&\includegraphics[width=\cellwidth]{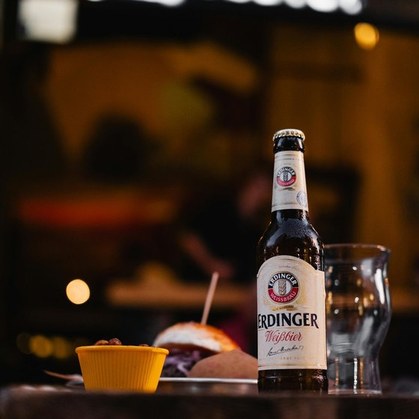}&\includegraphics[width=\cellwidth]{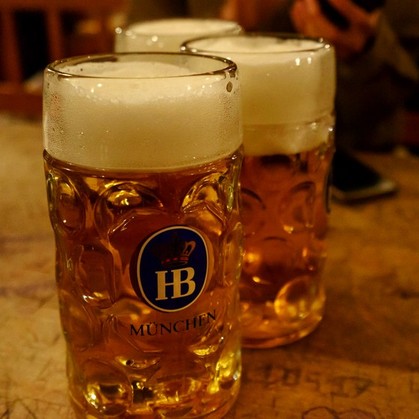}&\includegraphics[width=\cellwidth]{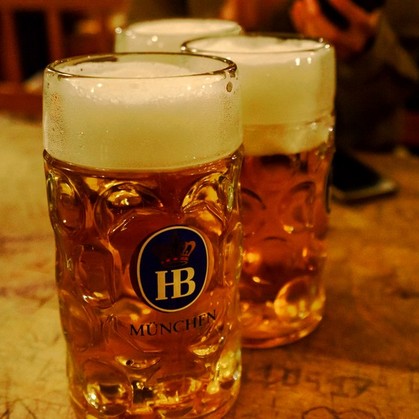} \\
    \includegraphics[width=\cellwidth]{figures/wild/wild20_x}&\includegraphics[width=\cellwidth]{figures/wild/wild20_y}&\includegraphics[width=\cellwidth]{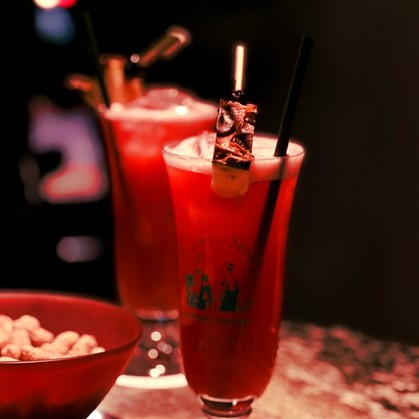}&\includegraphics[width=\cellwidth]{figures/wild/wild20_v}&\includegraphics[width=\cellwidth]{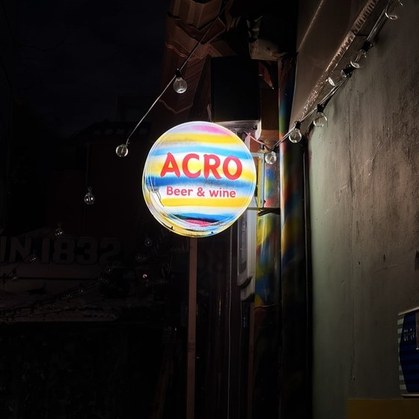}&\includegraphics[width=\cellwidth]{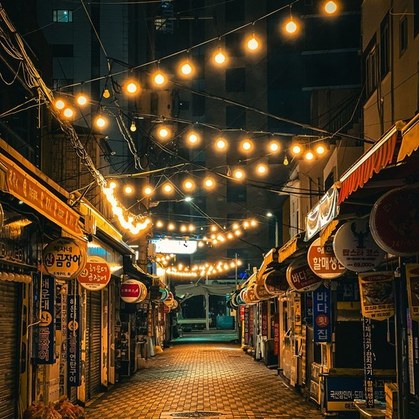}&\includegraphics[width=\cellwidth]{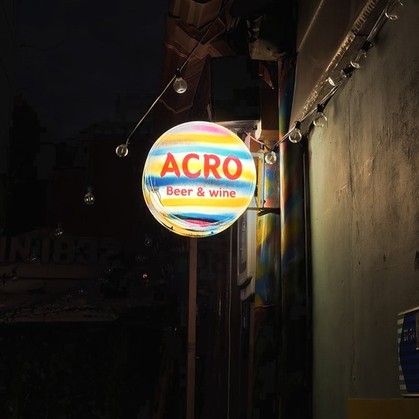}&\includegraphics[width=\cellwidth]{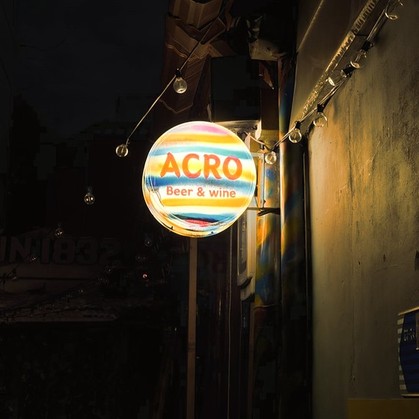} \\
    \includegraphics[width=\cellwidth]{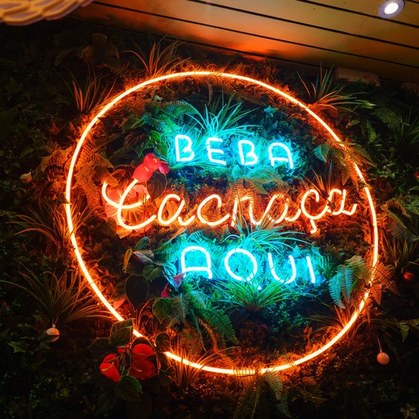}&\includegraphics[width=\cellwidth]{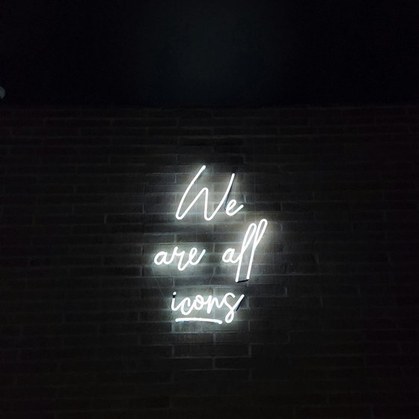}&\includegraphics[width=\cellwidth]{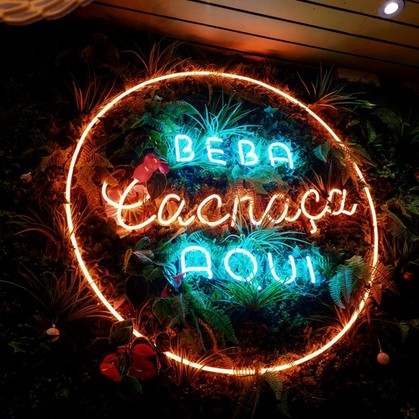}&\includegraphics[width=\cellwidth]{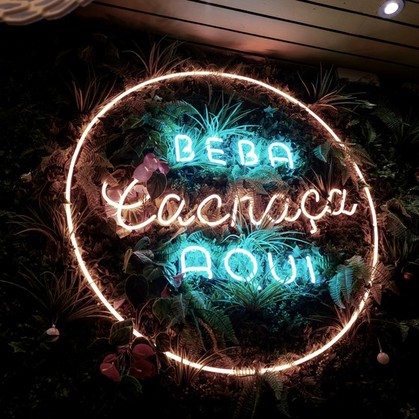}&\includegraphics[width=\cellwidth]{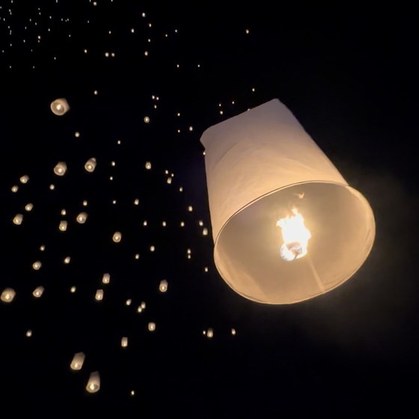}&\includegraphics[width=\cellwidth]{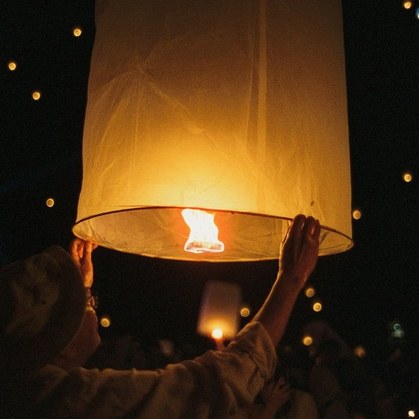}&\includegraphics[width=\cellwidth]{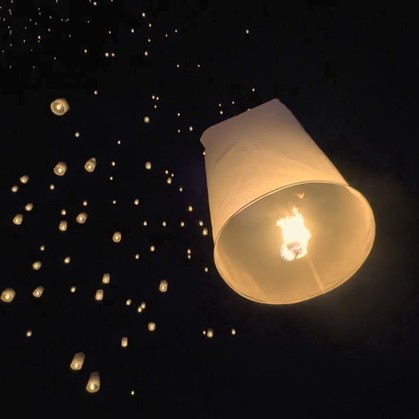}&\includegraphics[width=\cellwidth]{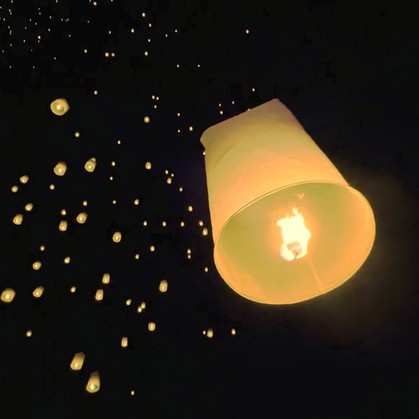} \\
  \end{tabular}
  \caption{Extended in-the-wild gallery (2/2): $X$, $Y$, PREEN ($\alpha\!=\!1$), and PREEN Vivid ($\alpha\!=\!5$).}
  \label{supp:fig:wild2}
\end{figure*}

\noindent\textbf{Extended in-the-wild gallery.} Beyond PST50, we apply PREEN to in-the-wild photographs spanning landscape, indoor, and low-light scenes, each paired with a selected reference image. The content images, including those in Figure~1 of the main paper, are the author's own photographs, and the reference images are free-to-use stock photographs from Pexels (\url{https://www.pexels.com/}). Figures~\ref{supp:fig:wild}--\ref{supp:fig:wild2} show these results. PREEN transfers the global color and tone character of the reference across diverse scenes. As on PST50, looks that hinge on local or otherwise non-global effects are only approximated, since the editor is restricted to global parameters.

\section{Backbone Ablation Checkpoints}

The exact pretrained checkpoints used in the main paper's backbone ablation are CLIP ViT-L/14~\cite{radford2021clip} (\texttt{openai/\allowbreak clip-vit-large-patch14}), DINOv3 ViT-L~\cite{simeoni2025dinov3} (\texttt{facebook/\allowbreak dinov3-vitl16-\allowbreak pretrain-lvd1689m}), SigLIP\,2 large~\cite{tschannen2025siglip2} (\texttt{google/\allowbreak siglip2-large-patch16-256}), and InternViT-300M~\cite{chen2024internvl,gao2024miniinternvl} (\texttt{OpenGVLab/\allowbreak InternViT-300M-448px}). Each backbone runs at its native input resolution, with the MLP head, training schedule, and loss held identical, so rows differ only in the frozen encoder and its native resolution.
\section{Additional Implementation Details}

The five handcrafted color-statistic features have fixed dimensionality: RGB log-ratio ($3$), HSL channel-mean difference ($3$), per-image RGB mean/std for both images ($12$), a soft RGB histogram with $16$ bins per channel for both images ($96$), and per-channel $p_{10}/p_{50}/p_{90}$ percentile differences ($9$), for $123$ dimensions alongside the two $768$-d CLIP image embeddings. Selected checkpoints occur at $1.1$k training steps (Phase~1) and $4.6$k (Phase~2). The Phase~1 inversion optimizes the pre-activation preset with AdamW (learning rate $0.2$, weight decay $0.01$) for $1{,}000$ steps per pair. AdamW's default decoupled weight decay ($0.01$) additionally shrinks the pre-activation, separate from the explicit $10^{-3}$ penalty stated in the main paper, both nudging the recovered preset toward neutral. For the single-stage ablation in the main paper, replacing the Phase~2-only variant's full MLP fine-tuning with a LoRA adapter on a randomly initialized (untrained) MLP base atop the same LoRA-adapted CLIP encoder reaches only $\Delta E^{*}_{ab}\,9.57$ on PST50, versus $9.07$ for the full pipeline and $9.12$ for Phase~2-only full fine-tuning, consistent with a low-rank update being underexpressive without the Phase~1 pretrained base.

\begin{table}[ht]
\centering
\caption{Training hyperparameters by phase. Throughout, the backbone is CLIP ViT-L/14 with frozen base weights and the regressor is two parallel residual MLP trunks $[1024,1024,1024,256]$ (tone and chroma); training uses the few hundred thousand $(X,Y)$ pairs described in the main paper.}
\small
\setlength{\tabcolsep}{4pt}
\begin{tabular}{@{}l p{0.74\columnwidth}@{}}
\toprule
\textbf{Phase} & \textbf{Settings} \\
\midrule
$1$ (inversion) & AdamW, lr $0.2$, $1{,}000$ steps per pair; CIELAB weight $\lambda_{\mathrm{lab}}{=}20$; pre-activation decay $10^{-3}$; AdamW weight decay $0.01$. \\
\addlinespace[2pt]
$1$ (regression) & AdamW, lr $3{\times}10^{-5}$; loss $=$ Huber($\beta{=}0.1$) on $(P_{\mathrm{pred}},\tilde{P})$ $+$ $\ell_2$ preset regularizer ($\lambda{=}10^{-2}$); checkpoint at $1.1$k steps. \\
\addlinespace[2pt]
$2$ (LoRA) & AdamW, lr $10^{-4}$, effective batch $64$ ($16{\times}4$ gradient accumulation), \texttt{bf16}; LoRA rank $16$, alpha $32$; image-level loss $w_{\mathrm{pix}}{=}0.2$, $w_{\mathrm{ab}}{=}5.0$, $w_{\mathrm{L}}{=}1.0$; $\ell_2$ preset regularizer $\lambda{=}10^{-2}$; identity ratio $0.1$; $Y_{\mathrm{aug}}{=}$ random $80$--$100\%$ crop $+$ horizontal flip; checkpoint at $4.6$k steps, i.e., ${\approx}294$k $(X,Y)$ pairs seen ($4.6\text{k}{\times}64$). \\
\bottomrule
\end{tabular}
\end{table}

\section{VLM Prompt Template}

The main paper reports off-the-shelf foundation-model baselines that predict the $29$-parameter preset directly. Each frontier VLM (Gemini~3.1~Pro~\cite{gemini31pro}, Claude~Opus~4.8~\cite{anthropic2025claude4}, and Qwen3-VL~235B~\cite{qwenteam2025qwen3vl}) receives the identical prompt (Figure~\ref{supp:fig:vlmprompt}(b)), with the content and reference images attached (SOURCE and TARGET in the prompt). Each returns a JSON object of $29$ named floats in $[-1,1]$, enforced by a structured-output response schema, and we apply the returned preset through the same editor $E$ used by PREEN. Table~\ref{supp:tab:vlmprompt} compares the two prompt variants of Figure~\ref{supp:fig:vlmprompt}. Under the neutral prompt (a), all three models produce substantially larger edits ($1.7$--$3.7\times$ the guided prompt's mean per-pixel change $|X'{-}X|$) and every metric degrades, while the conservative guidance of (b) recovers the levels reported in the main paper. The guided prompt supplies only generic slider semantics and a magnitude prior (real edits are usually small), with no PST50- or target-specific values, and its two worked examples are a generic neutral-to-warm-vintage preset and a near-zero no-op. Even granted this prior, the VLMs miscalibrate, consistent with editor-grounded calibration being hard to obtain from text prompts alone. VLM outputs are non-deterministic, and we query once per image.

\begin{table}[ht]
\centering
\caption{Prompt variants on PST50: neutral prompt (a) versus guided (b, reported).}
\small
\setlength{\tabcolsep}{3.5pt}
\label{supp:tab:vlmprompt}
\resizebox{\columnwidth}{!}{%
\begin{tabular}{l l c c c c}
\toprule
\textbf{Model} & \textbf{Prompt} & \textbf{PSNR$\uparrow$} & \textbf{SSIM$\uparrow$} & \textbf{HCorr$\uparrow$} & \textbf{LPIPS$\downarrow$} \\
\midrule
Gemini 3.1 Pro  & neutral & 18.27 & 0.675 & 0.211 & 0.234 \\
                & guided  & 21.40 & 0.788 & 0.343 & 0.157 \\
\addlinespace[5pt]
Claude Opus 4.8 & neutral & 19.57 & 0.725 & 0.261 & 0.199 \\
                & guided  & 22.49 & 0.857 & 0.364 & 0.124 \\
\addlinespace[5pt]
Qwen3-VL 235B   & neutral & 15.91 & 0.713 & 0.163 & 0.293 \\
                & guided  & 21.87 & 0.886 & 0.352 & 0.124 \\
\bottomrule
\end{tabular}}
\end{table}

\begin{figure*}[t]
\centering\footnotesize
{\normalsize\bfseries (a) Neutral prompt (unguided)}\\[2pt]
\begin{minipage}[t]{0.48\linewidth}
\begin{verbatim}
You are a precise photo color-grading assistant.
Predict 29 global photo-editing parameters to
transform SOURCE (first image) to match TARGET
(second image)'s style.

Each parameter is a float in [-1.0, 1.0], where
0.0 means no change and the sign sets direction
(below). Pick the magnitude that best matches
the target's look.

GLOBAL parameters (8):
- temperature: -1 cool/blue -> +1 warm/orange
- contrast:    -1 flat      -> +1 high
- tint:        -1 green     -> +1 magenta
- saturation:  -1 desatur.  -> +1 vivid
- brightness:  -1 darker    -> +1 brighter
- exposure:    -1 under     -> +1 over
- highlights:  -1 recover   -> +1 brighten
- shadows:     -1 crush     -> +1 lift
\end{verbatim}
\end{minipage}\hfill
\begin{minipage}[t]{0.48\linewidth}
\begin{verbatim}
PER-HUE parameters (21 = 7 colors x hue/sat/lum):
red/orange/yellow/green/cyan/blue/magenta, each:
- *_hue: shift the hue of that color range
- *_sat: saturation change for that color
- *_lum: luminance change for that color

Now analyze the two images and predict the
parameters that transform SOURCE into TARGET.
\end{verbatim}
\end{minipage}

\medskip
{\normalsize\bfseries (b) Guided prompt (used for the reported results)}\\[2pt]
\begin{minipage}[t]{0.48\linewidth}
\begin{verbatim}
You are a precise photo color-grading assistant.
Predict 29 global photo-editing parameters to
transform SOURCE (first image) to match TARGET
(second image)'s style.

SCALE GUIDANCE (CRITICAL):
Parameters range [-1.0, 1.0], but REAL edits
rarely exceed +/-0.3.
- +/-0.05 ~ +/-0.15: subtle, fine-tuning
- +/-0.15 ~ +/-0.30: moderate, visible (COMMON)
- +/-0.30 ~ +/-0.50: strong, dramatic (rare)
- +/-0.50 ~ +/-1.0: extreme (very rare)
When in doubt, USE SMALLER VALUES. Most per-hue
parameters should be near 0.0. Over-editing
destroys the image. Prefer subtlety.

GLOBAL parameters (8):
- temperature: -1 cool/blue -> +1 warm/orange
- contrast:    -1 flat      -> +1 high
- tint:        -1 green     -> +1 magenta
- saturation:  -1 desatur.  -> +1 vivid
- brightness:  -1 darker    -> +1 brighter
- exposure:    -1 under     -> +1 over
- highlights:  -1 recover   -> +1 brighten
- shadows:     -1 crush     -> +1 lift
\end{verbatim}
\end{minipage}\hfill
\begin{minipage}[t]{0.48\linewidth}
\begin{verbatim}
PER-HUE parameters (21 = 7 colors x hue/sat/lum):
red/orange/yellow/green/cyan/blue/magenta, each:
- *_hue: shift hue within that color range
- *_sat: saturation delta (small unless boosted)
- *_lum: luminance delta (usually small)
Most per-hue values should be 0.0 or tiny. Only
adjust if a specific color is clearly different.

EXAMPLES:

Ex.1 -- neutral source, warm-vintage target:
{"temperature": 0.15, "contrast": 0.05,
 "tint": 0.0, "saturation": -0.1,
 "brightness": -0.05, "exposure": 0.0,
 "highlights": -0.1, "shadows": 0.15,
 ... 21 per-hue fields, mostly 0.0 ...}

Ex.2 -- source already matches target:
all 29 values near 0.0 (+/-0.03).

Now analyze the two images carefully and predict
the parameters. Be conservative -- prefer small
values.
\end{verbatim}
\end{minipage}
\caption{Prompt templates for the VLM baselines. Both the content and reference images (SOURCE and TARGET) are attached, and the model returns a $29$-field JSON object enforced by a response schema; the worked example abbreviates its 21 per-hue fields, as in the prompt sent, and $\pm$/$\rightarrow$ are written as \texttt{+/-}/\texttt{->}. The reported results use the guided variant (b); see Table~\ref{supp:tab:vlmprompt}.}
\label{supp:fig:vlmprompt}
\end{figure*}

\section{User-Study Protocol}

The pairwise preference studies of the main paper (Section~4.7) were conducted on Prolific (\url{https://www.prolific.com/}). Each study recruited $50$ raters ($150$ in total), assigned in balanced groups targeting five independent ratings per image pair. Invalid responses were excluded before analysis. The seven pairings of Figure~6 of the main paper were collected in three studies. Study~1 compared the default against Deep Preset and against Claude. Study~2 compared Vivid against the default, Deep Preset, and Claude. Study~3 compared the default and Vivid against the SA-LUT teacher. Raters saw the reference $Y$ alongside the two candidate edits. The left--right order was randomized per pair, image filenames were anonymized with content hashes, and methods were never named. The two questions, verbatim: ``Q1. Which better matches the Style?'' (``Color tone, contrast, brightness, overall feel.'') and ``Q2. Which looks more beautiful?'' (``Independent of Style --- which one looks better as a standalone photo?''). Participation was restricted to desktop users. Within a pair's ratings, majority agreement averages $74$--$81\%$ across the three studies and both questions (Fleiss' $\kappa$ $0.15$--$0.34$), the slight-to-fair agreement range.

Table~\ref{supp:tab:userci} lists every per-opponent win rate with its confidence interval. The pooled rates of the main paper also count PREEN's own default as an opponent, a within-system $\alpha$ comparison rather than a baseline.

\begin{table}[ht]
\centering
\caption{Every pairing of Figure~6 of the main paper: win rate of the first-named PREEN mode, its $95\%$ pair-clustered bootstrap interval ($10$k pair resamples), and $n$ valid judgments per question.}
\small
\setlength{\tabcolsep}{3.5pt}
\label{supp:tab:userci}
\resizebox{\columnwidth}{!}{%
\begin{tabular}{c l c c c}
\toprule
\textbf{Study} & \textbf{Pairing} & \textbf{Style match (\%)} & \textbf{Aesthetic (\%)} & $n$ \\
\midrule
1 & Default vs.\ Deep Preset ($\hat{P}$) & $61.8\,[53.0,70.6]$ & $67.5\,[60.6,74.0]$ & $246$ \\
  & Default vs.\ Claude Opus 4.8 & $52.0\,[43.2,60.8]$ & $52.5\,[44.7,60.6]$ & $244$ \\
2 & Vivid vs.\ Deep Preset ($\hat{P}$) & $61.6\,[53.3,69.7]$ & $47.2\,[38.4,56.0]$ & $250$ \\
  & Vivid vs.\ Claude Opus 4.8 & $65.7\,[56.3,74.4]$ & $40.3\,[33.2,47.6]$ & $248$ \\
  & Vivid vs.\ Default & $73.4\,[66.4,79.9]$ & $42.5\,[35.6,49.4]$ & $252$ \\
3 & Default vs.\ SA-LUT & $28.3\,[22.9,34.0]$ & $68.8\,[60.8,76.3]$ & $240$ \\
  & Vivid vs.\ SA-LUT & $44.6\,[36.2,53.3]$ & $73.8\,[66.1,81.0]$ & $240$ \\
\bottomrule
\end{tabular}}
\end{table}

\section{Full PST50 Results}

\begin{figure*}[p]
  \centering
  \setlength{\tabcolsep}{0pt}
  \renewcommand{\arraystretch}{0}
  \newcommand{\cellwidth}{0.095\linewidth}
  \begin{tabular}{@{}c@{}c@{}c@{}c@{}c@{\hspace{4pt}}c@{}c@{}c@{}c@{}c@{}}
    \scriptsize\bfseries $X$ & \scriptsize\bfseries $Y$ & \scriptsize\bfseries PREEN & \scriptsize\bfseries PREEN Vivid & \scriptsize\bfseries Target & \scriptsize\bfseries $X$ & \scriptsize\bfseries $Y$ & \scriptsize\bfseries PREEN & \scriptsize\bfseries PREEN Vivid & \scriptsize\bfseries Target \\
    \includegraphics[width=\cellwidth]{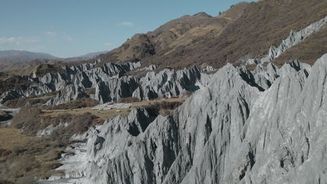}&\includegraphics[width=\cellwidth]{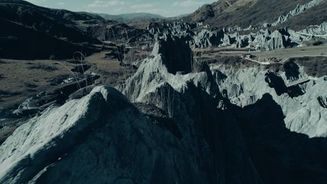}&\includegraphics[width=\cellwidth]{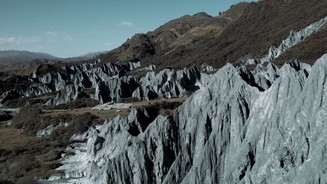}&\includegraphics[width=\cellwidth]{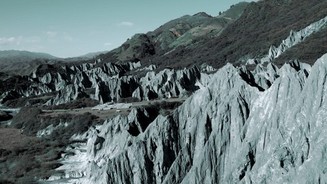}&\includegraphics[width=\cellwidth]{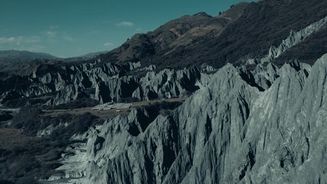}&\includegraphics[width=\cellwidth]{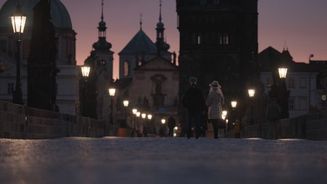}&\includegraphics[width=\cellwidth]{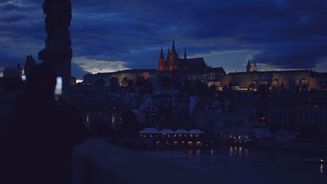}&\includegraphics[width=\cellwidth]{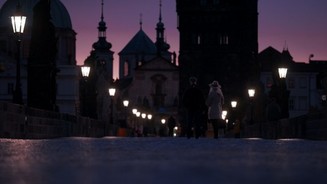}&\includegraphics[width=\cellwidth]{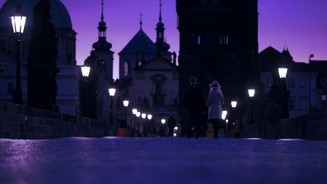}&\includegraphics[width=\cellwidth]{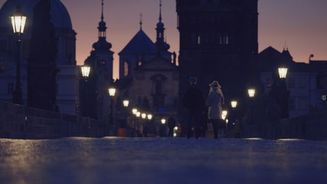} \\
    \includegraphics[width=\cellwidth]{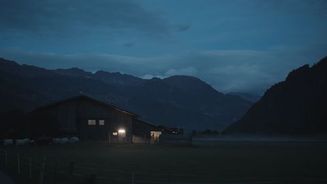}&\includegraphics[width=\cellwidth]{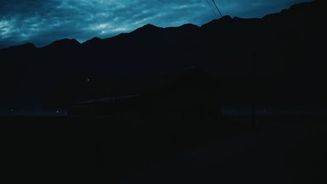}&\includegraphics[width=\cellwidth]{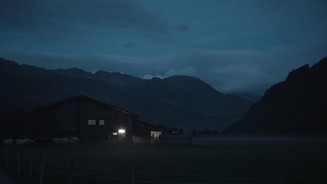}&\includegraphics[width=\cellwidth]{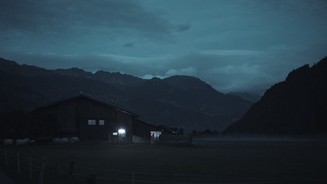}&\includegraphics[width=\cellwidth]{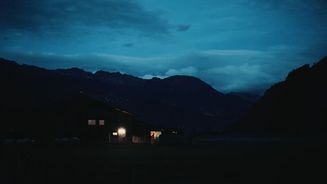}&\includegraphics[width=\cellwidth]{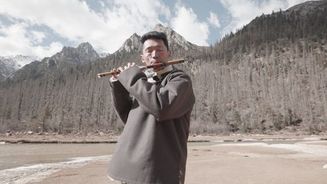}&\includegraphics[width=\cellwidth]{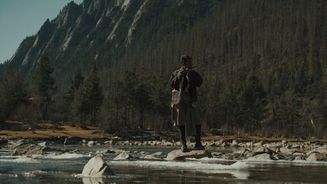}&\includegraphics[width=\cellwidth]{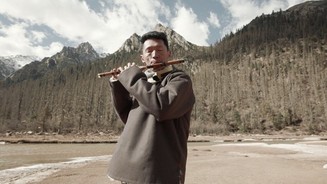}&\includegraphics[width=\cellwidth]{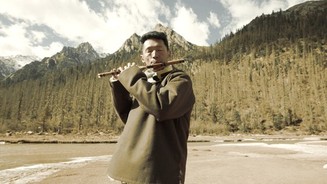}&\includegraphics[width=\cellwidth]{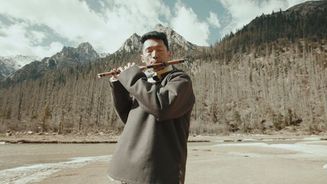} \\
    \includegraphics[width=\cellwidth]{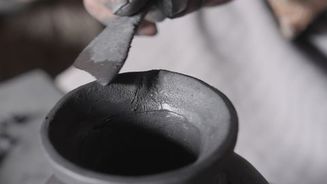}&\includegraphics[width=\cellwidth]{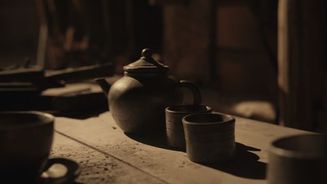}&\includegraphics[width=\cellwidth]{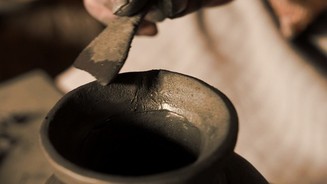}&\includegraphics[width=\cellwidth]{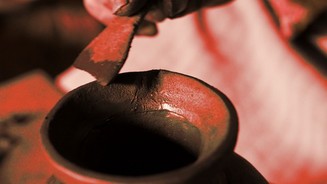}&\includegraphics[width=\cellwidth]{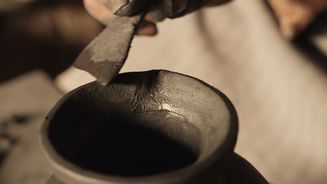}&\includegraphics[width=\cellwidth]{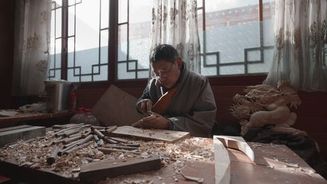}&\includegraphics[width=\cellwidth]{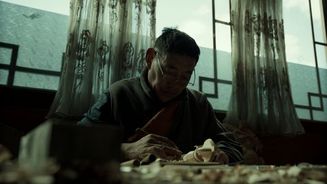}&\includegraphics[width=\cellwidth]{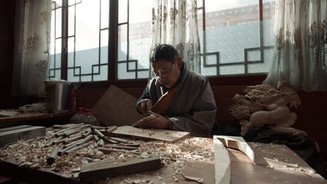}&\includegraphics[width=\cellwidth]{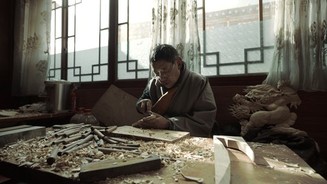}&\includegraphics[width=\cellwidth]{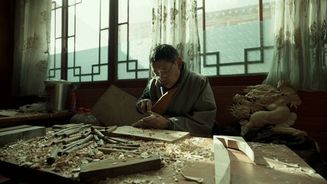} \\
    \includegraphics[width=\cellwidth]{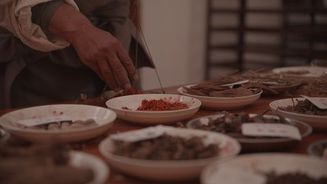}&\includegraphics[width=\cellwidth]{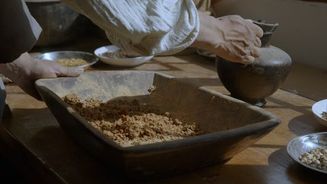}&\includegraphics[width=\cellwidth]{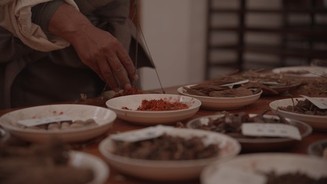}&\includegraphics[width=\cellwidth]{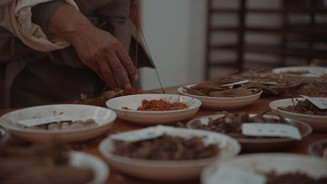}&\includegraphics[width=\cellwidth]{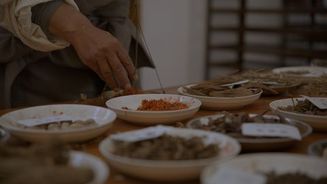}&\includegraphics[width=\cellwidth]{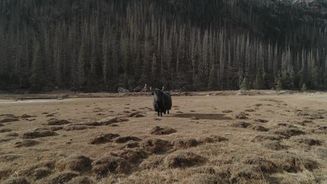}&\includegraphics[width=\cellwidth]{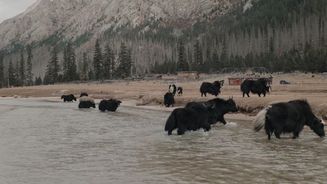}&\includegraphics[width=\cellwidth]{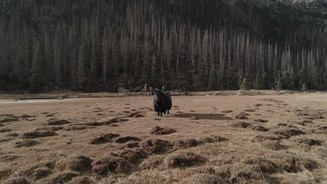}&\includegraphics[width=\cellwidth]{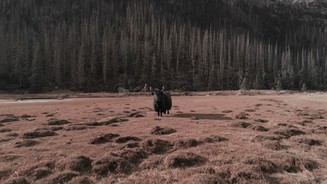}&\includegraphics[width=\cellwidth]{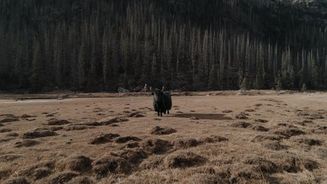} \\
    \includegraphics[width=\cellwidth]{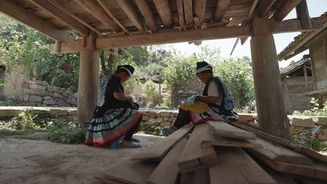}&\includegraphics[width=\cellwidth]{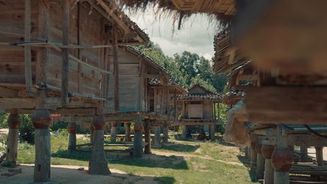}&\includegraphics[width=\cellwidth]{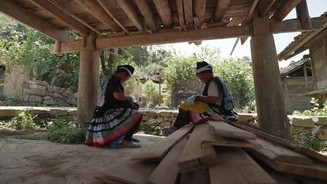}&\includegraphics[width=\cellwidth]{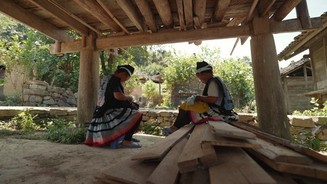}&\includegraphics[width=\cellwidth]{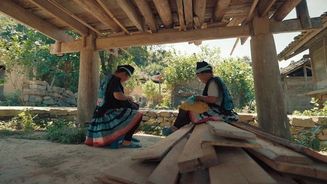}&\includegraphics[width=\cellwidth]{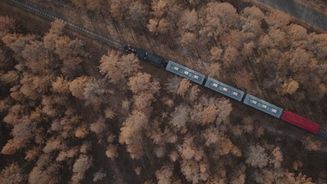}&\includegraphics[width=\cellwidth]{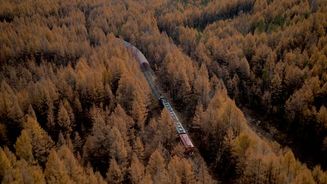}&\includegraphics[width=\cellwidth]{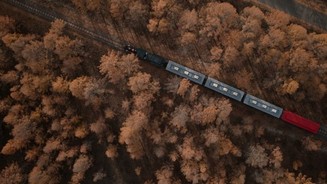}&\includegraphics[width=\cellwidth]{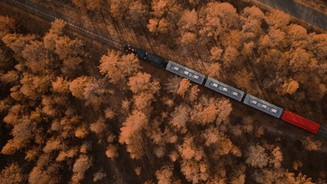}&\includegraphics[width=\cellwidth]{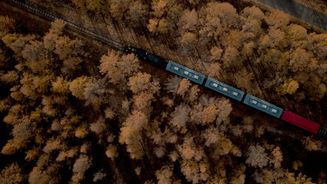} \\
    \includegraphics[width=\cellwidth]{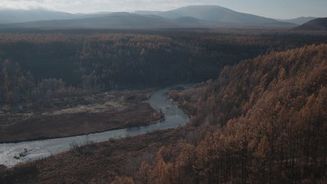}&\includegraphics[width=\cellwidth]{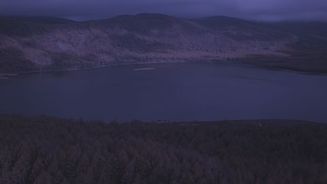}&\includegraphics[width=\cellwidth]{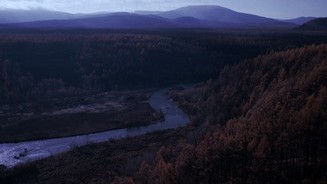}&\includegraphics[width=\cellwidth]{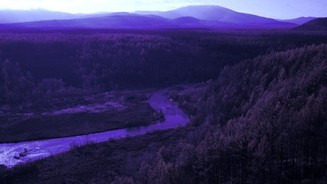}&\includegraphics[width=\cellwidth]{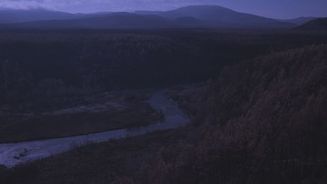}&\includegraphics[width=\cellwidth]{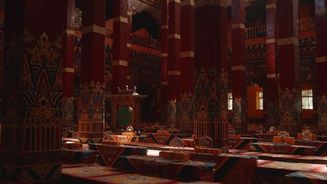}&\includegraphics[width=\cellwidth]{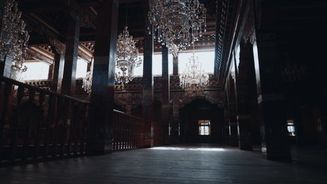}&\includegraphics[width=\cellwidth]{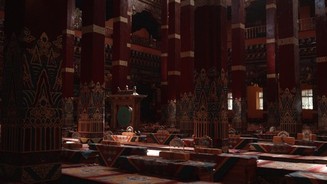}&\includegraphics[width=\cellwidth]{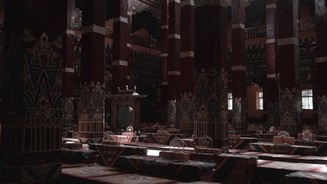}&\includegraphics[width=\cellwidth]{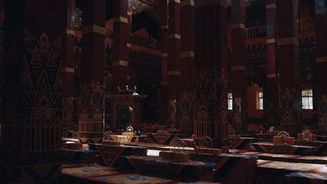} \\
    \includegraphics[width=\cellwidth]{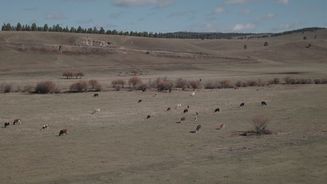}&\includegraphics[width=\cellwidth]{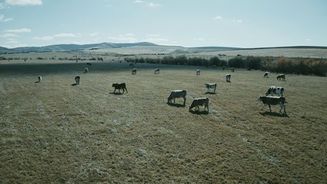}&\includegraphics[width=\cellwidth]{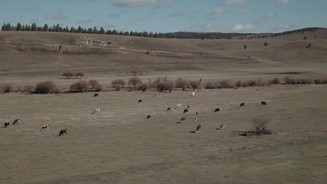}&\includegraphics[width=\cellwidth]{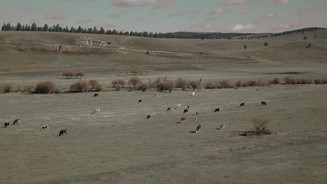}&\includegraphics[width=\cellwidth]{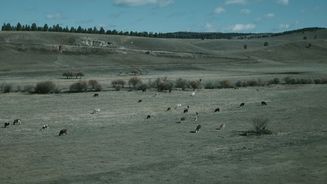}&\includegraphics[width=\cellwidth]{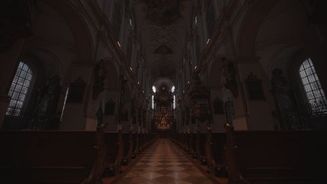}&\includegraphics[width=\cellwidth]{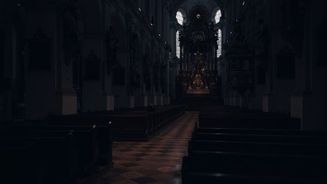}&\includegraphics[width=\cellwidth]{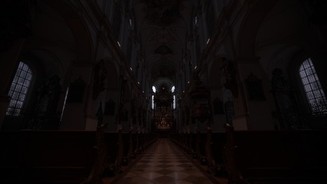}&\includegraphics[width=\cellwidth]{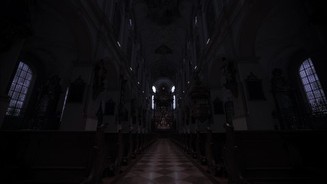}&\includegraphics[width=\cellwidth]{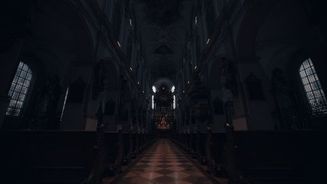} \\
    \includegraphics[width=\cellwidth]{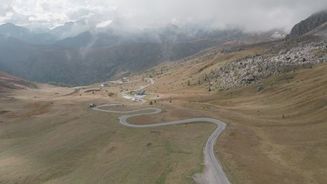}&\includegraphics[width=\cellwidth]{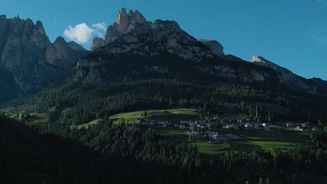}&\includegraphics[width=\cellwidth]{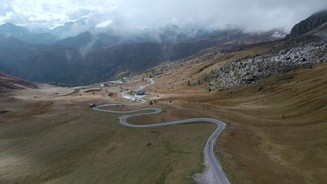}&\includegraphics[width=\cellwidth]{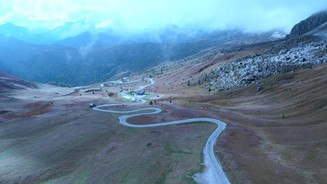}&\includegraphics[width=\cellwidth]{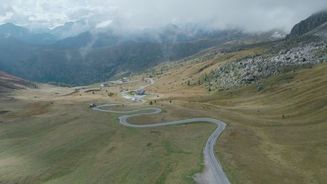}&\includegraphics[width=\cellwidth]{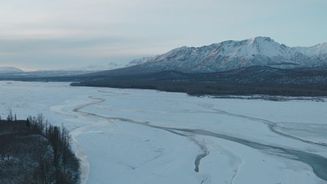}&\includegraphics[width=\cellwidth]{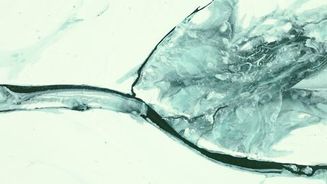}&\includegraphics[width=\cellwidth]{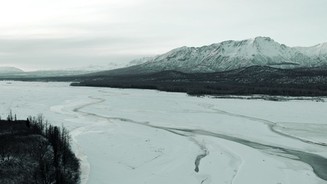}&\includegraphics[width=\cellwidth]{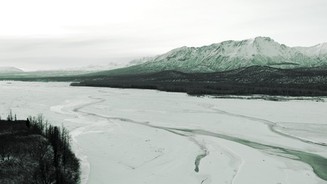}&\includegraphics[width=\cellwidth]{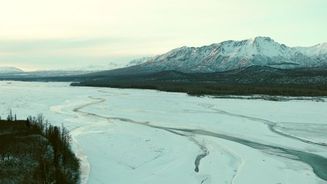} \\
    \includegraphics[width=\cellwidth]{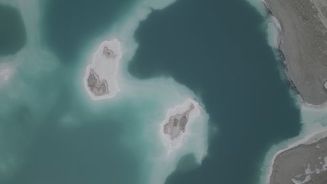}&\includegraphics[width=\cellwidth]{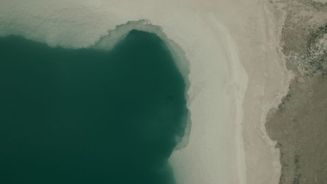}&\includegraphics[width=\cellwidth]{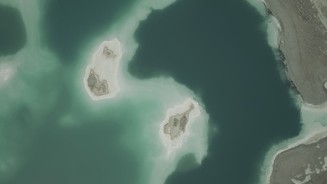}&\includegraphics[width=\cellwidth]{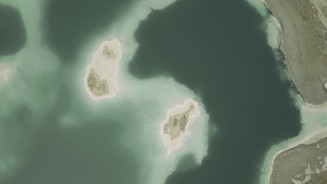}&\includegraphics[width=\cellwidth]{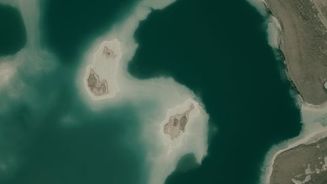}&\includegraphics[width=\cellwidth]{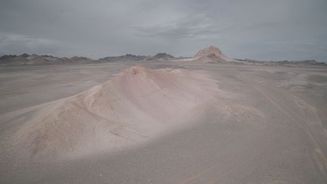}&\includegraphics[width=\cellwidth]{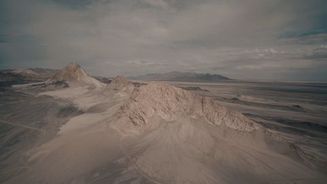}&\includegraphics[width=\cellwidth]{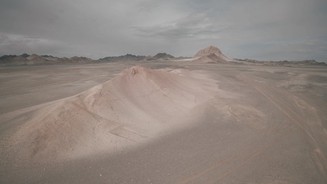}&\includegraphics[width=\cellwidth]{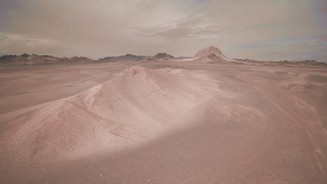}&\includegraphics[width=\cellwidth]{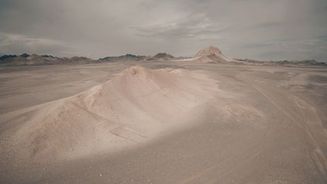} \\
    \includegraphics[width=\cellwidth]{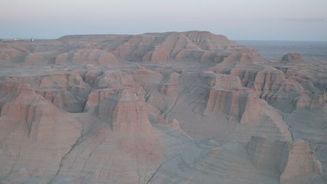}&\includegraphics[width=\cellwidth]{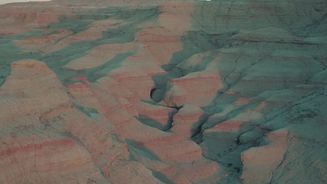}&\includegraphics[width=\cellwidth]{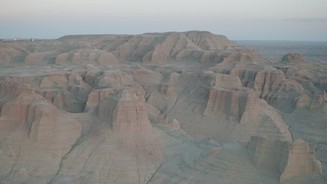}&\includegraphics[width=\cellwidth]{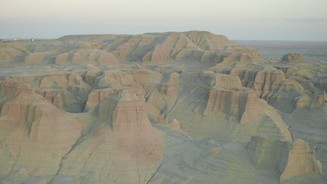}&\includegraphics[width=\cellwidth]{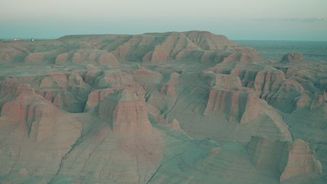}&\includegraphics[width=\cellwidth]{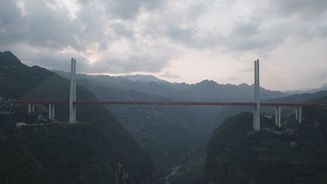}&\includegraphics[width=\cellwidth]{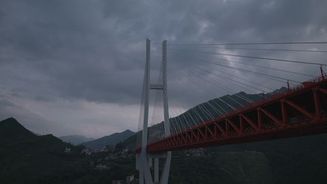}&\includegraphics[width=\cellwidth]{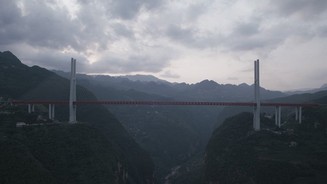}&\includegraphics[width=\cellwidth]{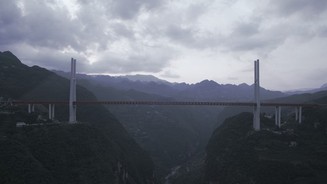}&\includegraphics[width=\cellwidth]{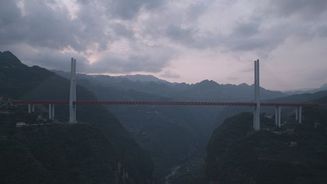} \\
    \includegraphics[width=\cellwidth]{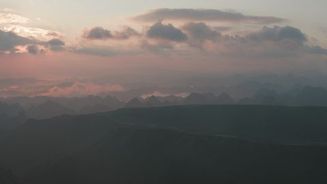}&\includegraphics[width=\cellwidth]{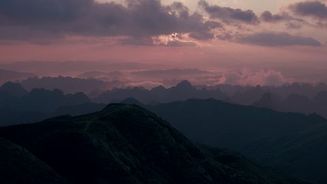}&\includegraphics[width=\cellwidth]{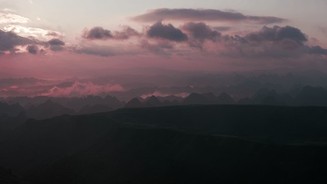}&\includegraphics[width=\cellwidth]{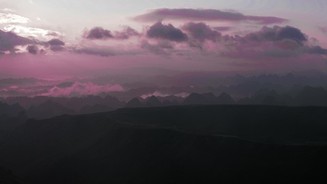}&\includegraphics[width=\cellwidth]{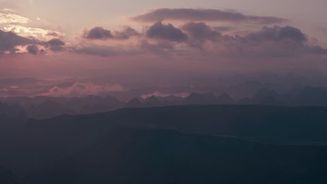}&\includegraphics[width=\cellwidth]{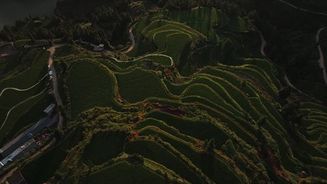}&\includegraphics[width=\cellwidth]{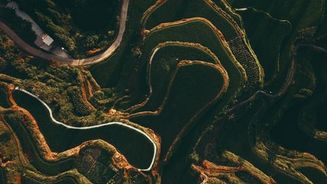}&\includegraphics[width=\cellwidth]{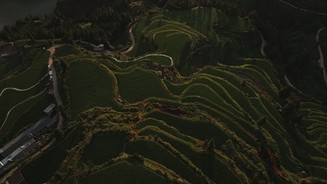}&\includegraphics[width=\cellwidth]{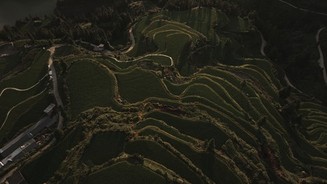}&\includegraphics[width=\cellwidth]{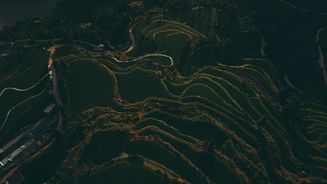} \\
    \includegraphics[width=\cellwidth]{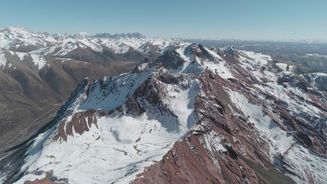}&\includegraphics[width=\cellwidth]{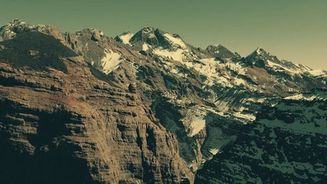}&\includegraphics[width=\cellwidth]{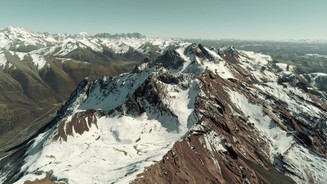}&\includegraphics[width=\cellwidth]{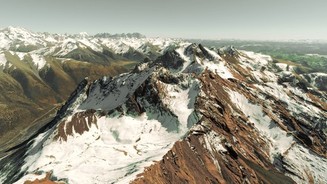}&\includegraphics[width=\cellwidth]{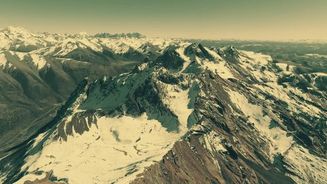}&\includegraphics[width=\cellwidth]{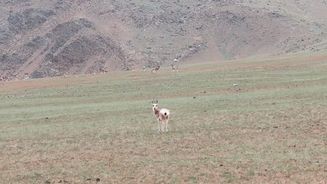}&\includegraphics[width=\cellwidth]{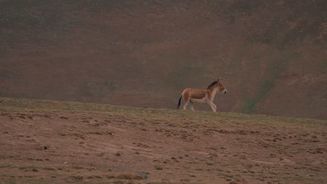}&\includegraphics[width=\cellwidth]{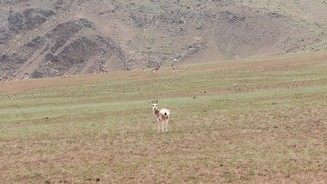}&\includegraphics[width=\cellwidth]{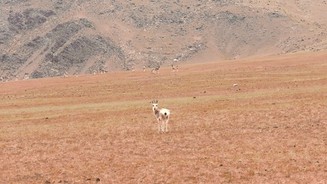}&\includegraphics[width=\cellwidth]{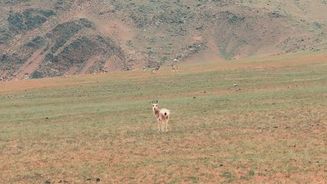} \\
    \includegraphics[width=\cellwidth]{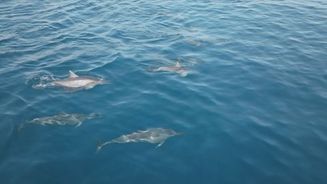}&\includegraphics[width=\cellwidth]{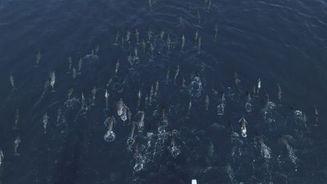}&\includegraphics[width=\cellwidth]{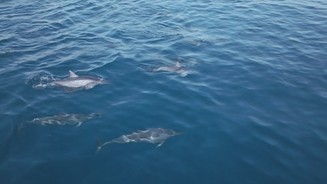}&\includegraphics[width=\cellwidth]{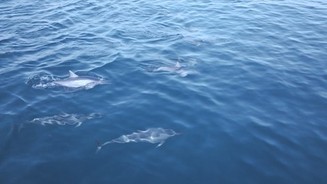}&\includegraphics[width=\cellwidth]{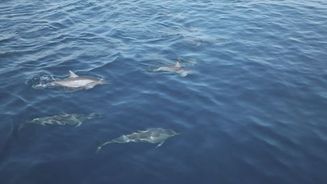}&\includegraphics[width=\cellwidth]{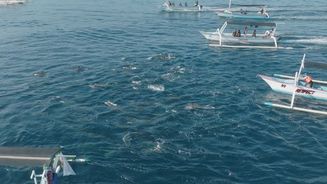}&\includegraphics[width=\cellwidth]{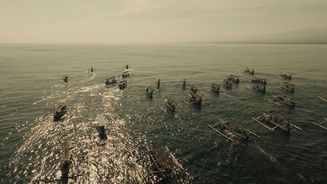}&\includegraphics[width=\cellwidth]{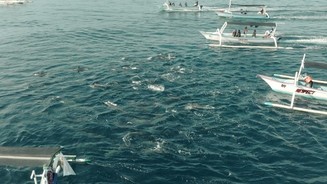}&\includegraphics[width=\cellwidth]{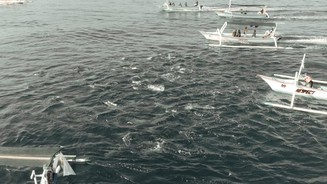}&\includegraphics[width=\cellwidth]{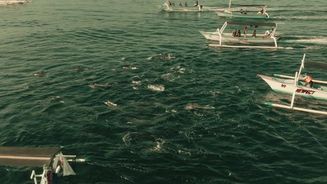} \\
  \end{tabular}
  \caption{Full PST50 paired test set (1/2). Each group: content $X$, reference $Y$, PREEN ($\alpha\!=\!1$), PREEN Vivid ($\alpha\!=\!5$), and the expert target.}
  \label{supp:fig:pst50full:p1}
\end{figure*}

Figures~\ref{supp:fig:pst50full:p1}--\ref{supp:fig:pst50full:u2} show every PST50 case: the $50$ \emph{paired} triples behind every reported number and $50$ \emph{unpaired} cases (each content image with a further benchmark reference and no expert target) for qualitative coverage only. Each group shows $X$, $Y$, PREEN ($\alpha\!=\!1$), PREEN Vivid ($\alpha\!=\!5$), and, for paired cases, the expert target.

\begin{figure*}[p]
  \centering
  \setlength{\tabcolsep}{0pt}
  \renewcommand{\arraystretch}{0}
  \newcommand{\cellwidth}{0.095\linewidth}
  \begin{tabular}{@{}c@{}c@{}c@{}c@{}c@{\hspace{4pt}}c@{}c@{}c@{}c@{}c@{}}
    \scriptsize\bfseries $X$ & \scriptsize\bfseries $Y$ & \scriptsize\bfseries PREEN & \scriptsize\bfseries PREEN Vivid & \scriptsize\bfseries Target & \scriptsize\bfseries $X$ & \scriptsize\bfseries $Y$ & \scriptsize\bfseries PREEN & \scriptsize\bfseries PREEN Vivid & \scriptsize\bfseries Target \\
    \includegraphics[width=\cellwidth]{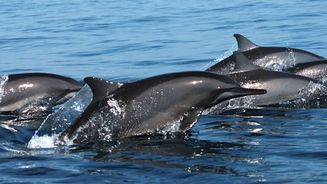}&\includegraphics[width=\cellwidth]{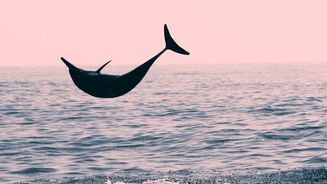}&\includegraphics[width=\cellwidth]{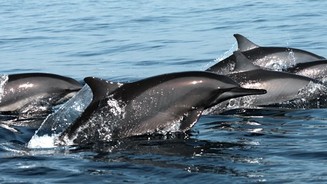}&\includegraphics[width=\cellwidth]{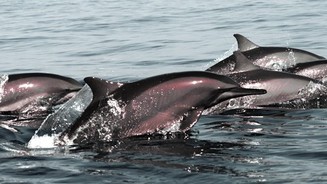}&\includegraphics[width=\cellwidth]{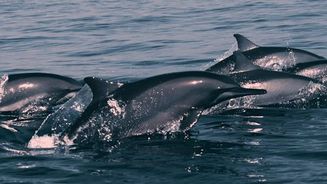}&\includegraphics[width=\cellwidth]{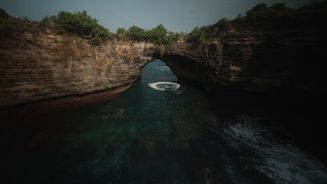}&\includegraphics[width=\cellwidth]{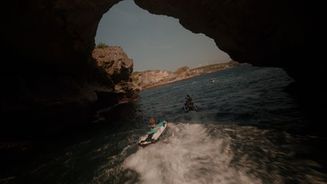}&\includegraphics[width=\cellwidth]{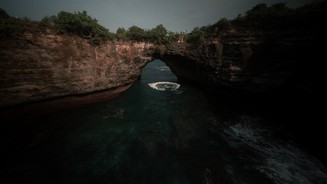}&\includegraphics[width=\cellwidth]{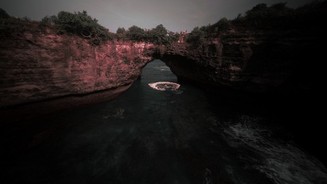}&\includegraphics[width=\cellwidth]{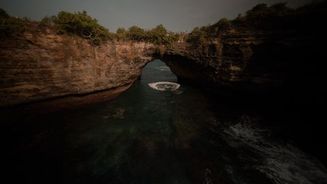} \\
    \includegraphics[width=\cellwidth]{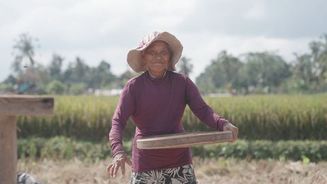}&\includegraphics[width=\cellwidth]{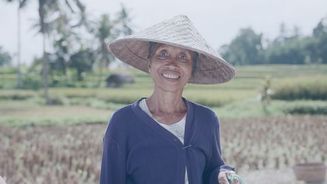}&\includegraphics[width=\cellwidth]{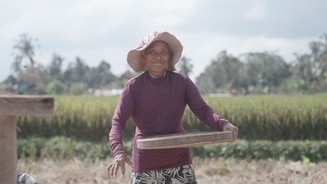}&\includegraphics[width=\cellwidth]{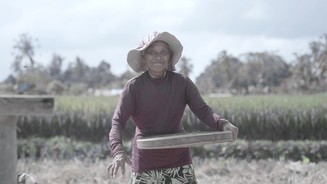}&\includegraphics[width=\cellwidth]{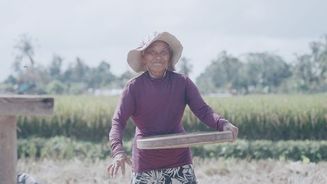}&\includegraphics[width=\cellwidth]{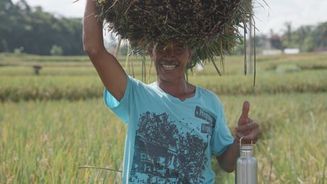}&\includegraphics[width=\cellwidth]{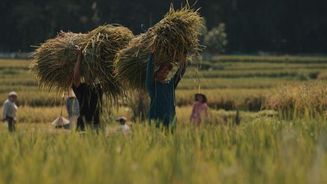}&\includegraphics[width=\cellwidth]{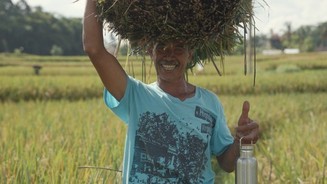}&\includegraphics[width=\cellwidth]{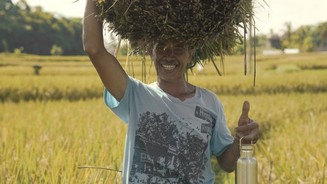}&\includegraphics[width=\cellwidth]{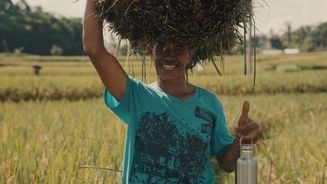} \\
    \includegraphics[width=\cellwidth]{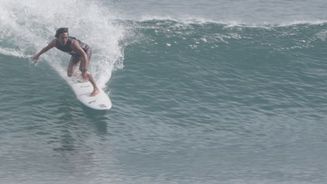}&\includegraphics[width=\cellwidth]{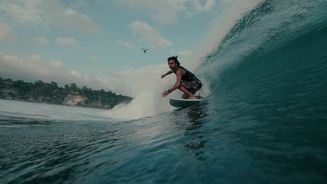}&\includegraphics[width=\cellwidth]{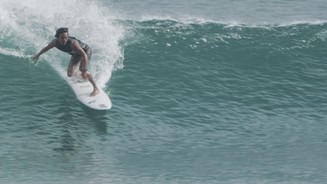}&\includegraphics[width=\cellwidth]{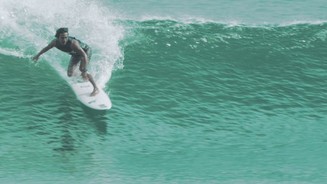}&\includegraphics[width=\cellwidth]{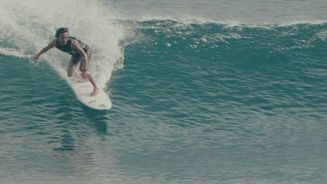}&\includegraphics[width=\cellwidth]{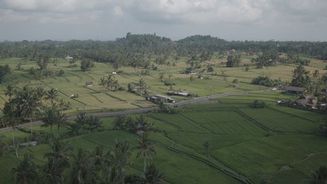}&\includegraphics[width=\cellwidth]{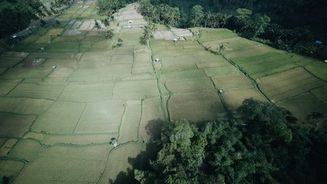}&\includegraphics[width=\cellwidth]{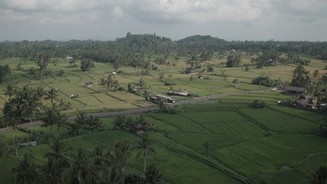}&\includegraphics[width=\cellwidth]{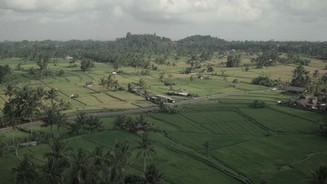}&\includegraphics[width=\cellwidth]{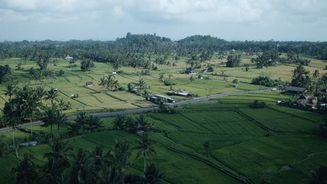} \\
    \includegraphics[width=\cellwidth]{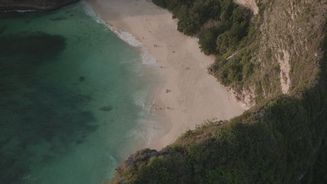}&\includegraphics[width=\cellwidth]{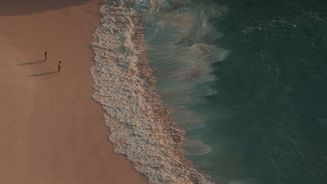}&\includegraphics[width=\cellwidth]{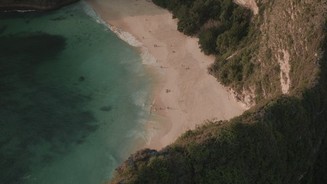}&\includegraphics[width=\cellwidth]{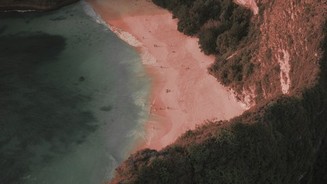}&\includegraphics[width=\cellwidth]{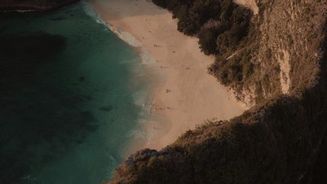}&\includegraphics[width=\cellwidth]{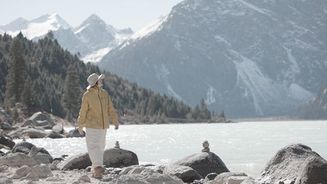}&\includegraphics[width=\cellwidth]{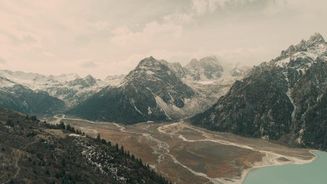}&\includegraphics[width=\cellwidth]{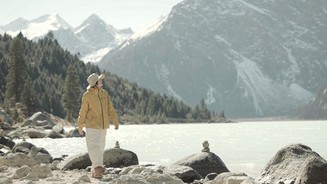}&\includegraphics[width=\cellwidth]{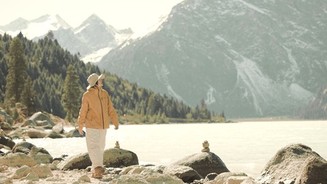}&\includegraphics[width=\cellwidth]{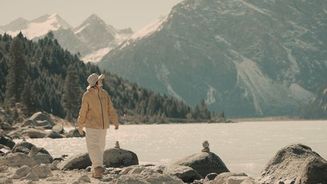} \\
    \includegraphics[width=\cellwidth]{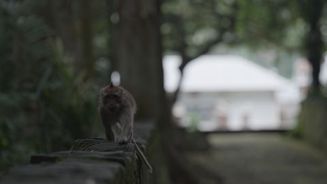}&\includegraphics[width=\cellwidth]{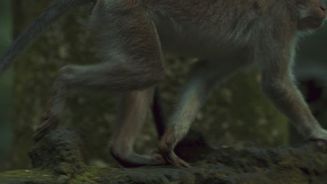}&\includegraphics[width=\cellwidth]{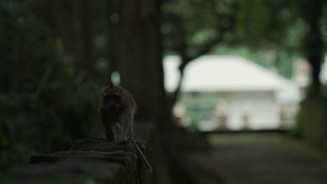}&\includegraphics[width=\cellwidth]{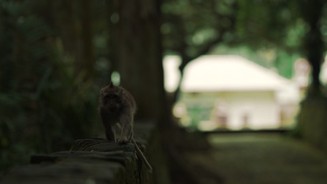}&\includegraphics[width=\cellwidth]{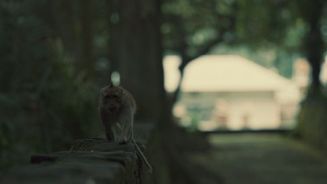}&\includegraphics[width=\cellwidth]{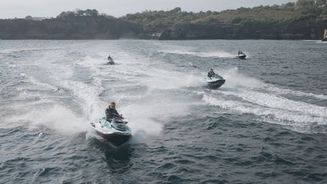}&\includegraphics[width=\cellwidth]{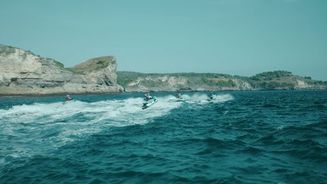}&\includegraphics[width=\cellwidth]{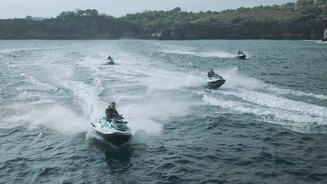}&\includegraphics[width=\cellwidth]{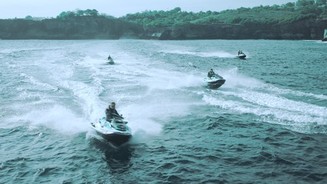}&\includegraphics[width=\cellwidth]{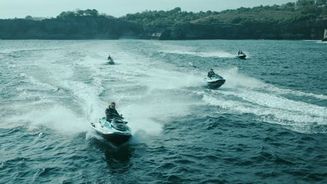} \\
    \includegraphics[width=\cellwidth]{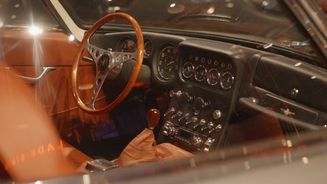}&\includegraphics[width=\cellwidth]{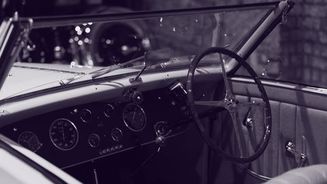}&\includegraphics[width=\cellwidth]{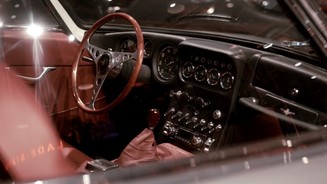}&\includegraphics[width=\cellwidth]{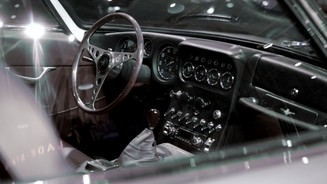}&\includegraphics[width=\cellwidth]{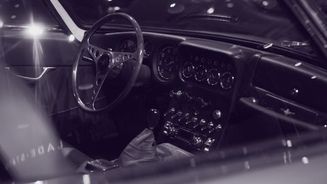}&\includegraphics[width=\cellwidth]{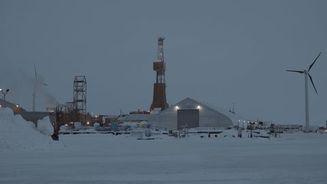}&\includegraphics[width=\cellwidth]{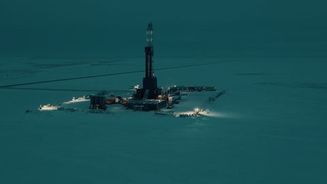}&\includegraphics[width=\cellwidth]{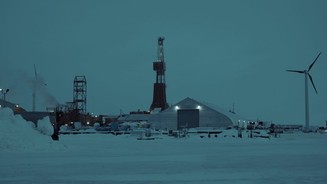}&\includegraphics[width=\cellwidth]{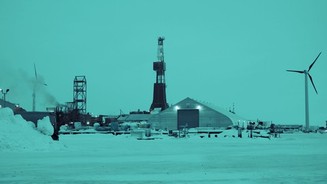}&\includegraphics[width=\cellwidth]{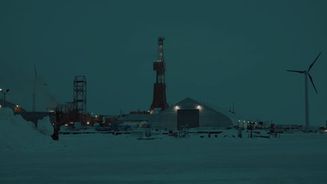} \\
    \includegraphics[width=\cellwidth]{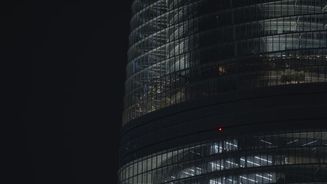}&\includegraphics[width=\cellwidth]{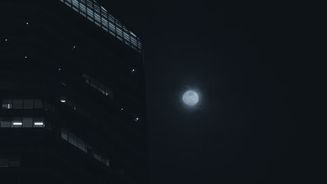}&\includegraphics[width=\cellwidth]{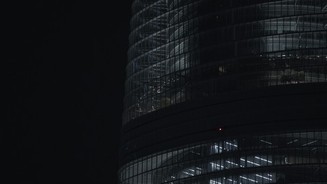}&\includegraphics[width=\cellwidth]{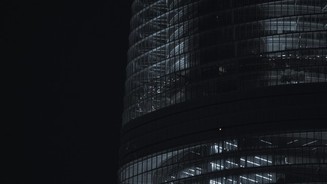}&\includegraphics[width=\cellwidth]{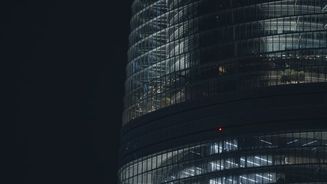}&\includegraphics[width=\cellwidth]{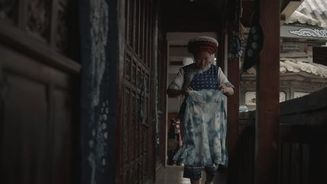}&\includegraphics[width=\cellwidth]{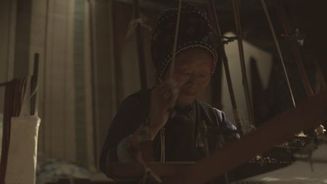}&\includegraphics[width=\cellwidth]{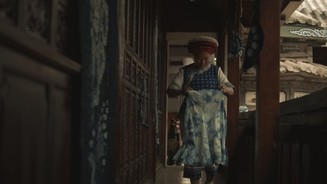}&\includegraphics[width=\cellwidth]{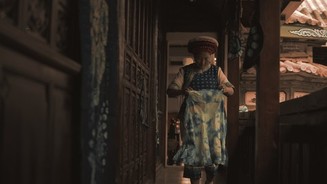}&\includegraphics[width=\cellwidth]{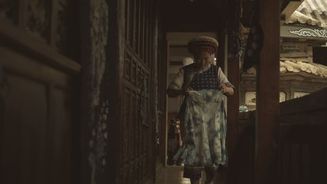} \\
    \includegraphics[width=\cellwidth]{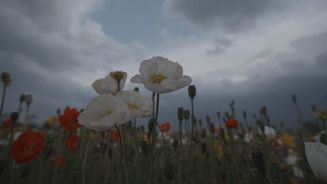}&\includegraphics[width=\cellwidth]{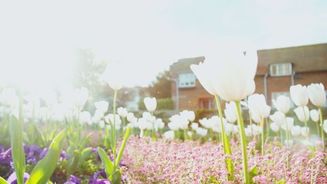}&\includegraphics[width=\cellwidth]{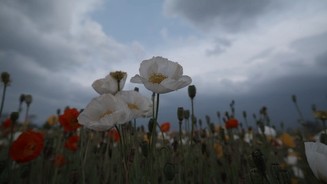}&\includegraphics[width=\cellwidth]{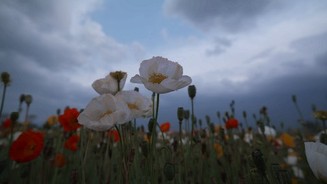}&\includegraphics[width=\cellwidth]{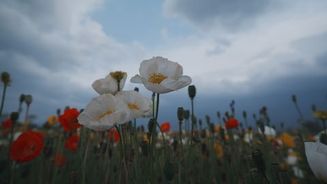}&\includegraphics[width=\cellwidth]{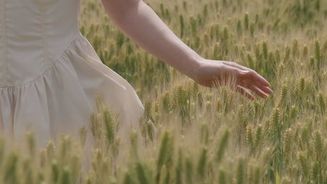}&\includegraphics[width=\cellwidth]{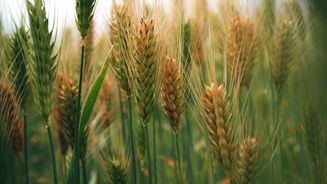}&\includegraphics[width=\cellwidth]{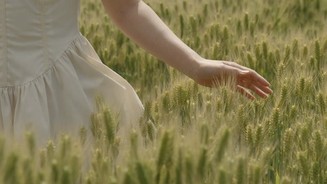}&\includegraphics[width=\cellwidth]{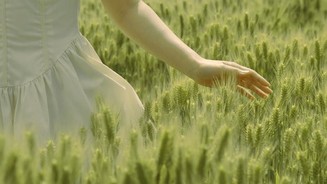}&\includegraphics[width=\cellwidth]{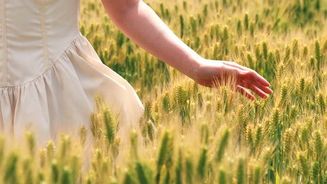} \\
    \includegraphics[width=\cellwidth]{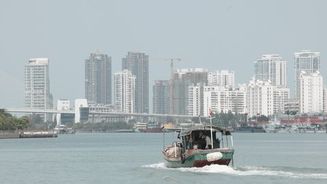}&\includegraphics[width=\cellwidth]{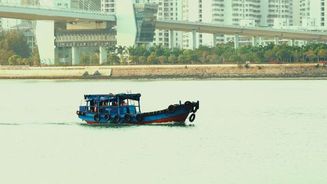}&\includegraphics[width=\cellwidth]{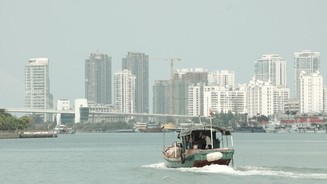}&\includegraphics[width=\cellwidth]{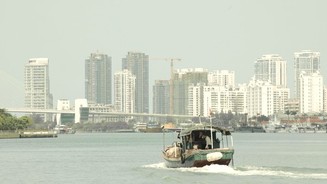}&\includegraphics[width=\cellwidth]{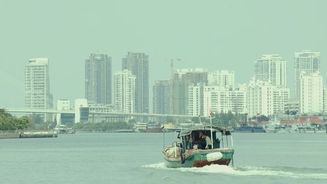}&\includegraphics[width=\cellwidth]{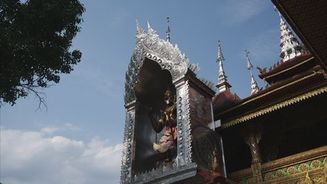}&\includegraphics[width=\cellwidth]{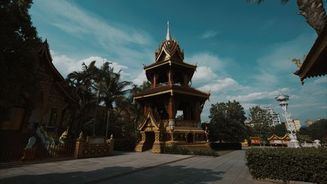}&\includegraphics[width=\cellwidth]{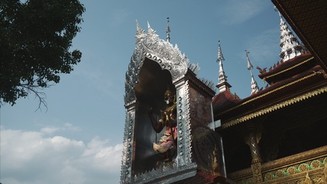}&\includegraphics[width=\cellwidth]{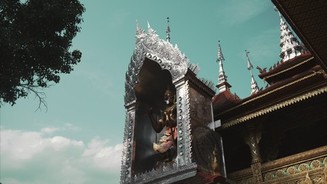}&\includegraphics[width=\cellwidth]{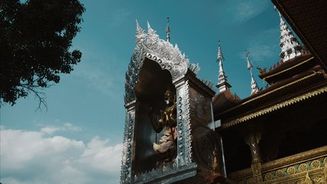} \\
    \includegraphics[width=\cellwidth]{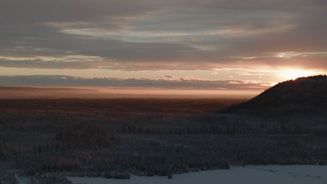}&\includegraphics[width=\cellwidth]{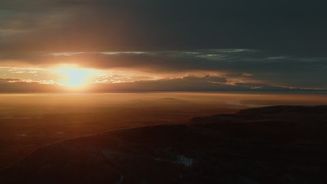}&\includegraphics[width=\cellwidth]{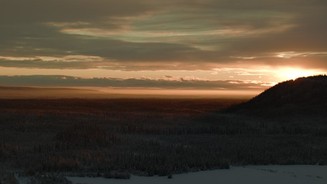}&\includegraphics[width=\cellwidth]{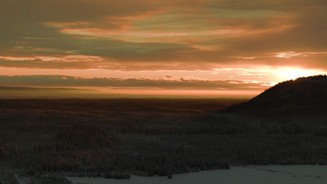}&\includegraphics[width=\cellwidth]{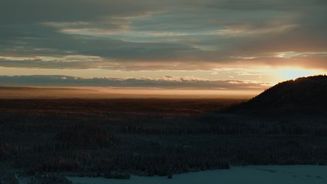}&\includegraphics[width=\cellwidth]{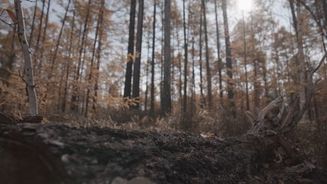}&\includegraphics[width=\cellwidth]{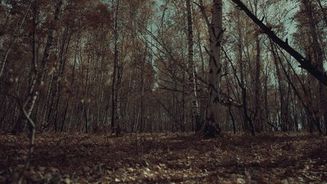}&\includegraphics[width=\cellwidth]{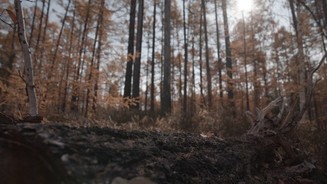}&\includegraphics[width=\cellwidth]{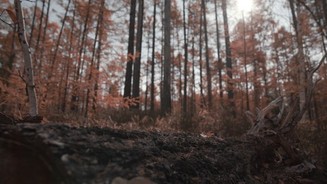}&\includegraphics[width=\cellwidth]{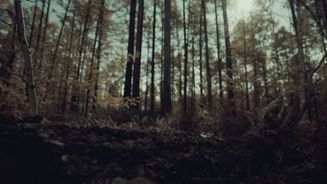} \\
    \includegraphics[width=\cellwidth]{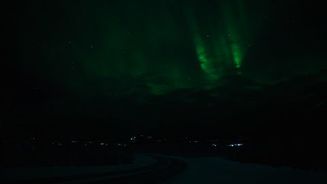}&\includegraphics[width=\cellwidth]{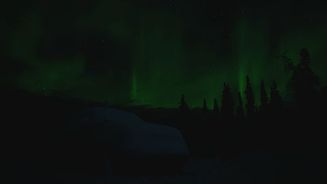}&\includegraphics[width=\cellwidth]{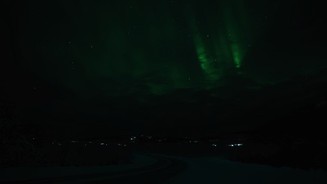}&\includegraphics[width=\cellwidth]{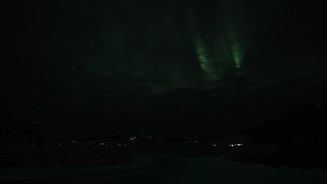}&\includegraphics[width=\cellwidth]{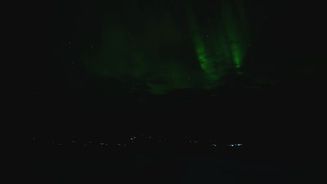}&\includegraphics[width=\cellwidth]{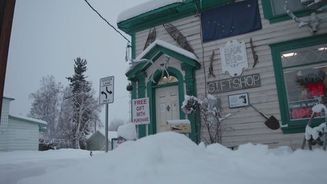}&\includegraphics[width=\cellwidth]{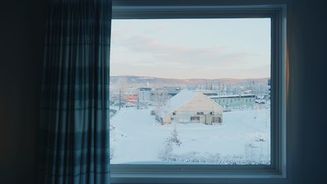}&\includegraphics[width=\cellwidth]{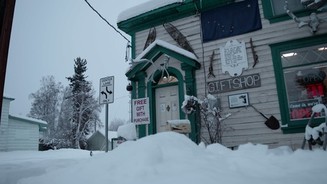}&\includegraphics[width=\cellwidth]{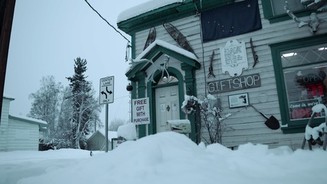}&\includegraphics[width=\cellwidth]{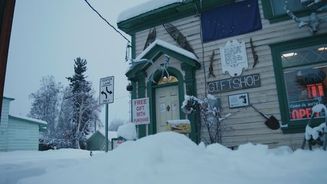} \\
    \includegraphics[width=\cellwidth]{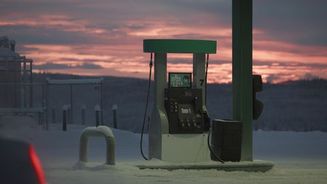}&\includegraphics[width=\cellwidth]{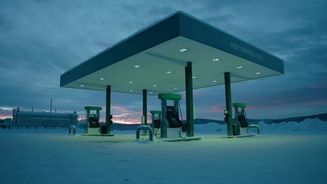}&\includegraphics[width=\cellwidth]{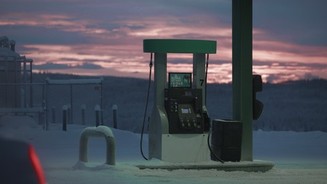}&\includegraphics[width=\cellwidth]{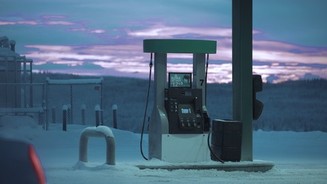}&\includegraphics[width=\cellwidth]{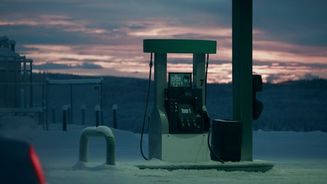}&\includegraphics[width=\cellwidth]{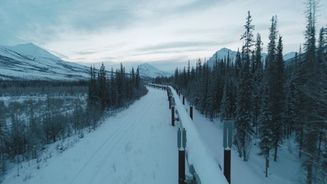}&\includegraphics[width=\cellwidth]{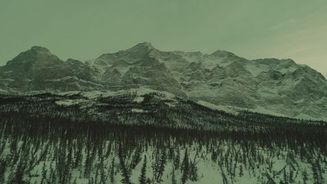}&\includegraphics[width=\cellwidth]{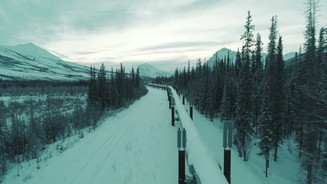}&\includegraphics[width=\cellwidth]{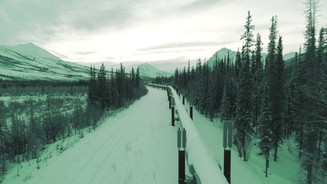}&\includegraphics[width=\cellwidth]{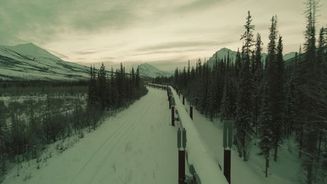} \\
  \end{tabular}
  \caption{Full PST50 paired test set (2/2). Each group: $X$, $Y$, PREEN ($\alpha\!=\!1$), PREEN Vivid ($\alpha\!=\!5$), and the expert target.}
\end{figure*}

\begin{figure*}[p]
  \centering
  \setlength{\tabcolsep}{0pt}
  \renewcommand{\arraystretch}{0}
  \newcommand{\cellwidth}{0.122\linewidth}
  \begin{tabular}{@{}c@{}c@{}c@{}c@{\hspace{4pt}}c@{}c@{}c@{}c@{}}
    \scriptsize\bfseries $X$ & \scriptsize\bfseries $Y$ & \scriptsize\bfseries PREEN & \scriptsize\bfseries PREEN Vivid & \scriptsize\bfseries $X$ & \scriptsize\bfseries $Y$ & \scriptsize\bfseries PREEN & \scriptsize\bfseries PREEN Vivid \\
    \includegraphics[width=\cellwidth]{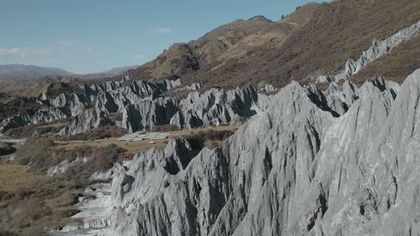}&\includegraphics[width=\cellwidth]{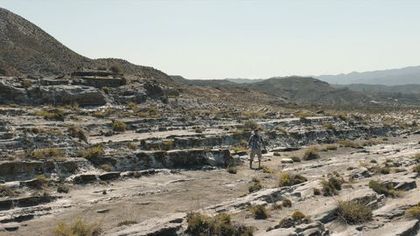}&\includegraphics[width=\cellwidth]{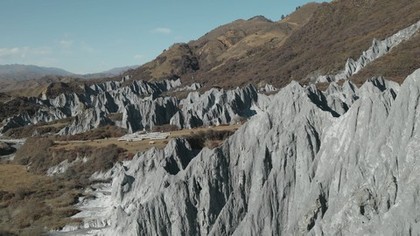}&\includegraphics[width=\cellwidth]{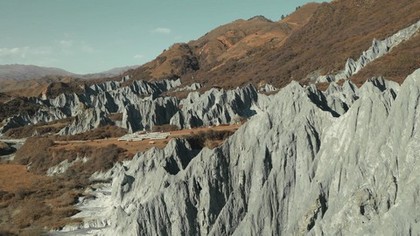}&\includegraphics[width=\cellwidth]{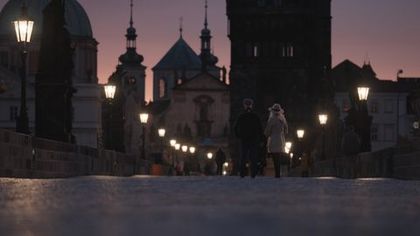}&\includegraphics[width=\cellwidth]{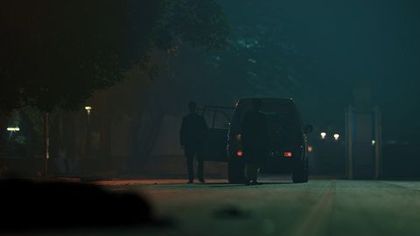}&\includegraphics[width=\cellwidth]{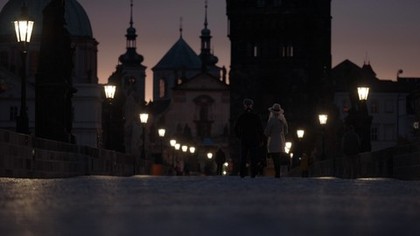}&\includegraphics[width=\cellwidth]{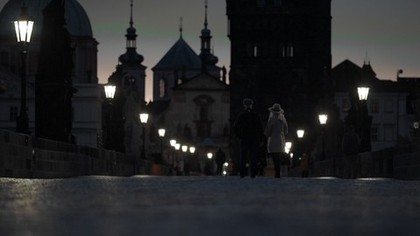} \\
    \includegraphics[width=\cellwidth]{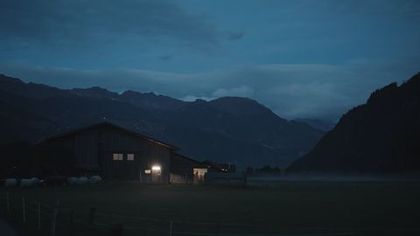}&\includegraphics[width=\cellwidth]{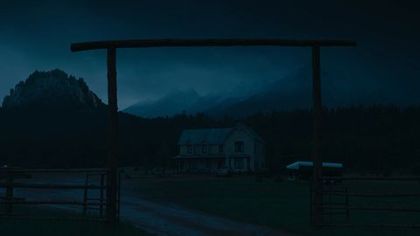}&\includegraphics[width=\cellwidth]{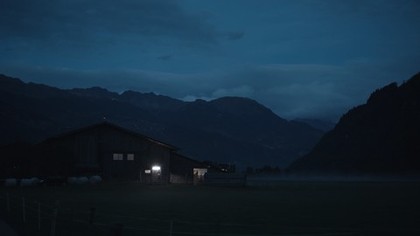}&\includegraphics[width=\cellwidth]{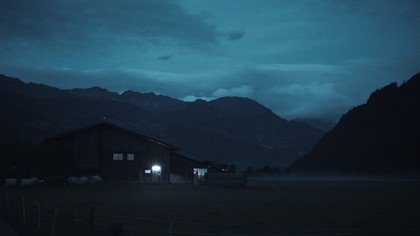}&\includegraphics[width=\cellwidth]{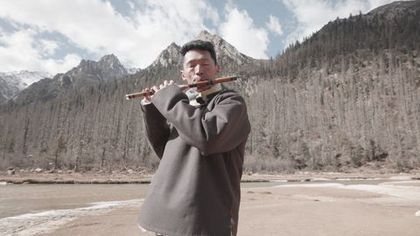}&\includegraphics[width=\cellwidth]{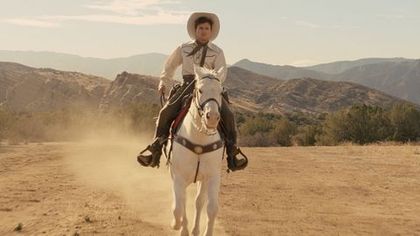}&\includegraphics[width=\cellwidth]{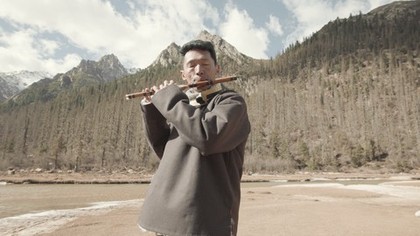}&\includegraphics[width=\cellwidth]{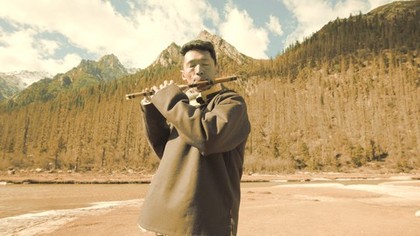} \\
    \includegraphics[width=\cellwidth]{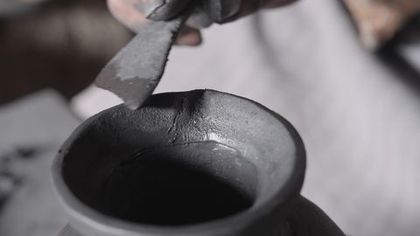}&\includegraphics[width=\cellwidth]{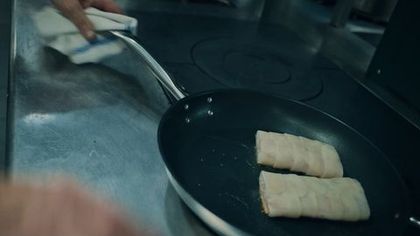}&\includegraphics[width=\cellwidth]{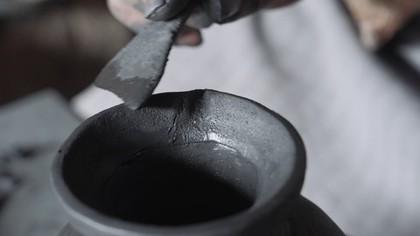}&\includegraphics[width=\cellwidth]{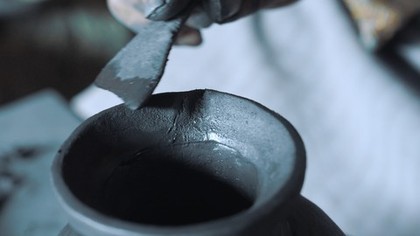}&\includegraphics[width=\cellwidth]{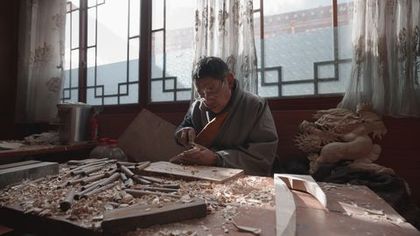}&\includegraphics[width=\cellwidth]{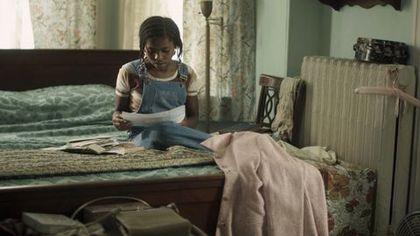}&\includegraphics[width=\cellwidth]{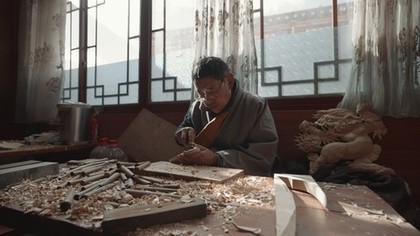}&\includegraphics[width=\cellwidth]{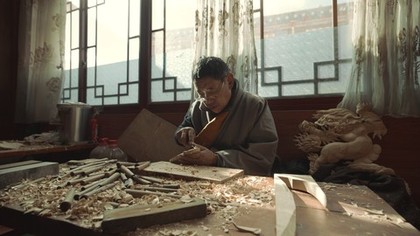} \\
    \includegraphics[width=\cellwidth]{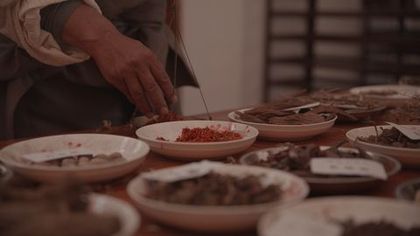}&\includegraphics[width=\cellwidth]{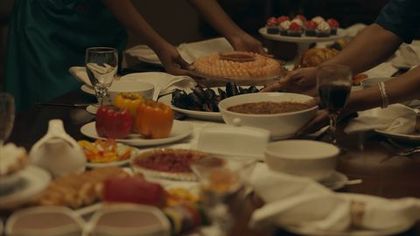}&\includegraphics[width=\cellwidth]{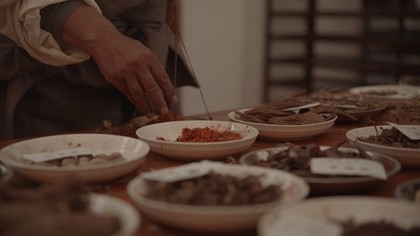}&\includegraphics[width=\cellwidth]{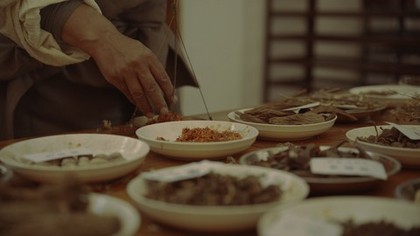}&\includegraphics[width=\cellwidth]{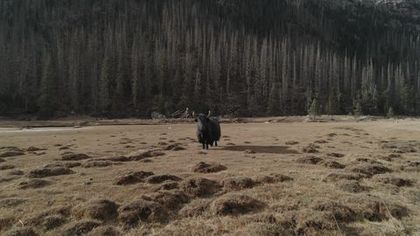}&\includegraphics[width=\cellwidth]{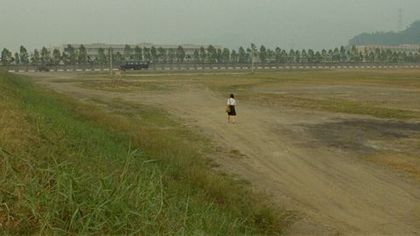}&\includegraphics[width=\cellwidth]{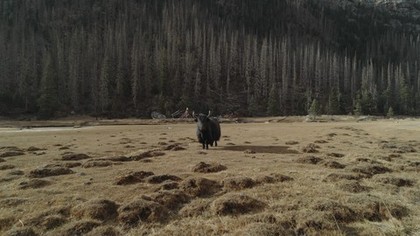}&\includegraphics[width=\cellwidth]{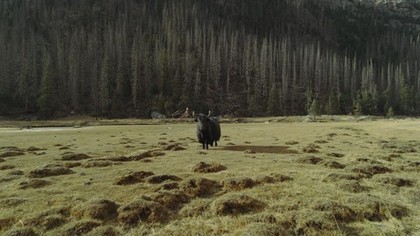} \\
    \includegraphics[width=\cellwidth]{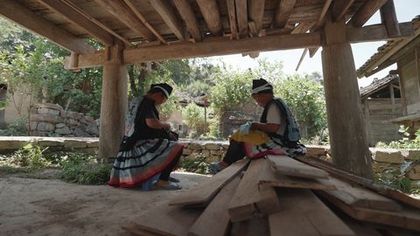}&\includegraphics[width=\cellwidth]{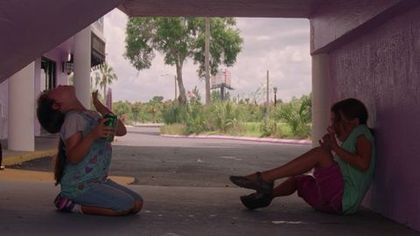}&\includegraphics[width=\cellwidth]{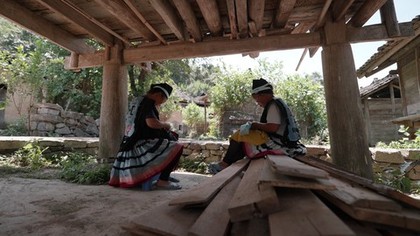}&\includegraphics[width=\cellwidth]{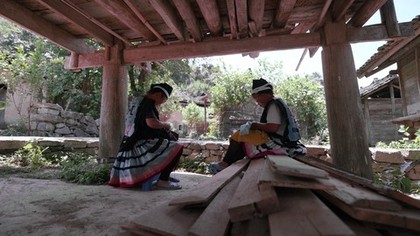}&\includegraphics[width=\cellwidth]{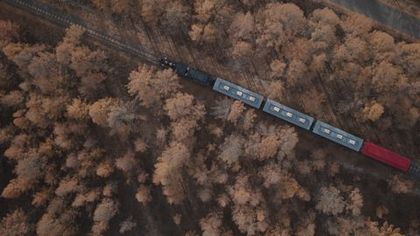}&\includegraphics[width=\cellwidth]{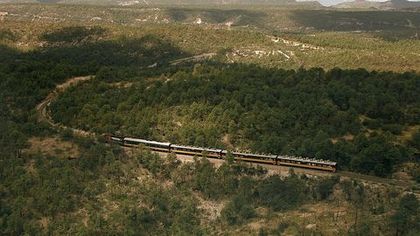}&\includegraphics[width=\cellwidth]{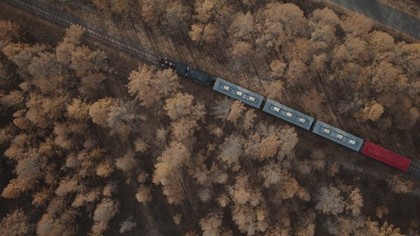}&\includegraphics[width=\cellwidth]{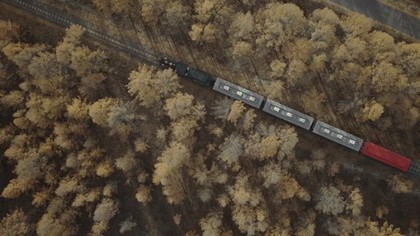} \\
    \includegraphics[width=\cellwidth]{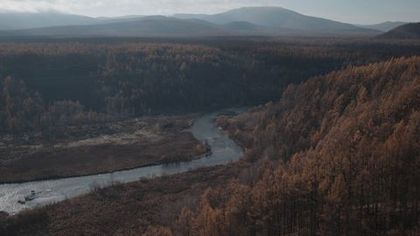}&\includegraphics[width=\cellwidth]{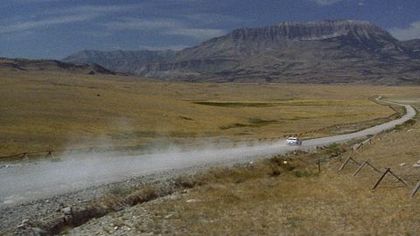}&\includegraphics[width=\cellwidth]{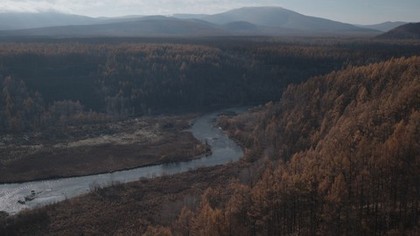}&\includegraphics[width=\cellwidth]{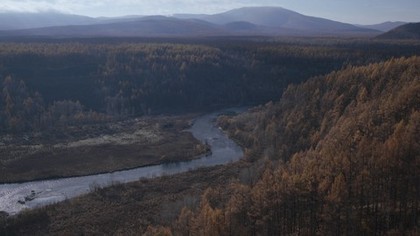}&\includegraphics[width=\cellwidth]{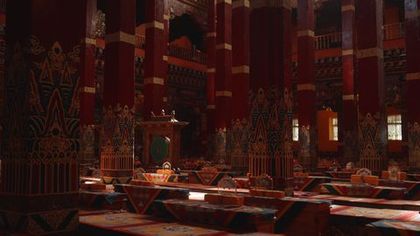}&\includegraphics[width=\cellwidth]{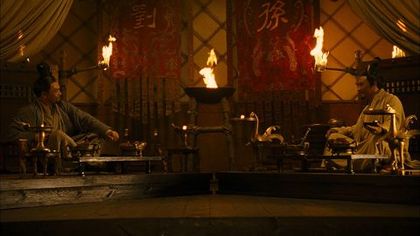}&\includegraphics[width=\cellwidth]{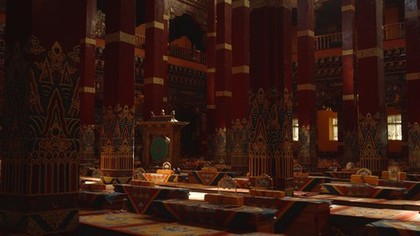}&\includegraphics[width=\cellwidth]{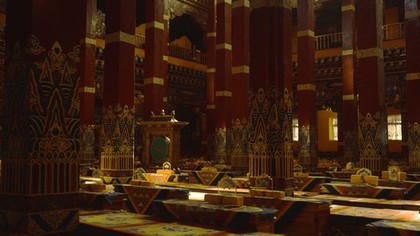} \\
    \includegraphics[width=\cellwidth]{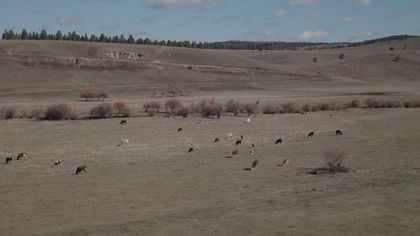}&\includegraphics[width=\cellwidth]{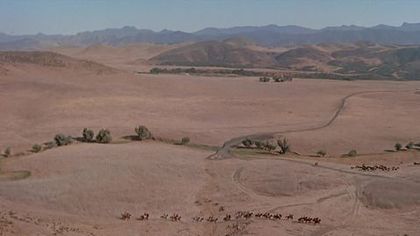}&\includegraphics[width=\cellwidth]{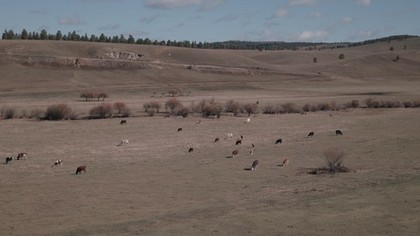}&\includegraphics[width=\cellwidth]{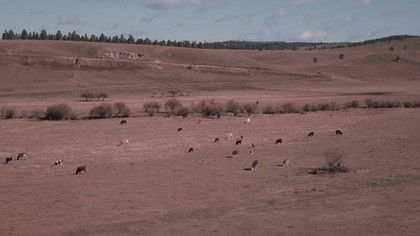}&\includegraphics[width=\cellwidth]{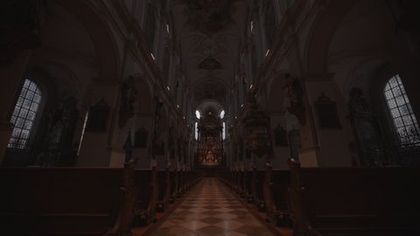}&\includegraphics[width=\cellwidth]{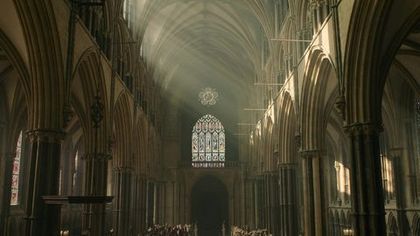}&\includegraphics[width=\cellwidth]{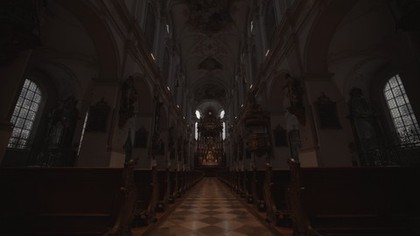}&\includegraphics[width=\cellwidth]{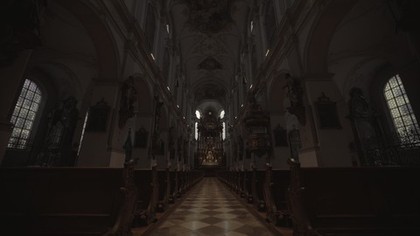} \\
    \includegraphics[width=\cellwidth]{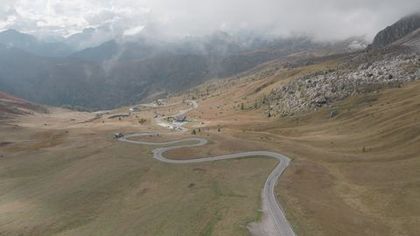}&\includegraphics[width=\cellwidth]{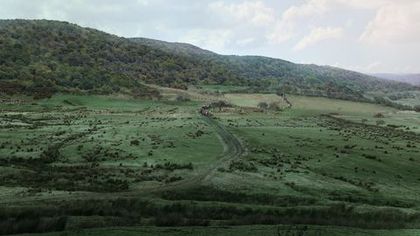}&\includegraphics[width=\cellwidth]{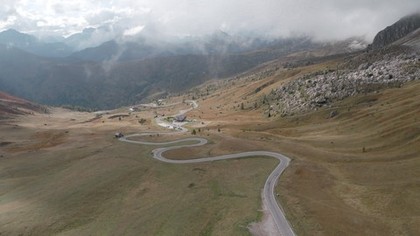}&\includegraphics[width=\cellwidth]{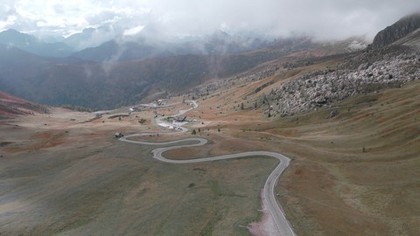}&\includegraphics[width=\cellwidth]{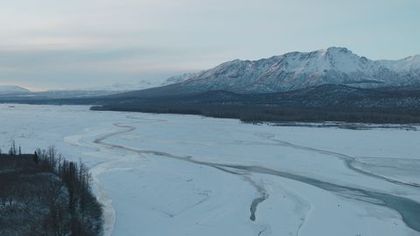}&\includegraphics[width=\cellwidth]{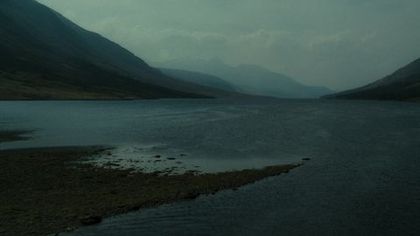}&\includegraphics[width=\cellwidth]{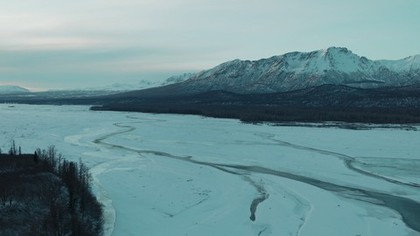}&\includegraphics[width=\cellwidth]{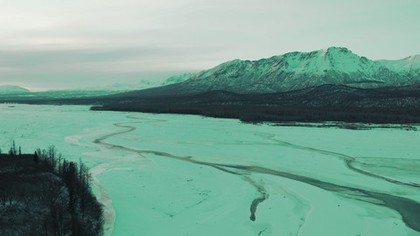} \\
    \includegraphics[width=\cellwidth]{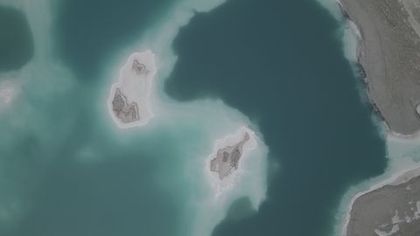}&\includegraphics[width=\cellwidth]{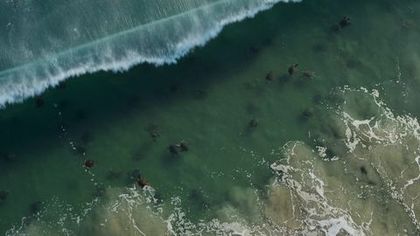}&\includegraphics[width=\cellwidth]{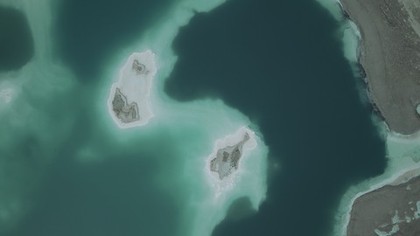}&\includegraphics[width=\cellwidth]{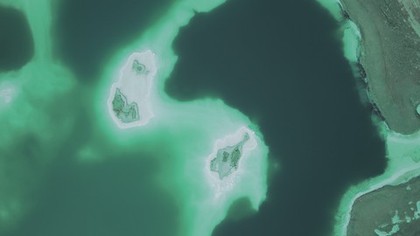}&\includegraphics[width=\cellwidth]{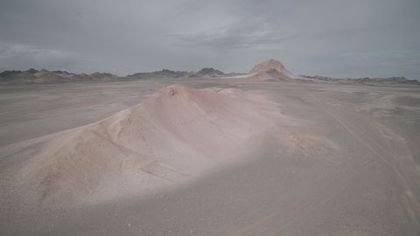}&\includegraphics[width=\cellwidth]{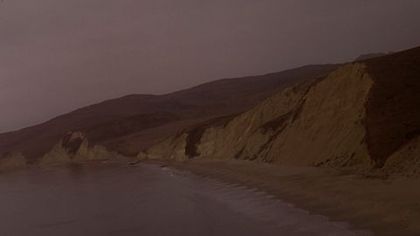}&\includegraphics[width=\cellwidth]{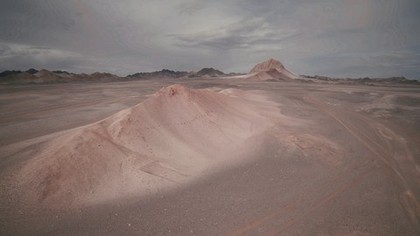}&\includegraphics[width=\cellwidth]{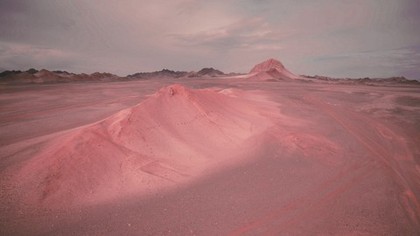} \\
    \includegraphics[width=\cellwidth]{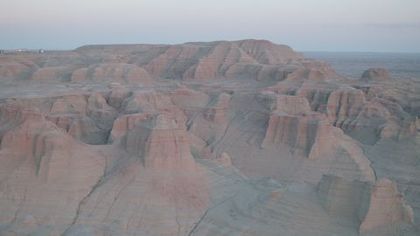}&\includegraphics[width=\cellwidth]{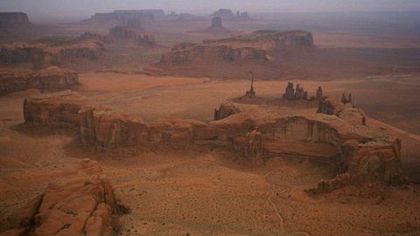}&\includegraphics[width=\cellwidth]{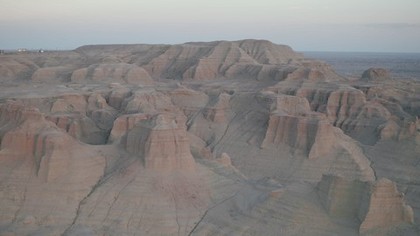}&\includegraphics[width=\cellwidth]{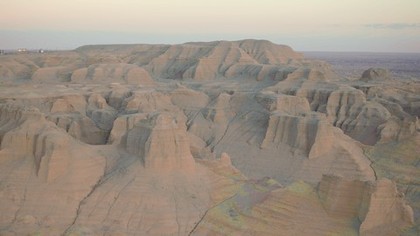}&\includegraphics[width=\cellwidth]{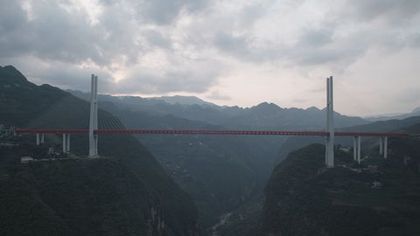}&\includegraphics[width=\cellwidth]{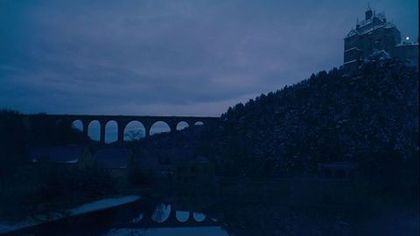}&\includegraphics[width=\cellwidth]{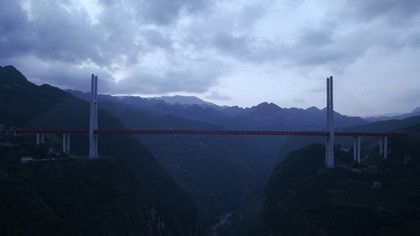}&\includegraphics[width=\cellwidth]{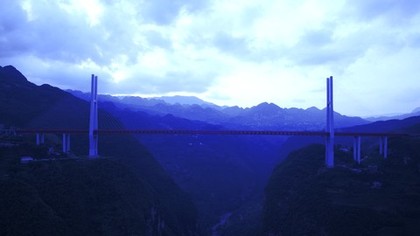} \\
    \includegraphics[width=\cellwidth]{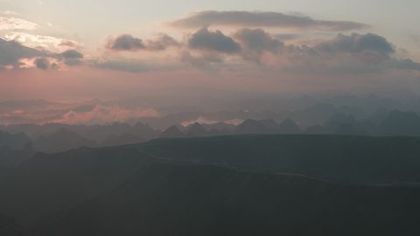}&\includegraphics[width=\cellwidth]{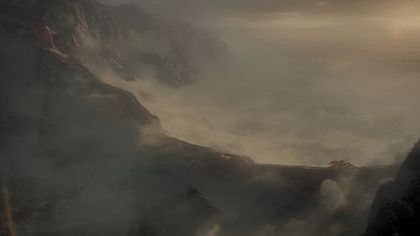}&\includegraphics[width=\cellwidth]{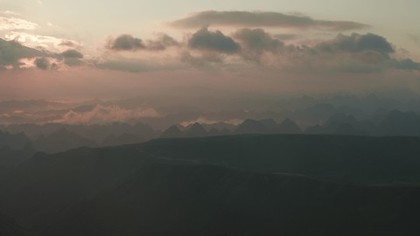}&\includegraphics[width=\cellwidth]{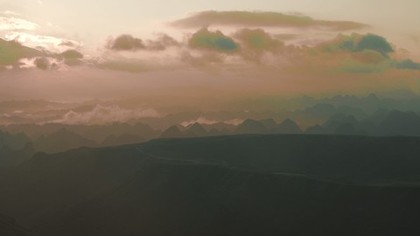}&\includegraphics[width=\cellwidth]{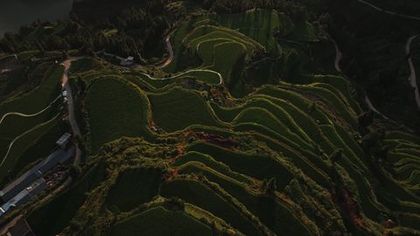}&\includegraphics[width=\cellwidth]{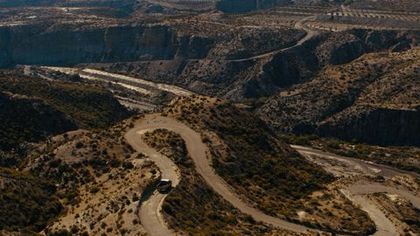}&\includegraphics[width=\cellwidth]{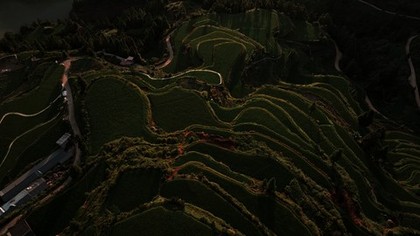}&\includegraphics[width=\cellwidth]{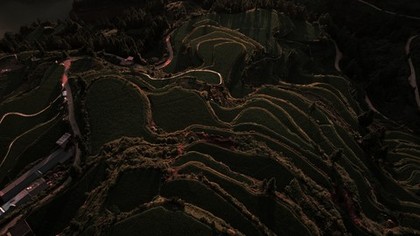} \\
    \includegraphics[width=\cellwidth]{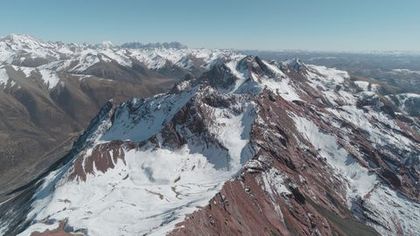}&\includegraphics[width=\cellwidth]{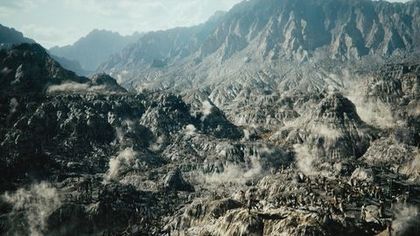}&\includegraphics[width=\cellwidth]{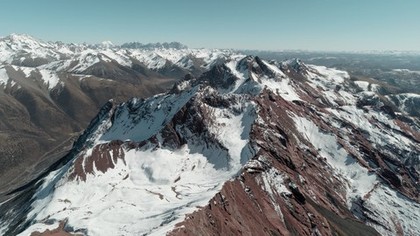}&\includegraphics[width=\cellwidth]{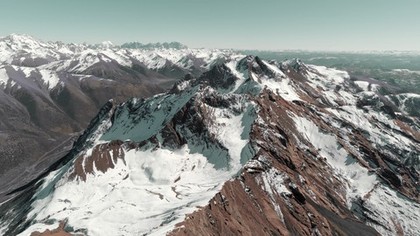}&\includegraphics[width=\cellwidth]{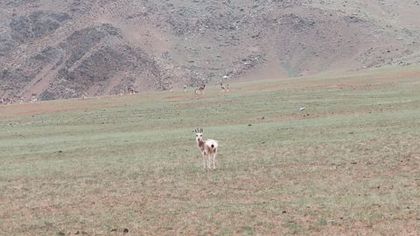}&\includegraphics[width=\cellwidth]{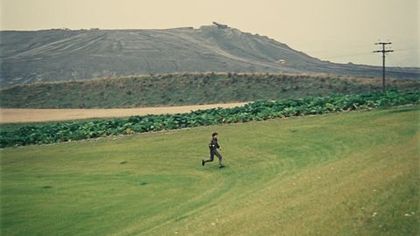}&\includegraphics[width=\cellwidth]{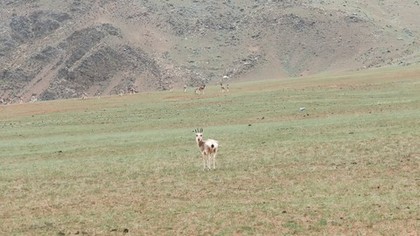}&\includegraphics[width=\cellwidth]{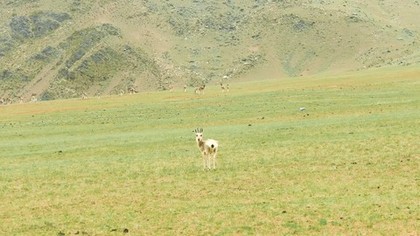} \\
    \includegraphics[width=\cellwidth]{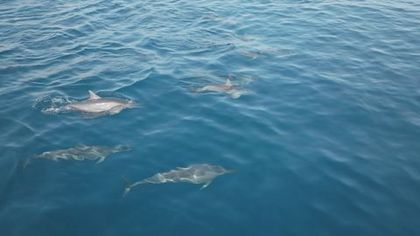}&\includegraphics[width=\cellwidth]{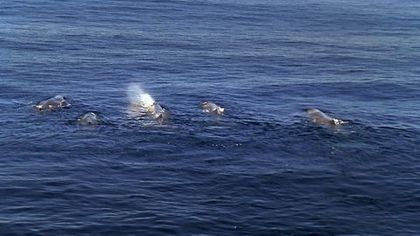}&\includegraphics[width=\cellwidth]{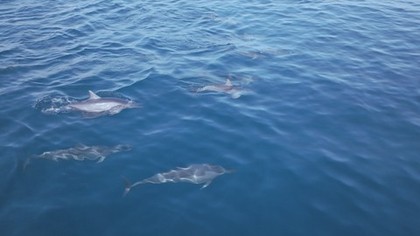}&\includegraphics[width=\cellwidth]{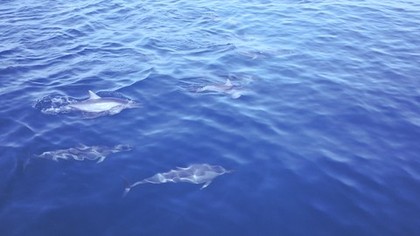}&\includegraphics[width=\cellwidth]{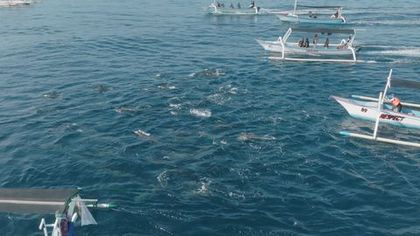}&\includegraphics[width=\cellwidth]{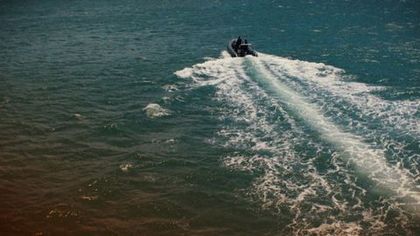}&\includegraphics[width=\cellwidth]{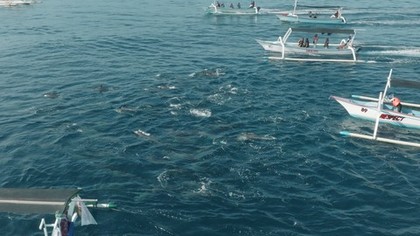}&\includegraphics[width=\cellwidth]{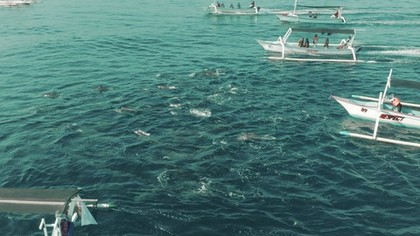} \\
  \end{tabular}
  \caption{Full PST50 unpaired test set (1/2). Each group: $X$, $Y$, PREEN ($\alpha\!=\!1$), and PREEN Vivid ($\alpha\!=\!5$).}
\end{figure*}

\begin{figure*}[p]
  \centering
  \setlength{\tabcolsep}{0pt}
  \renewcommand{\arraystretch}{0}
  \newcommand{\cellwidth}{0.122\linewidth}
  \begin{tabular}{@{}c@{}c@{}c@{}c@{\hspace{4pt}}c@{}c@{}c@{}c@{}}
    \scriptsize\bfseries $X$ & \scriptsize\bfseries $Y$ & \scriptsize\bfseries PREEN & \scriptsize\bfseries PREEN Vivid & \scriptsize\bfseries $X$ & \scriptsize\bfseries $Y$ & \scriptsize\bfseries PREEN & \scriptsize\bfseries PREEN Vivid \\
    \includegraphics[width=\cellwidth]{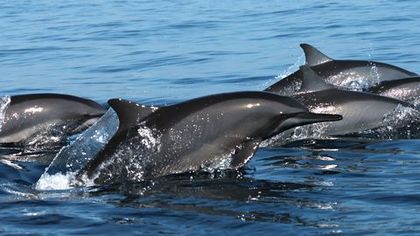}&\includegraphics[width=\cellwidth]{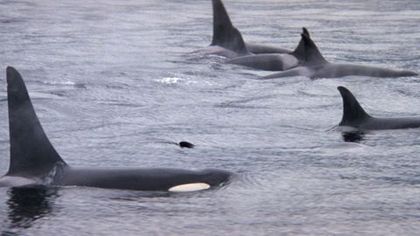}&\includegraphics[width=\cellwidth]{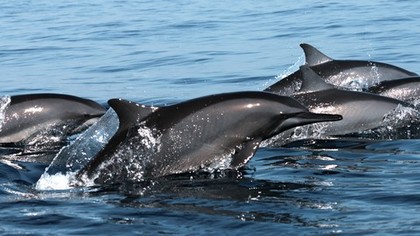}&\includegraphics[width=\cellwidth]{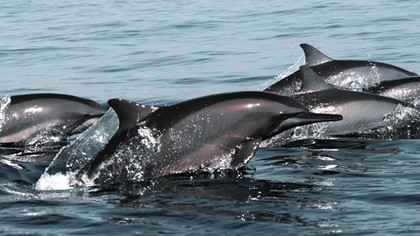}&\includegraphics[width=\cellwidth]{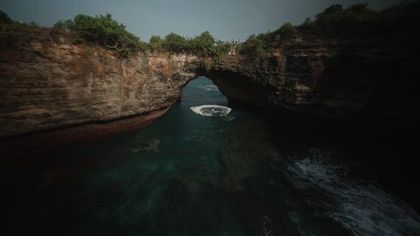}&\includegraphics[width=\cellwidth]{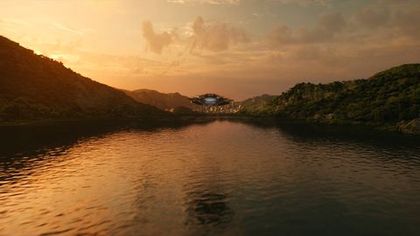}&\includegraphics[width=\cellwidth]{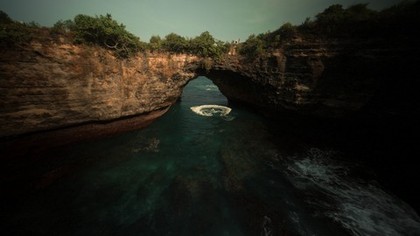}&\includegraphics[width=\cellwidth]{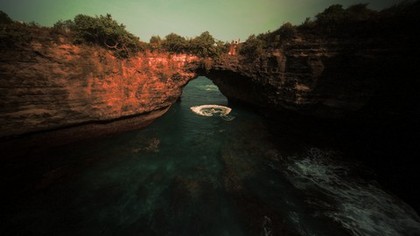} \\
    \includegraphics[width=\cellwidth]{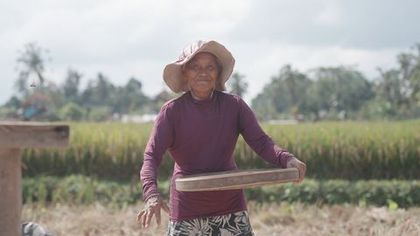}&\includegraphics[width=\cellwidth]{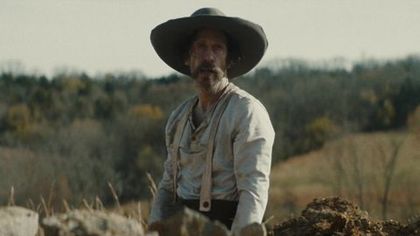}&\includegraphics[width=\cellwidth]{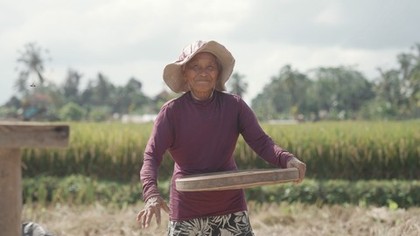}&\includegraphics[width=\cellwidth]{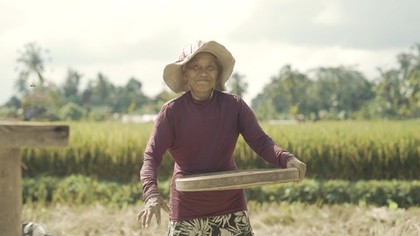}&\includegraphics[width=\cellwidth]{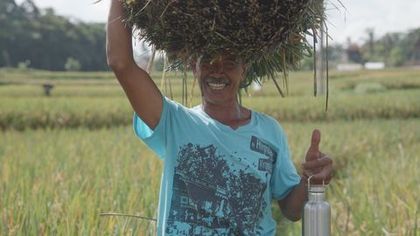}&\includegraphics[width=\cellwidth]{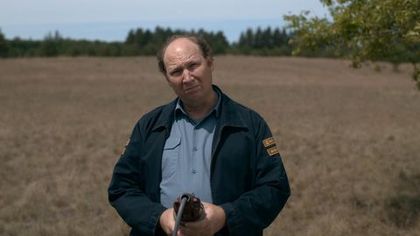}&\includegraphics[width=\cellwidth]{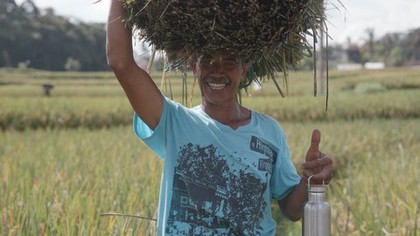}&\includegraphics[width=\cellwidth]{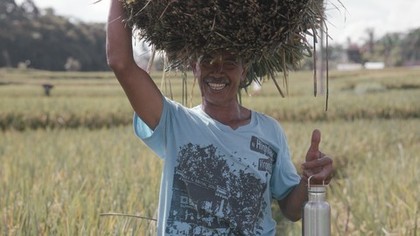} \\
    \includegraphics[width=\cellwidth]{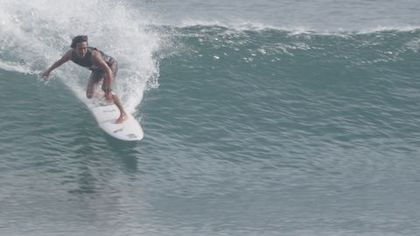}&\includegraphics[width=\cellwidth]{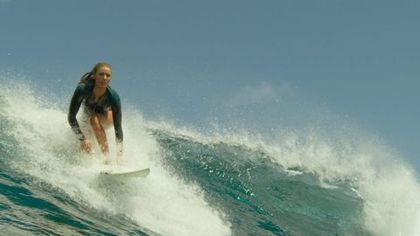}&\includegraphics[width=\cellwidth]{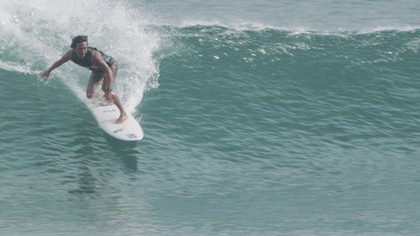}&\includegraphics[width=\cellwidth]{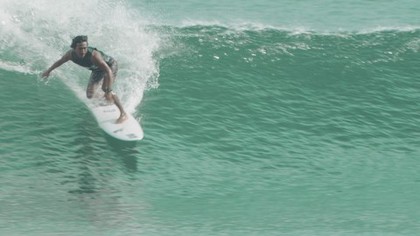}&\includegraphics[width=\cellwidth]{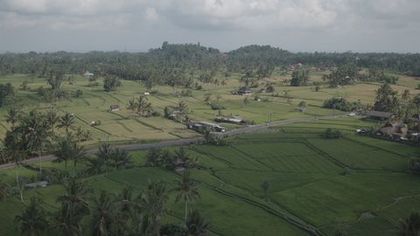}&\includegraphics[width=\cellwidth]{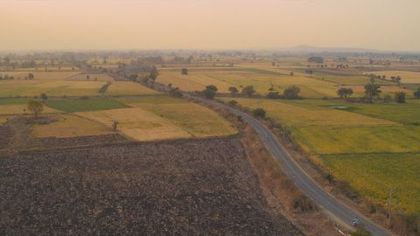}&\includegraphics[width=\cellwidth]{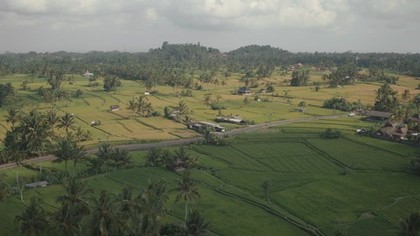}&\includegraphics[width=\cellwidth]{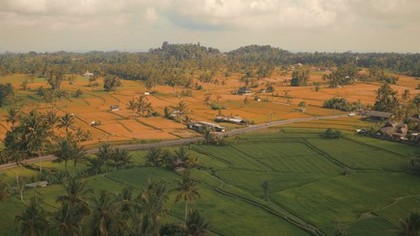} \\
    \includegraphics[width=\cellwidth]{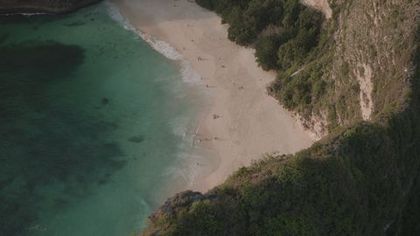}&\includegraphics[width=\cellwidth]{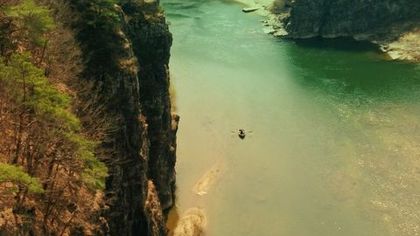}&\includegraphics[width=\cellwidth]{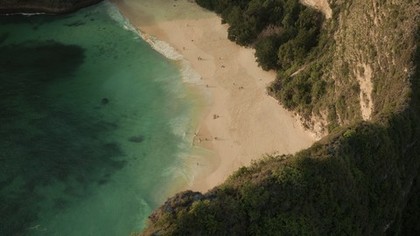}&\includegraphics[width=\cellwidth]{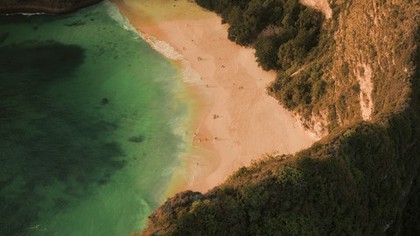}&\includegraphics[width=\cellwidth]{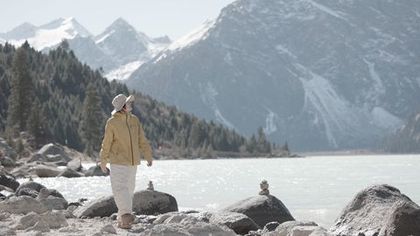}&\includegraphics[width=\cellwidth]{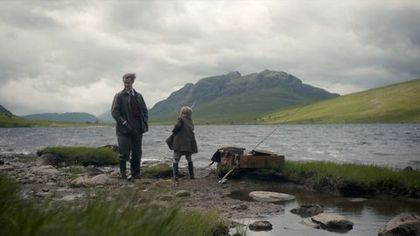}&\includegraphics[width=\cellwidth]{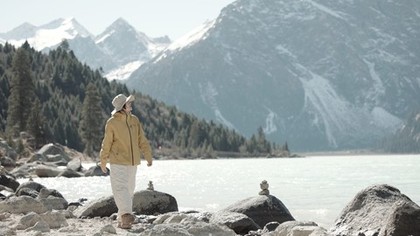}&\includegraphics[width=\cellwidth]{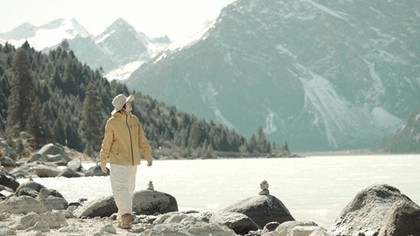} \\
    \includegraphics[width=\cellwidth]{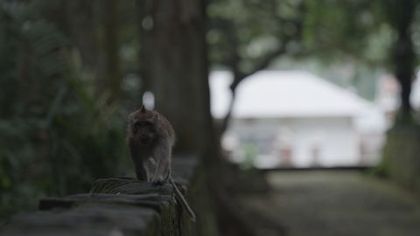}&\includegraphics[width=\cellwidth]{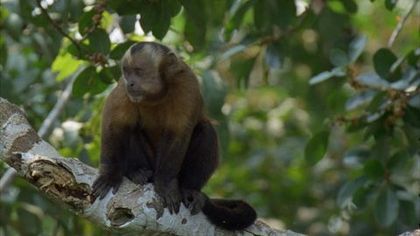}&\includegraphics[width=\cellwidth]{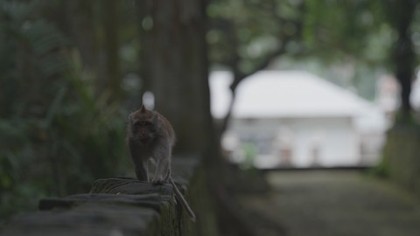}&\includegraphics[width=\cellwidth]{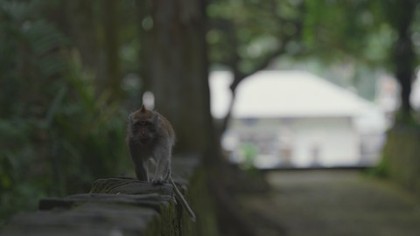}&\includegraphics[width=\cellwidth]{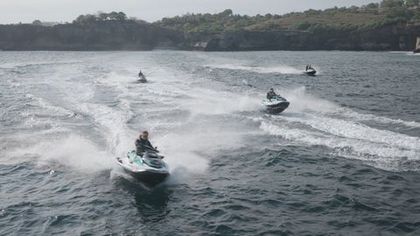}&\includegraphics[width=\cellwidth]{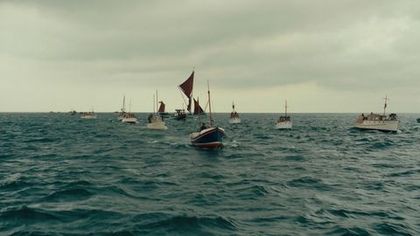}&\includegraphics[width=\cellwidth]{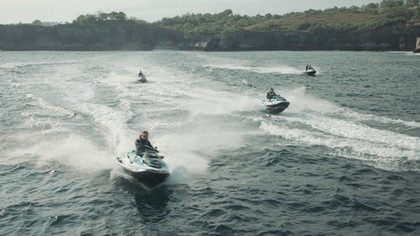}&\includegraphics[width=\cellwidth]{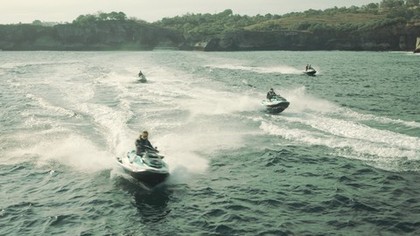} \\
    \includegraphics[width=\cellwidth]{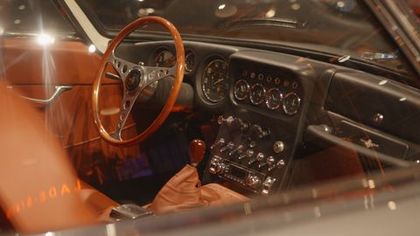}&\includegraphics[width=\cellwidth]{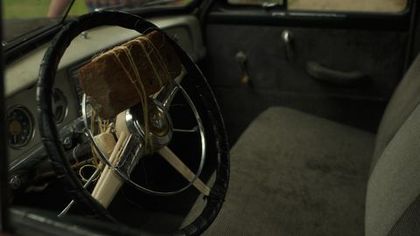}&\includegraphics[width=\cellwidth]{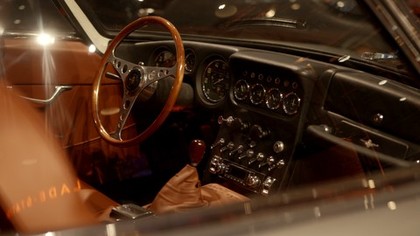}&\includegraphics[width=\cellwidth]{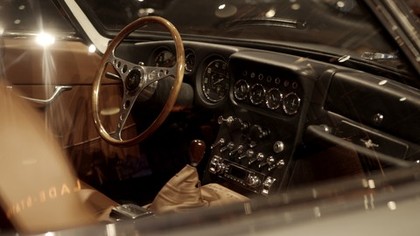}&\includegraphics[width=\cellwidth]{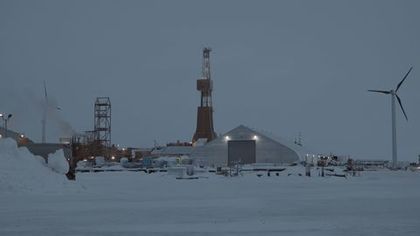}&\includegraphics[width=\cellwidth]{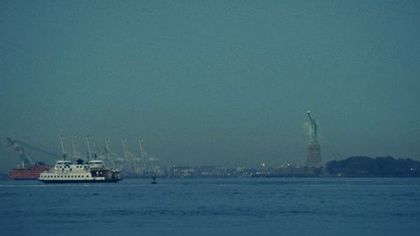}&\includegraphics[width=\cellwidth]{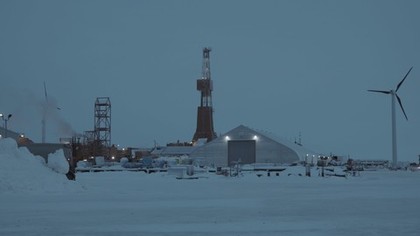}&\includegraphics[width=\cellwidth]{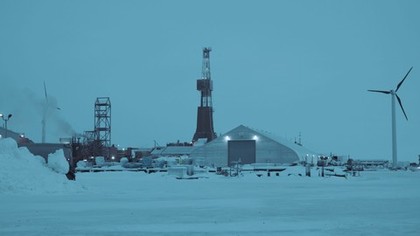} \\
    \includegraphics[width=\cellwidth]{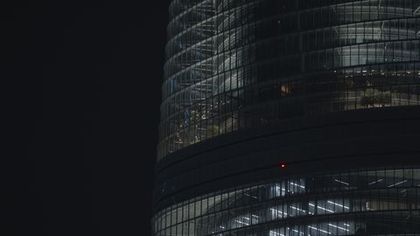}&\includegraphics[width=\cellwidth]{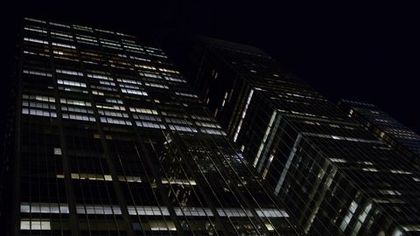}&\includegraphics[width=\cellwidth]{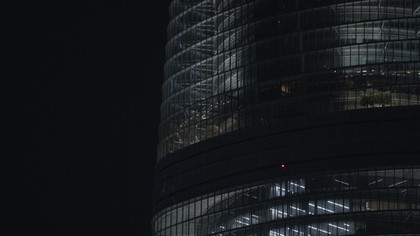}&\includegraphics[width=\cellwidth]{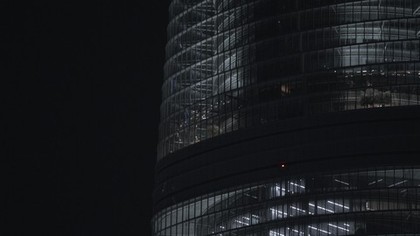}&\includegraphics[width=\cellwidth]{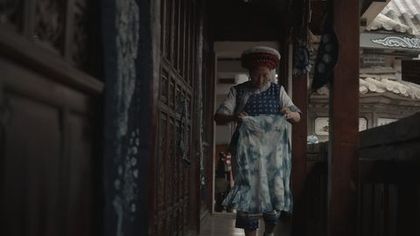}&\includegraphics[width=\cellwidth]{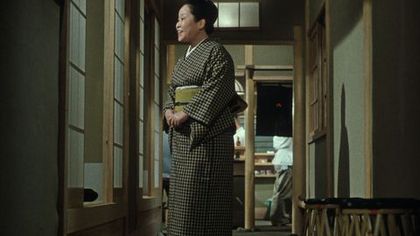}&\includegraphics[width=\cellwidth]{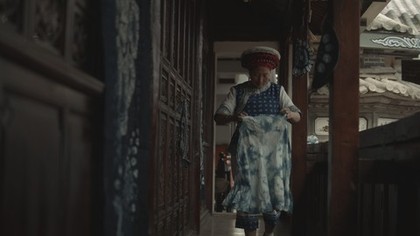}&\includegraphics[width=\cellwidth]{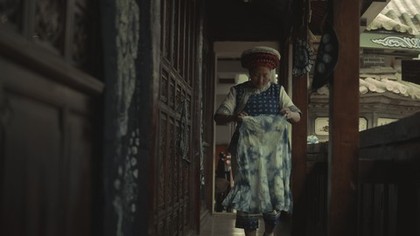} \\
    \includegraphics[width=\cellwidth]{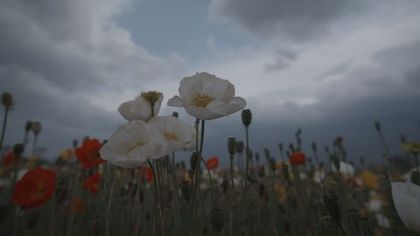}&\includegraphics[width=\cellwidth]{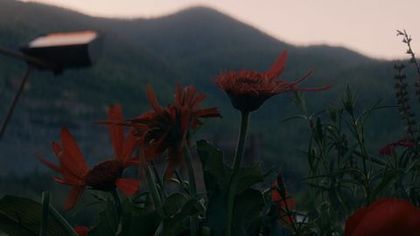}&\includegraphics[width=\cellwidth]{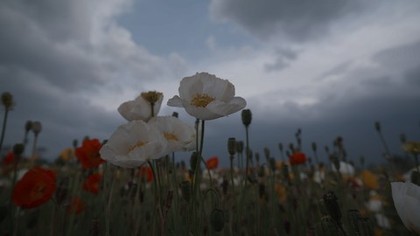}&\includegraphics[width=\cellwidth]{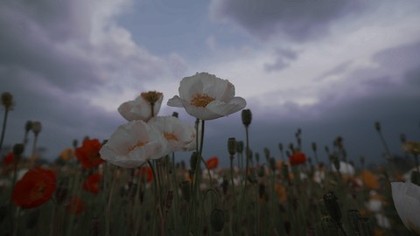}&\includegraphics[width=\cellwidth]{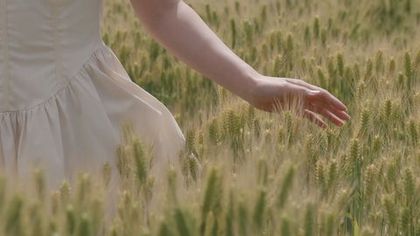}&\includegraphics[width=\cellwidth]{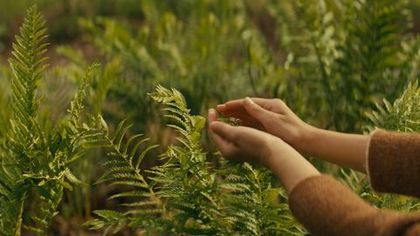}&\includegraphics[width=\cellwidth]{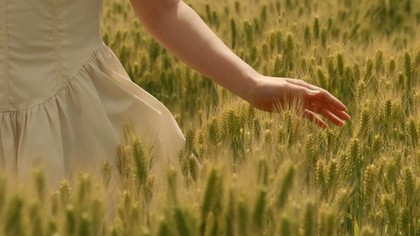}&\includegraphics[width=\cellwidth]{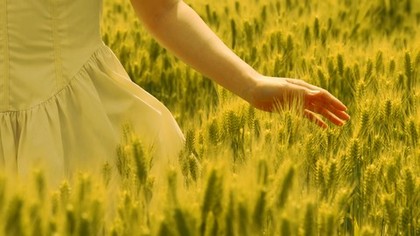} \\
    \includegraphics[width=\cellwidth]{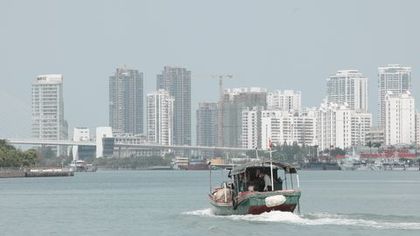}&\includegraphics[width=\cellwidth]{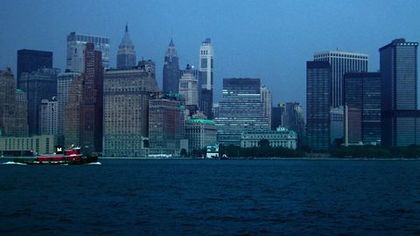}&\includegraphics[width=\cellwidth]{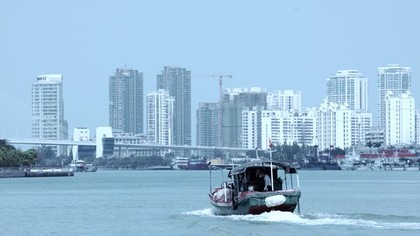}&\includegraphics[width=\cellwidth]{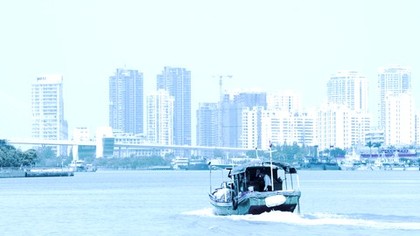}&\includegraphics[width=\cellwidth]{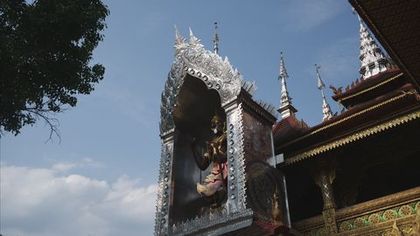}&\includegraphics[width=\cellwidth]{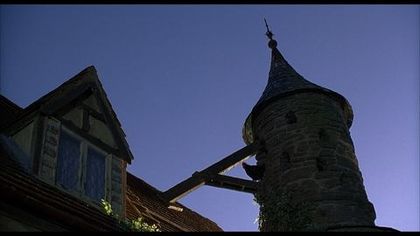}&\includegraphics[width=\cellwidth]{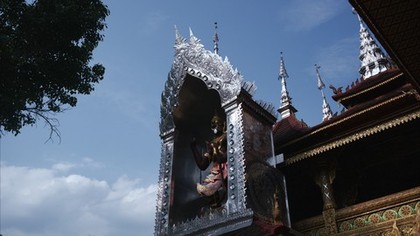}&\includegraphics[width=\cellwidth]{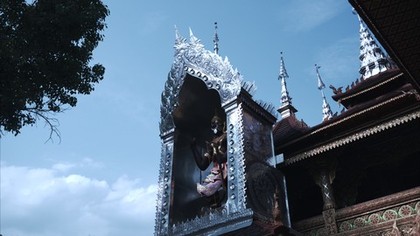} \\
    \includegraphics[width=\cellwidth]{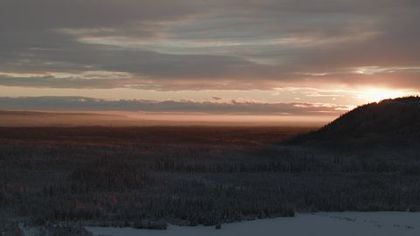}&\includegraphics[width=\cellwidth]{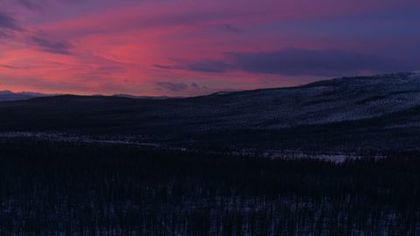}&\includegraphics[width=\cellwidth]{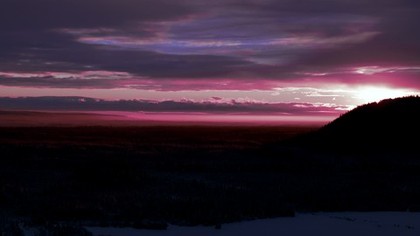}&\includegraphics[width=\cellwidth]{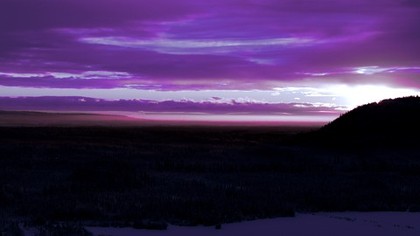}&\includegraphics[width=\cellwidth]{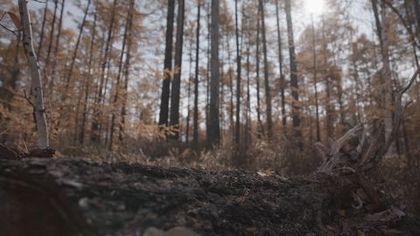}&\includegraphics[width=\cellwidth]{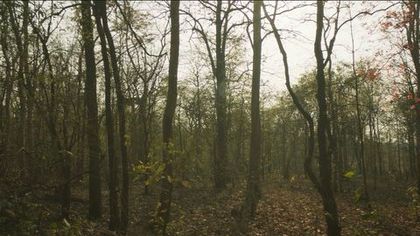}&\includegraphics[width=\cellwidth]{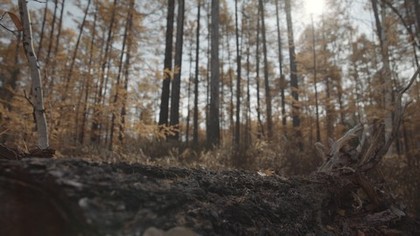}&\includegraphics[width=\cellwidth]{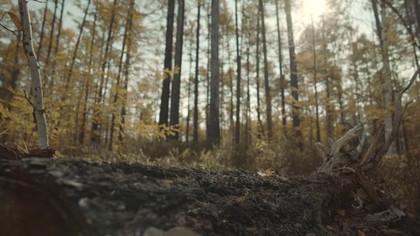} \\
    \includegraphics[width=\cellwidth]{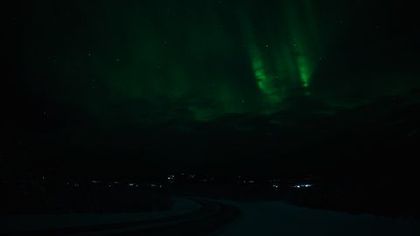}&\includegraphics[width=\cellwidth]{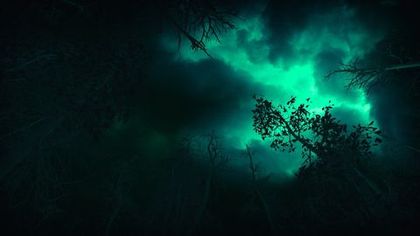}&\includegraphics[width=\cellwidth]{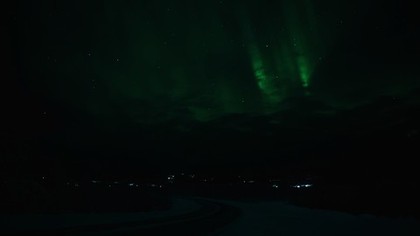}&\includegraphics[width=\cellwidth]{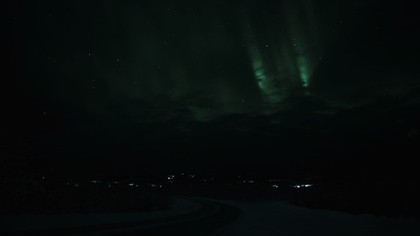}&\includegraphics[width=\cellwidth]{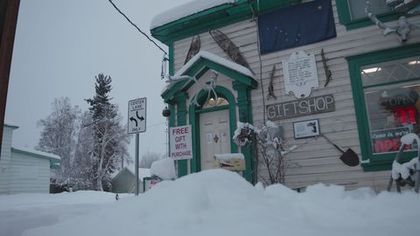}&\includegraphics[width=\cellwidth]{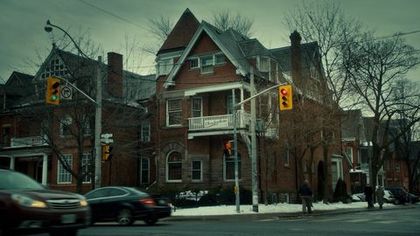}&\includegraphics[width=\cellwidth]{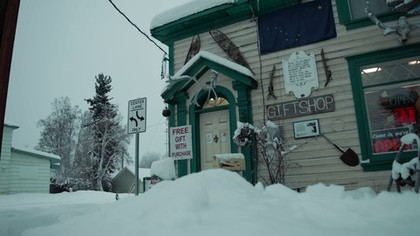}&\includegraphics[width=\cellwidth]{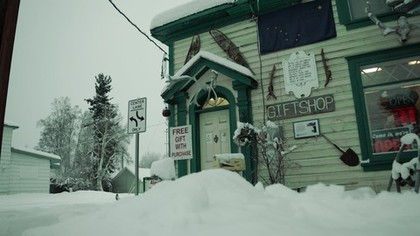} \\
    \includegraphics[width=\cellwidth]{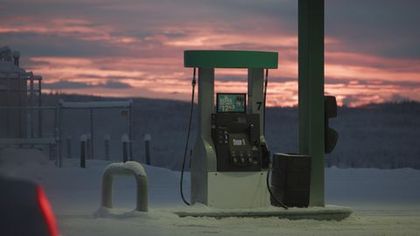}&\includegraphics[width=\cellwidth]{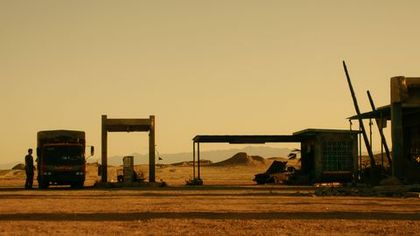}&\includegraphics[width=\cellwidth]{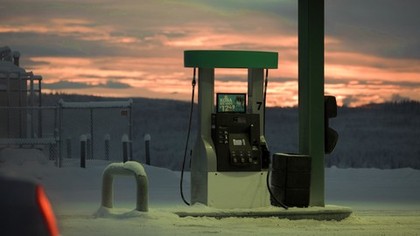}&\includegraphics[width=\cellwidth]{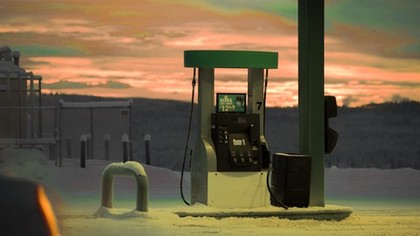}&\includegraphics[width=\cellwidth]{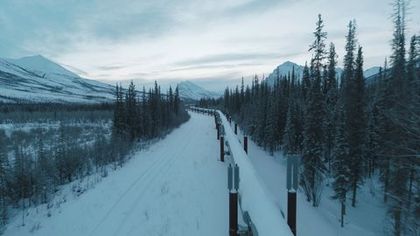}&\includegraphics[width=\cellwidth]{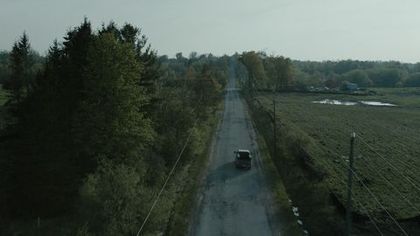}&\includegraphics[width=\cellwidth]{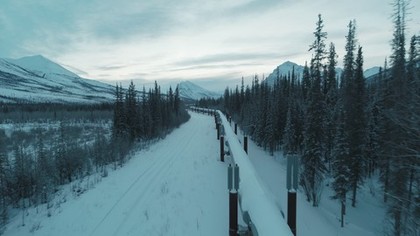}&\includegraphics[width=\cellwidth]{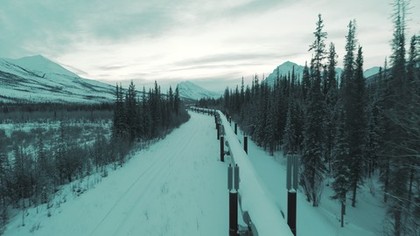} \\
  \end{tabular}
  \caption{Full PST50 unpaired test set (2/2). Each group: $X$, $Y$, PREEN ($\alpha\!=\!1$), and PREEN Vivid ($\alpha\!=\!5$).}
  \label{supp:fig:pst50full:u2}
\end{figure*}

\end{document}